\documentclass[10pt,twocolumn,letterpaper]{article}

\usepackage{iccv}
\usepackage{times}
\usepackage{epsfig}
\usepackage{graphicx}
\usepackage{amsmath}
\usepackage{amssymb}

\usepackage{booktabs}
\usepackage{indentfirst}
\usepackage{array}
\usepackage{cellspace}
\usepackage{adjustbox}
\usepackage{comment}
\graphicspath{ {./images/} }

\usepackage[breaklinks=true,bookmarks=false]{hyperref}

\iccvfinalcopy 

\def\iccvPaperID{****} 

\ificcvfinal\fi

\begin{document}

\title{Neural Style Transfer for Vector Graphics}

\author{Valeria Efimova\\
ITMO University\\
{\tt\small vefimova@itmo.ru}
\and
Artyom Chebykin\\
SUAI\\
{\tt\small chebykin.aa@students.guap.ru}
\and
Ivan Jarsky\\
ITMO University\\
{\tt\small ivanjarsky@niuitmo.ru}
\and
Evgenii Prosvirnin\\
ITMO University\\
{\tt\small proevgenii19@gmail.com}
\and
Andrey Filchenkov\\
ITMO University\\
{\tt\small afilchenkov@itmo.ru}
}

\maketitle
\ificcvfinal\thispagestyle{empty}\fi

\begin{abstract}
   Neural style transfer draws researchers' attention, but the interest focuses on bitmap images. Various models have been developed for bitmap image generation both online and offline with arbitrary and pre-trained styles. However, the style transfer between vector images has not almost been considered. Our research shows that applying standard content and style losses insignificantly changes the vector image drawing style because the structure of vector primitives differs a lot from pixels. To handle this problem, we introduce new loss functions. We also develop a new method based on differentiable rasterization that uses these loss functions and can change the color and shape parameters of the content image corresponding to the drawing of the style image. 
   Qualitative experiments demonstrate the effectiveness of the proposed VectorNST method compared with the state-of-the-art neural style transfer approaches for bitmap images and the only existing approach for stylizing vector images, DiffVG.
   Although the proposed model does not achieve the quality and smoothness of style transfer between bitmap images, we consider our work an important early step in this area.
   VectorNST code and demo service are available at~\url{https://github.com/IzhanVarsky/VectorNST}.
\end{abstract}

\begin{figure}[t]
\begin{center}
\begin{tabular}{>{\centering\arraybackslash}m{2cm}
>{\centering\arraybackslash}m{2cm}
>{\centering\arraybackslash}m{2cm}}
    {\small Content Image} &
    {\small Style Image} &
    {\small Result} \\
    \midrule
    \includegraphics[width=2cm, height=2.5cm]{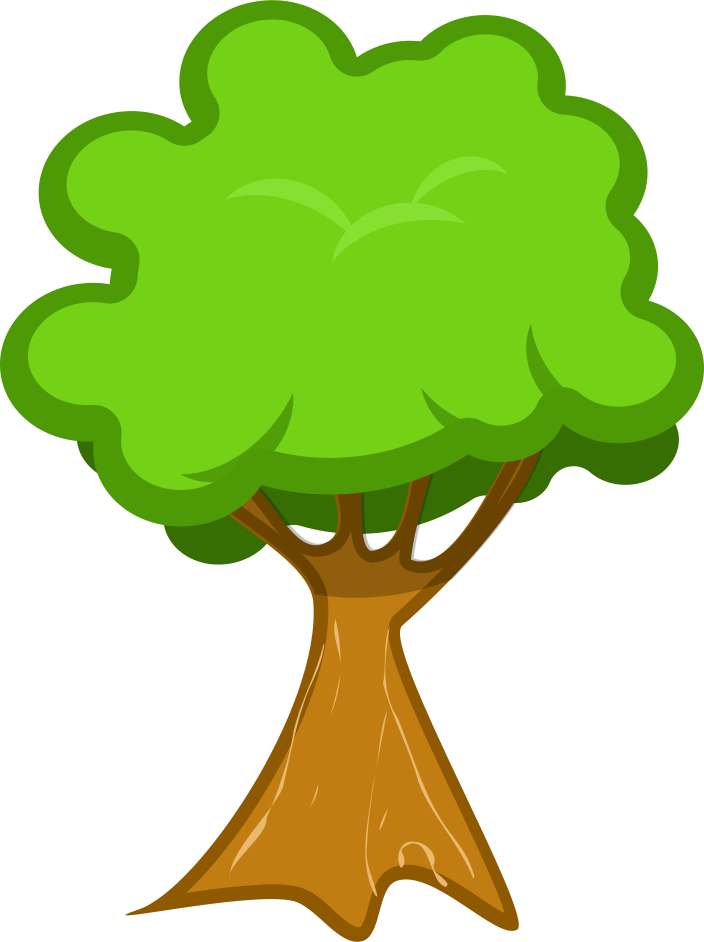} &
    \includegraphics[width=2cm, height=2.5cm]{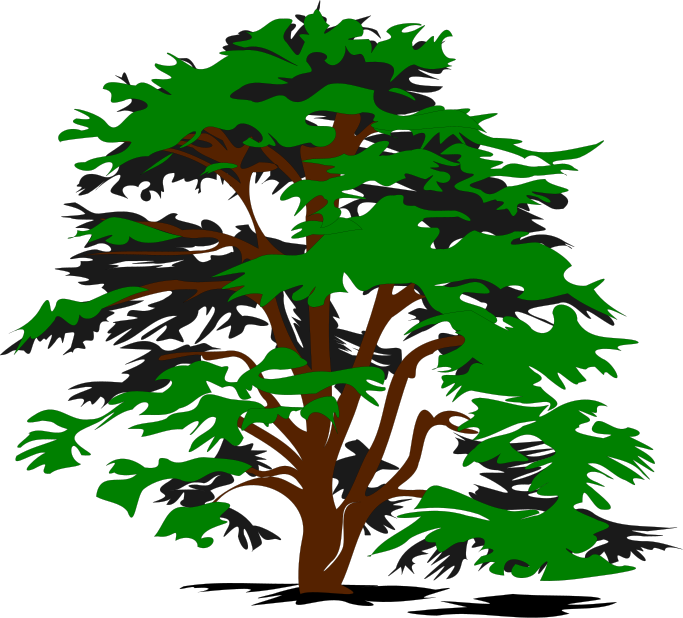}&
    \includegraphics[width=2cm, height=2.5cm]{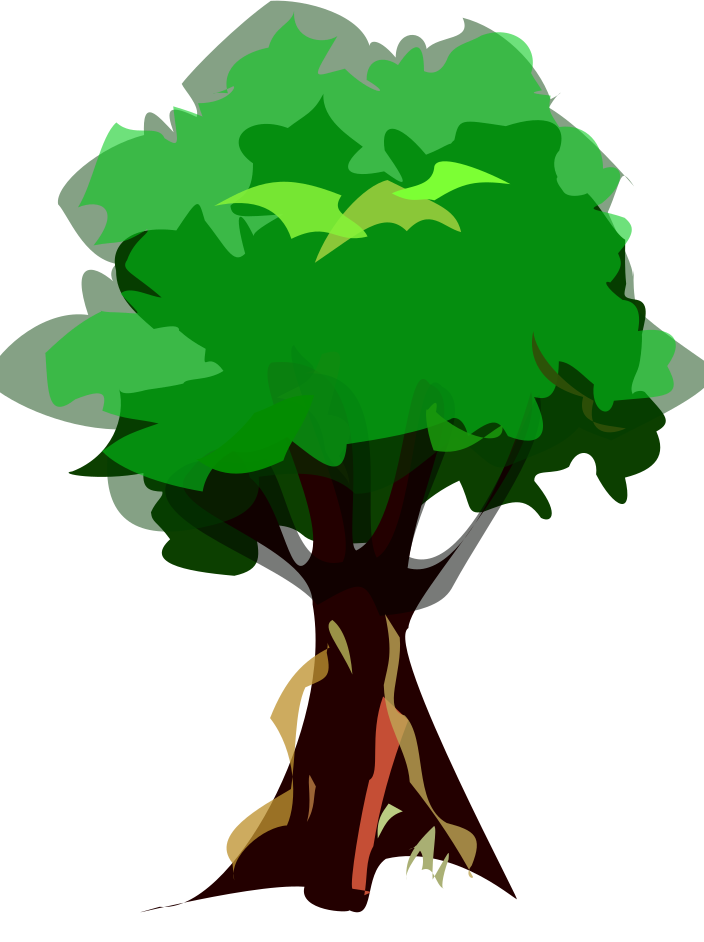} \\
    \midrule
    \includegraphics[width=2cm, height=1.2cm]{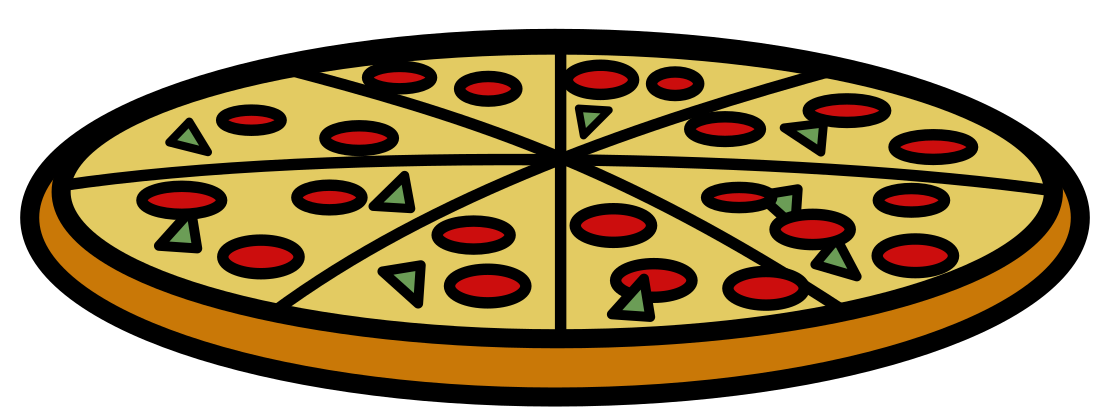}&
    \includegraphics[width=2cm, height=1.2cm]{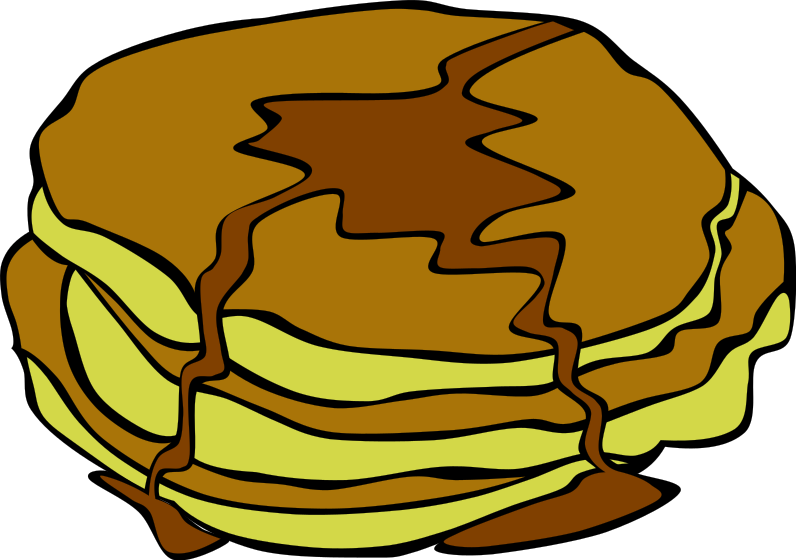}&
    \includegraphics[width=2cm, height=1.2cm]{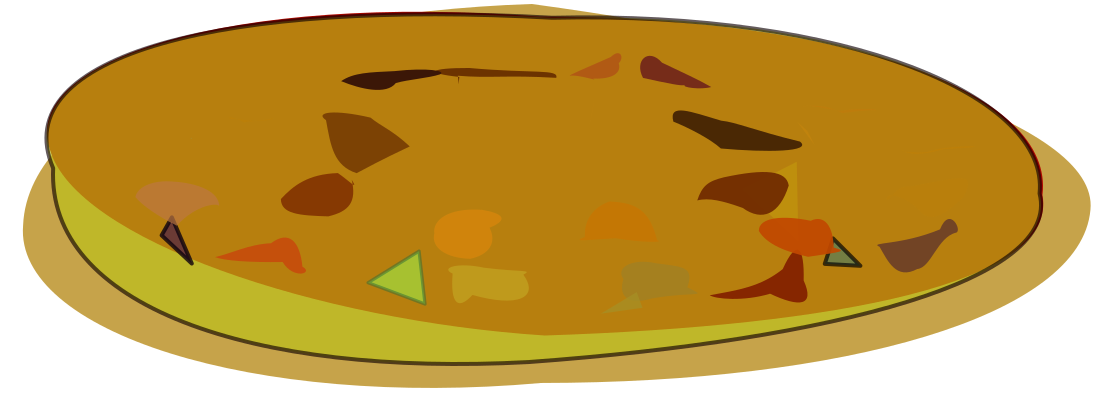} \\
    \midrule
    \includegraphics[width=2cm, height=1.6cm]{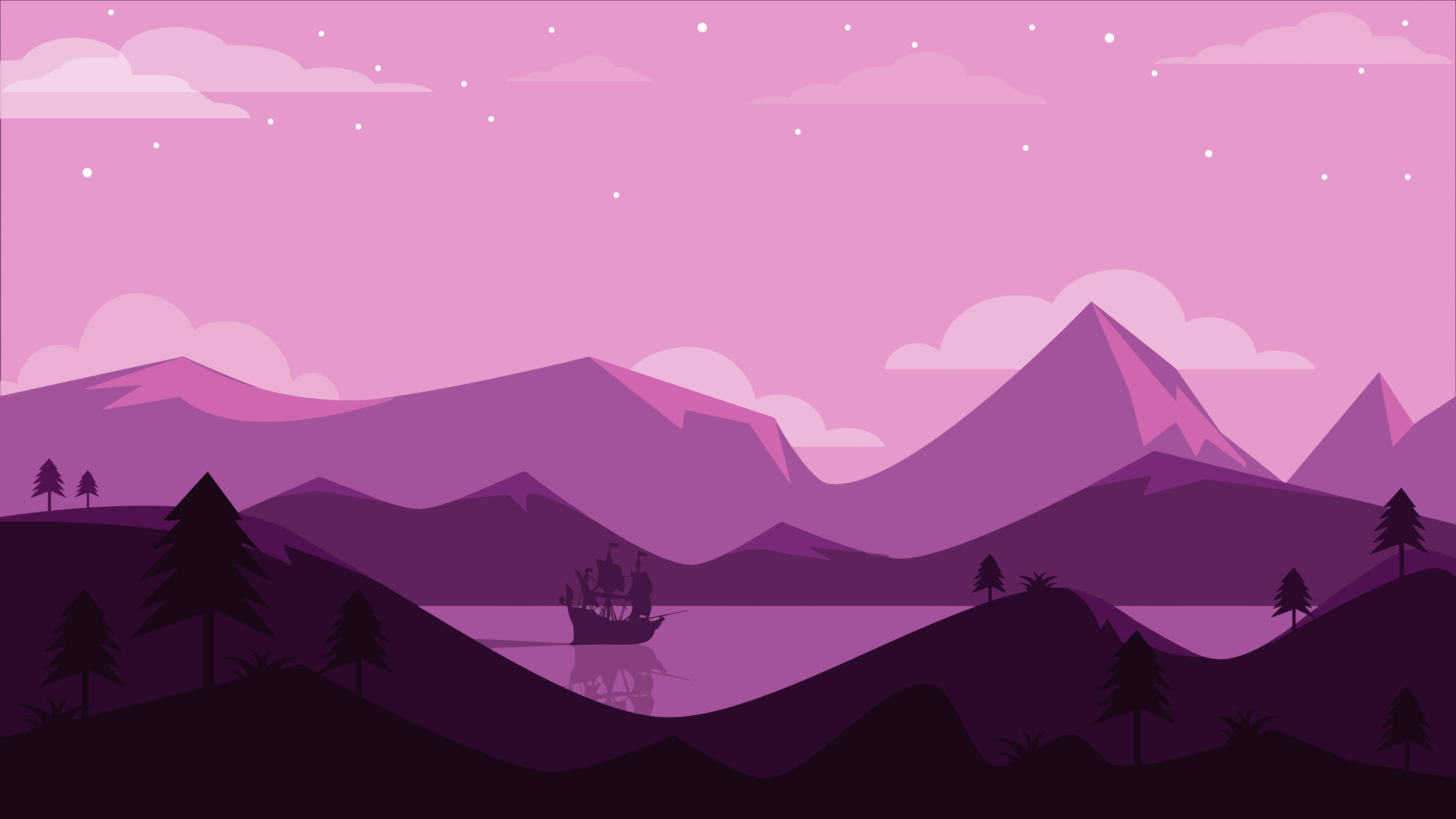}&
    \includegraphics[width=2cm, height=1.6cm]{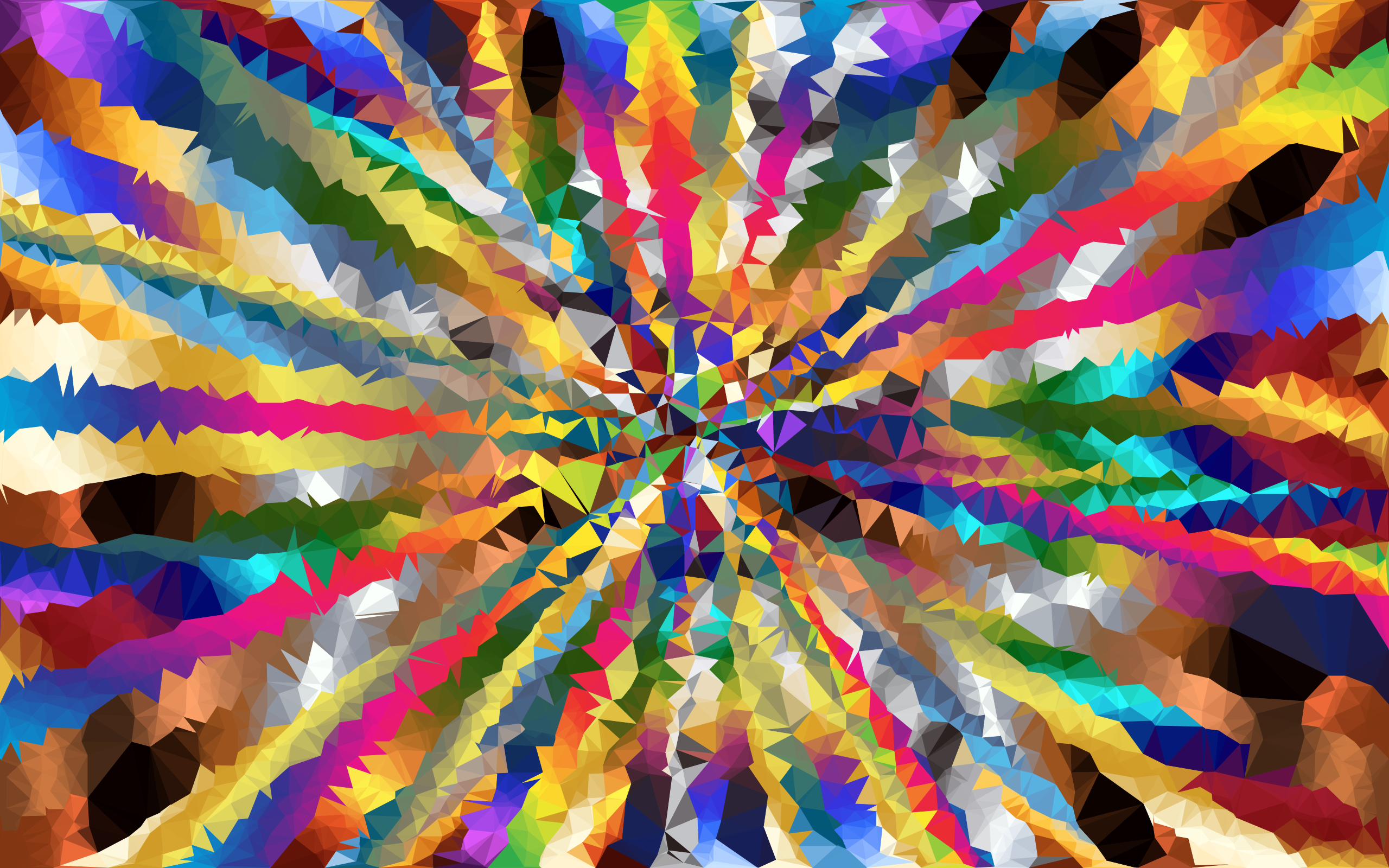}&
    \includegraphics[width=2cm, height=1.6cm]{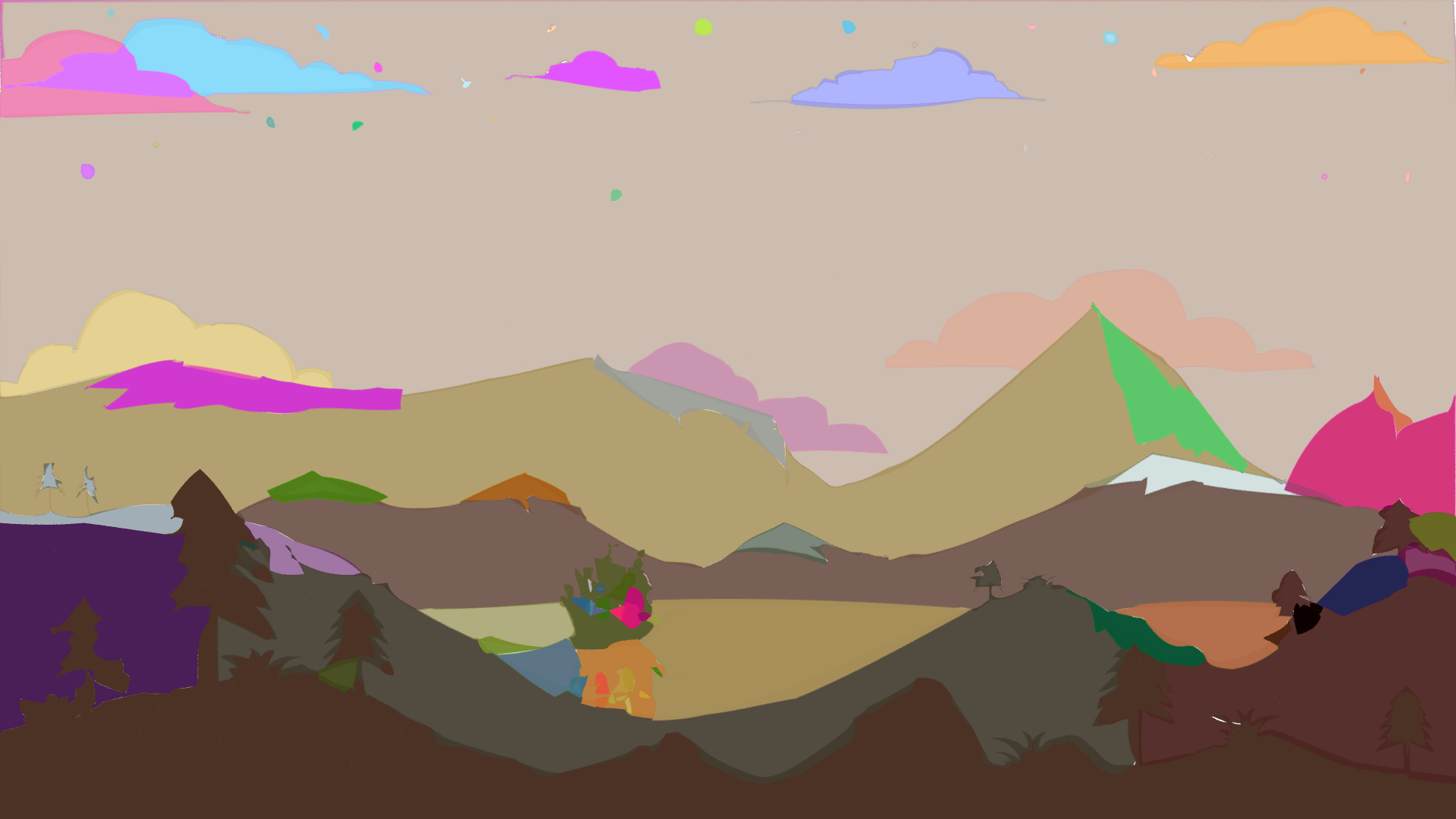}
\end{tabular}
\end{center}
\caption{We propose a novel neural style transfer method VectorNST for vector graphics. It takes as inputs a vector content image and some style image and produces a resulting vector image with a style from the style image transferred to the content image.}\label{fig:res}
\end{figure}


\section{Introduction}
\label{sec:intro}
Style transfer is a task of computer vision aiming to create new visual art objects. 
Its objective is to synthesize an image, which combines recognizable style patterns of a style image and preserves the subject of a content image.

The pioneering work of Gatys~\etal~\cite{gatys2015neural} in the field of neural style transfer (NST) showed that correlations between image representations extracted from deep neural networks could capture the visual style of an image. Based on this, they proposed the first NST method.
Using Gram matrices-based loss functions and training feed-forward neural networks~\cite{li2017demystifying, ulyanov2016texture, li2016combining, johnson2016perceptual}, utilizing one model for multiple styles~\cite{dumoulin2016learned} and many other essential improvements to the basic method have been proposed. 
The authors of~\cite{deng2022stytr2} suggested an approach for stylizing images using transformer-based architecture. 
Contrastive learning strategy is used in the CAST~\cite{ma2022towards} for training style transfer generator.
Dual language-image encoder CLIP~\cite{radford2021learning} was used for the image generation and stylization~\cite{kwon2022clipstyler} using natural language prompts.

One of the main limitations of these methods is that they process bitmap images only of a fixed resolution, which is an essential constraint preventing manipulations with high-resolution images. 
Image scaling is not applicable for bitmap images without a decrease in quality. 
Meanwhile, scalability is the feature of vector graphics. 

Prior researches tackle the task of vector graphic processing for fonts~\cite{wang2021deepvecfont} and simple graphics such as icons and emoji~\cite{carlier2020deepsvg,reddy2021im2vec}, and sketch-like image generation~\cite{frans2021clipdraw,schaldenbrand2022styleclipdraw} using natural language prompts. 
The study~\cite{efimova2022conditional} suggests an approach for generating vector images consisting of multiple B\'ezier curves conditioned by a music track and its emotion.

To the best of our knowledge, the only work that is relevant to NST for vector graphics is the DiffVG method proposed in~\cite{li2020differentiable}. However, the authors do not address the NST problem directly, but only provide tools applicable to it. We can thus conclude that the field of vector NST remains majorly untouched.


Two paths exist that lead to styled vector images: (1) rasterize vector input, apply bitmap style transfer algorithms, and then vectorize the result and (2) apply style transfer without directly to the input vector image.

We believe that the second path is preferable due to the following two reasons.
First, VGG~\cite{simonyan2014very} and other backbones that are used for feature extraction are trained on ImageNet, which makes them bitmap-based and unable to classify vector images. Therefore, it is necessary to separately train the network for feature extraction.
Second, the first path has a bottleneck: stylized bitmap images must be converted into vector form, which can be done using software algorithms, which produce various artifacts on the image and produce $10-500$ times more curves. A large number of curves makes the vector images difficult to edit. Vectorization approaches without this disadvantage are DiffVG~\cite{li2020differentiable}, which produces artifacts on the resulting image, and  LIVE~\cite{ma2022towards}, which is a very resource- and time-consuming.
Besides, both DiffVG and LIVE require a predetermined number of curves, which strongly affects the fidelity of the result. These and other shortcomings of raster-then-vectorize stylization are discussed in subsection~\ref{subsec:rtv}.

%
Being motivated by this, we decided to find if it is possible to transfer style of a vector image. 
Our contribution is a novel style transfer method for vector images based on learning how to transform an image via backpropagating contour and perceptual losses through differentiable rasterization transformation, \emph{VectorNST}. Some samples of stylized vector images are presented in Fig.~\ref{fig:res}.


\section{Related Work}
\label{sec:rel}


\subsection{Neural Style Transfer for Raster Graphics}

Gatys~\etal~\cite{gatys2015neural} discovered the possibility to separate representations of content and style obtained using a pre-trained CNN~\cite{krizhevsky2017imagenet}. 
They proposed an NST algorithm combining the content of one image with the style of another. 
It jointly optimizes the loss function responsible for style synthesis and loss for content reconstruction using multiple feature maps from a pre-trained VGG network~\cite{simonyan2014very}. 
The algorithm starts with random noise and changes pixel values with gradient-based optimization to obtain a stylized image. 
While producing high-quality results and flexibility, this method is computationally expensive since it requires many forward and backward passes.

To overcome this shortcoming, Johnson~\etal~\cite{johnson2016perceptual} proposed a feed-forward style transfer network, which synthesizes stylized images in one forward pass; the pre-trained VGG model is used as a loss network.
Its performance is similar to the results of Gatys~\etal, but reduces the inference time. 
However, the algorithm limitation is that one trained style transfer network can only be used for one style.


Dumoulin~\etal~\cite{dumoulin2016learned} tackled this problem by introducing a conditional style transfer network that can handle multiple styles and is based on a conditional instance normalization algorithm. 
Defining a specific style requires only trainable parameters of scaling and shifting.  
Moreover, the latent space of these trainable parameters can be used to interpolate between styles and capture new artistic styles.

To address the problem of high-resolution image generation, Yoo~\etal~\cite{yoo2019photorealistic} proposed an algorithm based on whitening and coloring transforms for the direct change of style representation to match the covariance matrix of content representation. 
Wavelet Corrected Transmission (WCT2) using Haar wavelet pooling and unpooling allows losing less structural information and maintains the statistical properties of VGG feature space during stylization. 
It can stylize a $1024 \times 1024$ resolution image in $4.7$ seconds and obtain a photorealistic result without postprocessing. 

A Transformer-based~\cite{vaswani2017attention} approach, initially proposed for language processing, can be an alternative to the classic CNN-based methods as it has achieved state-of-the-art results in many computer vision tasks.
Park~\etal~\cite{park2019arbitrary} proposed the SANet method using the attention mechanism and the identity loss function, which heavily monitors the preservation of image content. However, such an encoder-transfer-decoder architecture cannot handle long-term dependencies, which leads to various distortions and loss of details in a stylized image.
Using transformers' ability to handle long-range dependencies, Deng~\etal~\cite{deng2022stytr2} introduced a transformer-based style transfer framework StyTr$^2$, which splits content and style images into patches and feeds them into different encoders, and then the transformer decoder stylizes the content sequence according to the style sequence.
However, due to the use of a patch-based mechanism, it is difficult to extract and preserve global and local features in a stylized image.
Zhang~\etal~\cite{zhang2022domain} presented a framework for style transfer and image style representation based on contrastive learning. Furthermore, style representations are learned directly from image features as well as the global distribution of style. The proposed multi-layer style projector with CNN layers taking as input feature maps from fine-tuned VGG19 encodes the image into a set of codes that are proper guidance for the style transfer generator.


\subsection{Vector Graphics}

Vector graphics is the most commonly used for various fonts, illustrations, icons, emblems, logos, and other resolution-independent images.
Vector graphics is usually declared as a set of primitives such as lines, curves, and circles with many geometric and color attributes.

The most common vector image format is SVG, which is an XML markup text file describing geometric shapes that are mathematically defined by control points. SVG supports many tags and attributes, but the most interesting is the $<$path$>$ tag, which can be used to describe a shape using B\'ezier curves. The main advantages of vector graphics are lossless scalability, simplicity, and the memory-efficiency.

Most of the existing methods for neural vector image generation are based on work by Li~\etal~\cite{li2020differentiable}. They introduced the differentiable rasterizer for vector graphics, DiffVG, that allows direct optimization of vector image components such as B\'ezier curves instead of a matrix of pixels.

On the basis of DiffVG, Frans~\etal~\cite{frans2021clipdraw} introduced the CLIPDraw method  that synthesizes vector images conditioned by natural language prompts. CLIPDraw iteratively optimizes a set of RGBA B\'ezier curves through gradient descent optimizing cosine distance between text encoding and image encoding from the pre-trained CLIP model~\cite{radford2021learning}. By adjusting text prompts, the model produces different stylized images, which, however, look like sketches rather than pictures. The model performs worse than generative models in high-resolution image generation tasks.

Model-free method for image vectorization, LIVE~\cite{ma2022towards} is an approach that offers a completely differentiable way to vectorize bitmap images.
Unlike the DiffVG method, which uses random path initialization, LIVE uses an initialization method that determines the best place to add a new path based on the color and size of the component.
Although this approach does not use any deep learning model, it implements an iterative image vectorization algorithm, and vectorization of more complex examples requires a lot of resources and takes a long time.

\section{Method}\label{sec:method}
To develop the style transfer for vector graphics, we use DiffVG to parse vector images and obtain shape parameters: anchor points of vector primitives, shape colors, and line widths. Anchor points are the basis for any vector image, they are used to build curves, which form the figures in the image. Each point is characterized by coordinates $[x,y]$. Also, any curve has a color, it is stored in RGBA format in the interval $[0;1]$, and a thickness, which is a float number.
Unlike, style transfer for raster images where only pixel values change, vector images have $3$ uncorrelated groups of shape parameters listed above, which can be updated. 
Changing the parameters of vector primitives is equivalent to transferring the drawing style for bitmap images. Compare NST approaches: in bitmap domain, to transfer style we can only change the color of the pixels in the particular pattern. In vector domain, the drawing style consists of uncorrelated groups of parameters, which can be updated simultaneously or separately.

\begin{figure}[h]
  \centering
  \includegraphics[width=1.0\columnwidth]{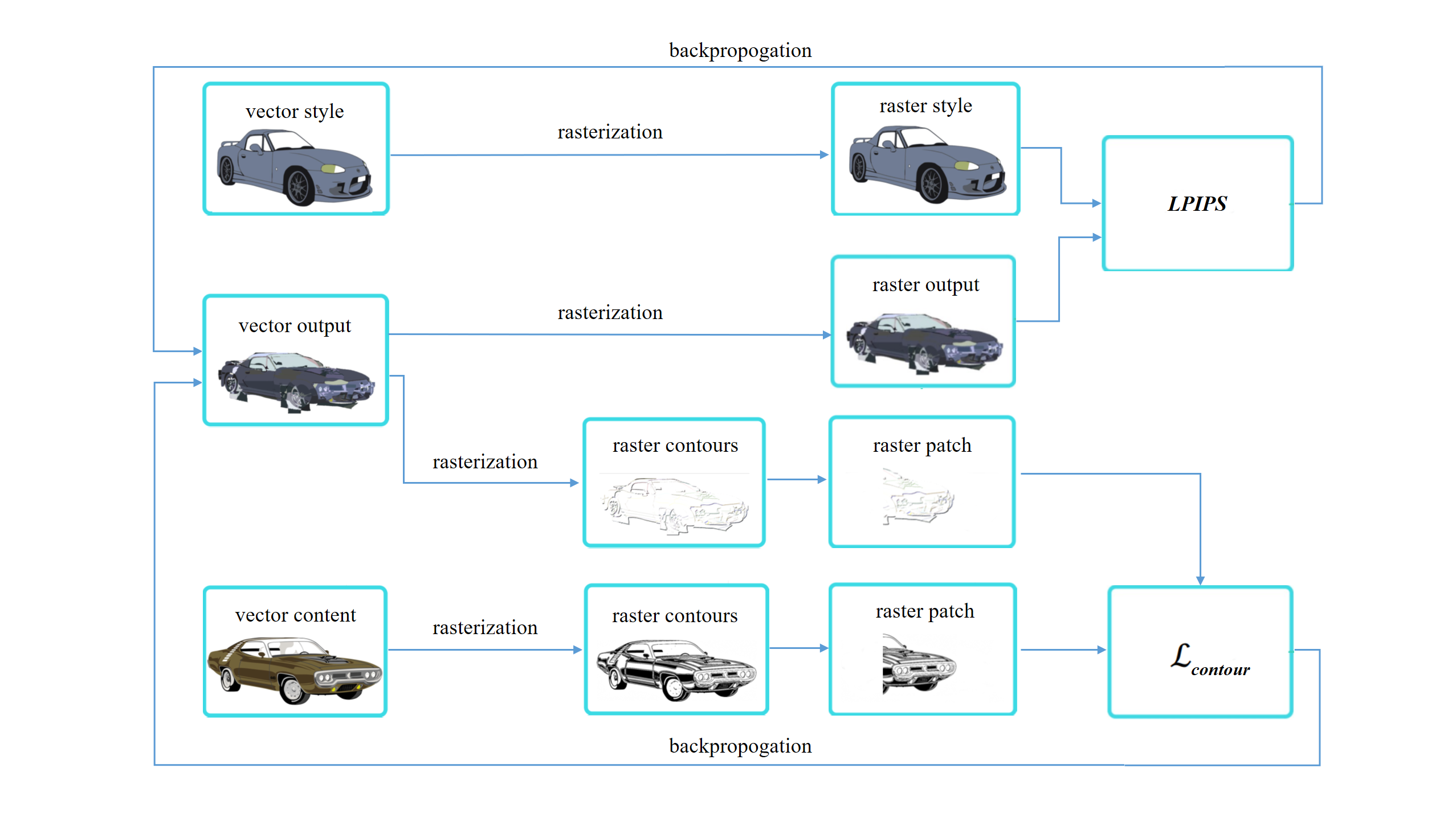}
  \caption{Method overview. We propose a method for real-time style transfer for vector graphics. The optimization consists of two parts. The upper part evaluates the perceptual similarity between the rasterized style image and the output image and aims to convey the style and color of the drawing. The lower part penalizes the differences between the contours of the rasterized content image and the output image to preserve the overall shape of the image.}
  \label{fig:method}
\end{figure}

Based on the above, we aim to develop a model capable of transferring the drawing style of one vector image called a \emph{style image} to another vector image called a \emph{content image} preserving its subject. We do not start the style transfer with a new empty or random vector image, but with a content image, which means that we only change existing shapes and do not create new vector primitives. The method we propose belongs to the iterative optimization methods category, it transfers the style by direct iterative updating shape parameters~\cite{jing2019neural}. 
The number of iterations determines the influence of the style image on the result of the style transfer. To allow evaluation of the resulting vector image, it should be rasterized using DiffVG. After that, the similarity between the current image and the style image is measured by the LPIPS method~\cite{zhang2018unreasonable} and the similarity between the current image and content image is measured with the Contour Loss. Both LPIPS and the Contour Loss are described in detail in subsection~\ref{sec:losses}. The scheme of the method is presented on Fig.~\ref{fig:method}.

\subsection{Differentiable Rasterization with DiffVG}
No algorithm exists to compare the similarity between two vector images. However, it is possible to rasterize them and then evaluate their similarity as bitmap images. In this case, rasterization must be performed by a differentiable operator, which is available in DiffVG, allowing thus to apply backpropagation for image updating.

However, DiffVG had some issues and errors preventing proper image rasterization. We mention two major ones that we encountered while using DiffVG and our suggestions for solving these problems.

First, SVG image parsing is implemented sequentially. If the current visual tag (for example, a linear gradient tag) refers to a tag that has not yet been processed by the method, this causes an error in DiffVG. The solution to this problem is a preliminary topological sorting of visual tags. 

Second, after saving an SVG image with path tags consisting of several subpaths, only the very first subpath remains, and the rest are ignored, which may dramatically change the image. 

In our implementation, we fixed these issues to rasterize images correctly. 
Other limitations and disadvantages of DiffVG are mentioned in Section~\ref{sec:lim}. 

\subsection{Feature Extraction}\label{sec:featextr}
As a feature extractor, we have chosen the standard VGG-19 network pre-trained on ImageNet.
We use the deep embeddings of the $16$ convolutional, $5$ max pooling, and $16$ ReLU activation functions of the $19$-layer VGG network\footnote{
\url{https://pytorch.org/hub/pytorch_vision_vgg/}
}.  
We group these $37$ deep embeddings into several intervals by their indices: $[0, 4), [9, 16), [16, 23), [23, 30), [30, 36)$ (we select features before ReLU) following paper~\cite{zhang2018unreasonable}. 
We did not take deep embeddings with indices $4$ to $8$ because otherwise, it leads to marred contours in the final image.

\subsection{Losses}\label{sec:losses}
Gatys~\etal~\cite{gatys2015neural} proposed to calculate the style loss based on a Gram matrix, which is effective at representing wide varieties of both natural and non-natural textures. The style loss was designed to capture global statistics but it tosses spatial arrangements, which leads to unsatisfying results for modeling shape parameters and obtaining indecent results for vector images. 
On contrary, loss evaluation can be done based on the perceptual distance between images. This can be a solution for our task because perceptual losses eliminate the aforementioned drawbacks of the basic method for vector graphics. We introduce our complete loss function:
\[ \mathcal{L} = LPIPS(x,y) + \lambda \cdot {\mathcal{L}_{contour}}(x,z) \, ,\]
where $LPIPS$ is the perceptual loss we discuss in detail in the next subsection, ${\mathcal{L}_{contour}}$ is the regularization on contours we discuss in subsection~\ref{sub:contour}.


\textbf{Learned Perceptual Image Patch Similarity (LPIPS) metric for vector graphics}
LPIPS~\cite{zhang2018unreasonable} has been used for many computer vision tasks, for example, image restoration and super-resolution. 
In E-LPIPS~\cite{kettunen2019lpips}, authors proposed to use random transformations before calculating the perceptual similarity between images. We found that most of these transformations lead to poorer results for vector images. 
We could modify LPIPS for NST for vector graphics.
After conducting experiments, we noticed that only this color scale transformation improves the quality of style transfer:
\[ color \, scale = random \; uniform(0.0, 1.0)\]
It results in more pleasing colors and smoother contours in the output image.

We use the $L_2$ term to normalize the feature dimension in all pixels and layers to unit length as it is more stable and computationally effective.
In the original paper~\cite{zhang2018unreasonable}, the authors took the sum of $L_2$ distances between the image activation maps, but we use the mean of $L_2$ distances to avoid a high range that can cause artifacts in the output image.

Obtaining the spatial average is important because shapes are often related to each other. The lack of focus on specific, salient regions may lead to the appearance of strange shapes and distortions. Therefore, we included it in our version of LPIPS:

\[LPIPS(x,y) = \sum_{l} \frac{1}{H_l W_l} \frac{1}{h w} \left\|w_l \odot (\hat{x}^l_{h w} - \hat{y}^l_{h w})\right\|_2^2\]

The equation illustrates how the distance between style and output images is obtained: we extract features from $L$ layers, apply the random transformation, normalize them in the channel dimension, and obtain $\hat{x}^l_{h w}$ and $\hat{y}^l_{h w}$. Then, we scale the activation maps channel-wise by vector $w^l$ and compute the mean of $L_2$ distances. Finally, we average spatially and sum channel-wise. 

\textbf{Contour Loss.}\label{sub:contour}
DiffVG is used to obtain contours of content and current images. We parse them and change the fill and stroke colors of shapes to black and white, respectively. With these vector primitives, we obtain new raster contour images with DiffVG. Then, we crop random patches from both images. We attempted to compute the difference between patches of various size and found that the size of $(W/4,H/4)$ is the most appropriate. After that, we calculated $L_1$ term:
\[ {\mathcal{L}_{contour}}(x,z) = \frac{1}{n} \sum_{i=1}^{n} \mid x - z \mid .\]
It forces the input image to respect the target image since the $L_1$ loss penalizes the distance between them. As a result, it makes images smoother, and more exact and helps obtain sharp outlines.

\section{Experiments and Results}

In this section, we investigate the behavior of our method and compare it with six other NST methods. 
We try various loss functions such as contour loss and explore with a subset of VGG layers is the most efficient for the extracting style of the vector image.

\textbf{Experiment setup}.
We used the Adam optimizer for each of $3$ parameter groups used in DiffVG to represent a vector image. 
The learning rate for color parameters and stroke width was $0.01$ and $0.1$ correspondingly. Point learning rate $lr$ was chosen depending on the number of shapes in the image, $n$: if $n \in (0, 300), lr = 0.2;$ $n \in (300, 1000), lr = 0.3$;    $ n \in (1000, 1600), lr = 0.4;$ else $lr = 0.8$. Weight of the contour loss $\lambda = 100$.

\textbf{Methods}.
We included the following methods in the comparison: 
(1) DiffVG, the only existing style transfer for vector images, similar to Gatys~\etal. 
We selected loss weights following the original implementation\footnote{\url{https://github.com/BachiLi/diffvg}}
: $\lambda_{style}=500, \lambda_{content}=1$.
(2) Gatys~\etal, the first and the most widespread method for bitmap images.
(3) SANet, (4) StyTr$^2$, and (5) CAST are three state-of-the-art methods for raster style transfer based on encoder-decoder structure. 
(6) AttentionedDeepPaint (ADP), a method for sketch colorization conditioned by given style image\footnote{\scriptsize{\url{https://github.com/ktaebum/AttentionedDeepPaint}}} based on GANs.

\textbf{Dataset}. 
To assess the quality of the resulting images, we collected a dataset of $500$ vector images, mostly sketchy animals, cars, and landscapes from the FreeSVG website\footnote{\scriptsize{\url{https://freesvg.org}}}.
It contains freely distributed SVG files of various domains with no specific focus. 

\textbf{Metrics}.
Evaluating the results in the field of NST is a sophisticated problem and there is no gold standard by which the best model can be identified. 
No method can determine how accurately the image style was reproduced, because this task is imprecise, and even a human is often unable to give a correct assessment.
Nevertheless, we made attempts to compare the models using style and content losses proposed in the original article by Gatys~\etal. However, using this approach, we encountered difficulties that did not allow us to make a comparison in this way. 
Instead, we evaluated generated images by ourselves, involved assessors for quality estimation, and compared the time of inference.

\subsection{Visual Comparison}

\begin{figure*}[h] 
\begin{center}
\begin{tabular}{
>{\centering\arraybackslash}m{1.5cm}
>{\centering\arraybackslash}m{1.5cm}
>{\centering\arraybackslash}m{1.5cm}
>{\centering\arraybackslash}m{1.5cm}
>{\centering\arraybackslash}m{1.5cm}
>{\centering\arraybackslash}m{1.5cm}
>{\centering\arraybackslash}m{1.5cm}
>{\centering\arraybackslash}m{1.5cm}
>{\centering\arraybackslash}m{1.5cm}
}
\toprule
Content Image&
Style Image&
{\bf Ours} (vector)&
DiffVG (vector)&
Gatys~\etal (raster)&
StyTr$^2$ (raster)&
SANet (raster)&
CAST (raster)&
ADP (raster) \\
\midrule
\includegraphics[width=1.6cm, height=1.5cm]{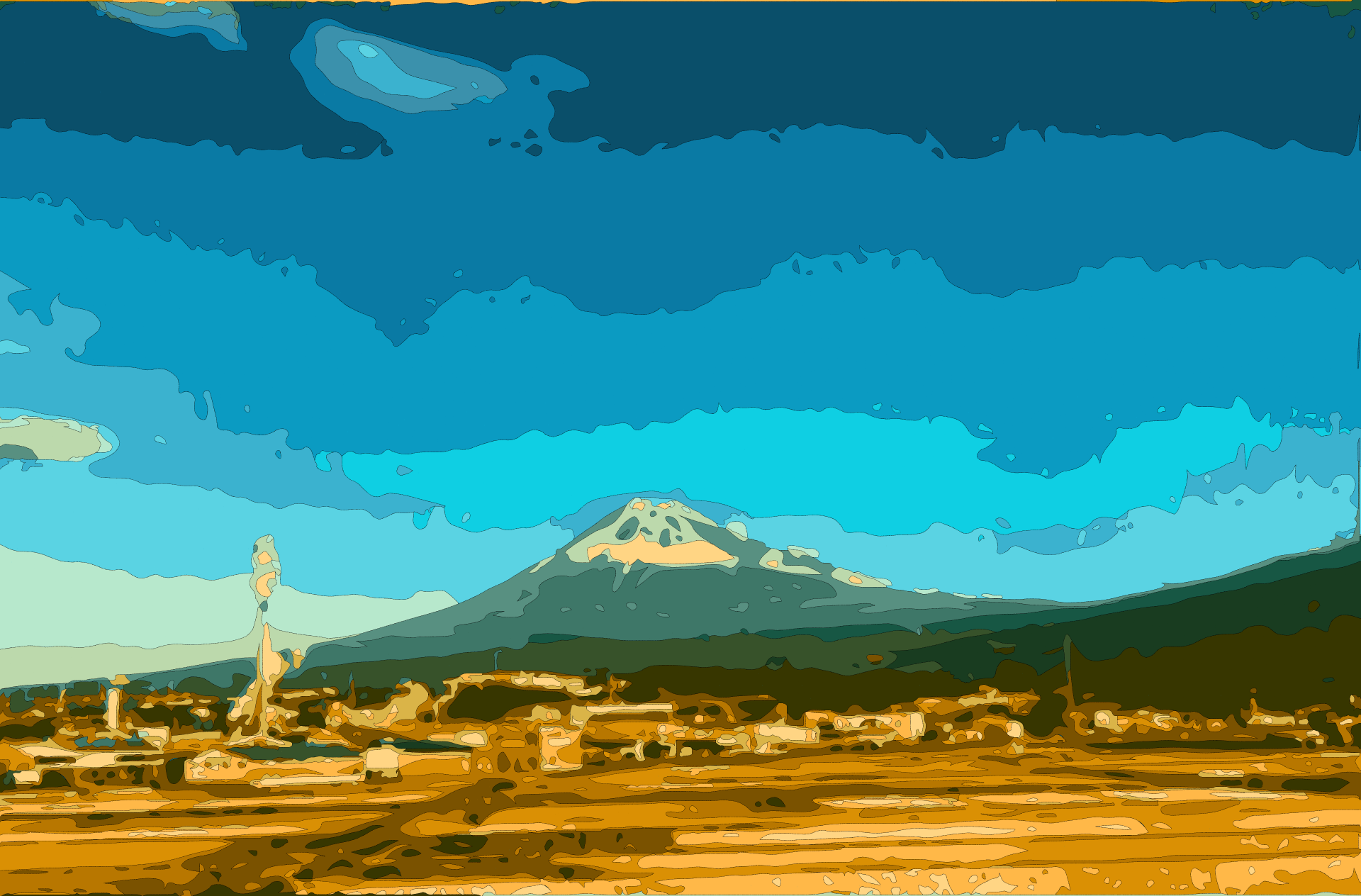} & \includegraphics[width=1.6cm, height=1.5cm]{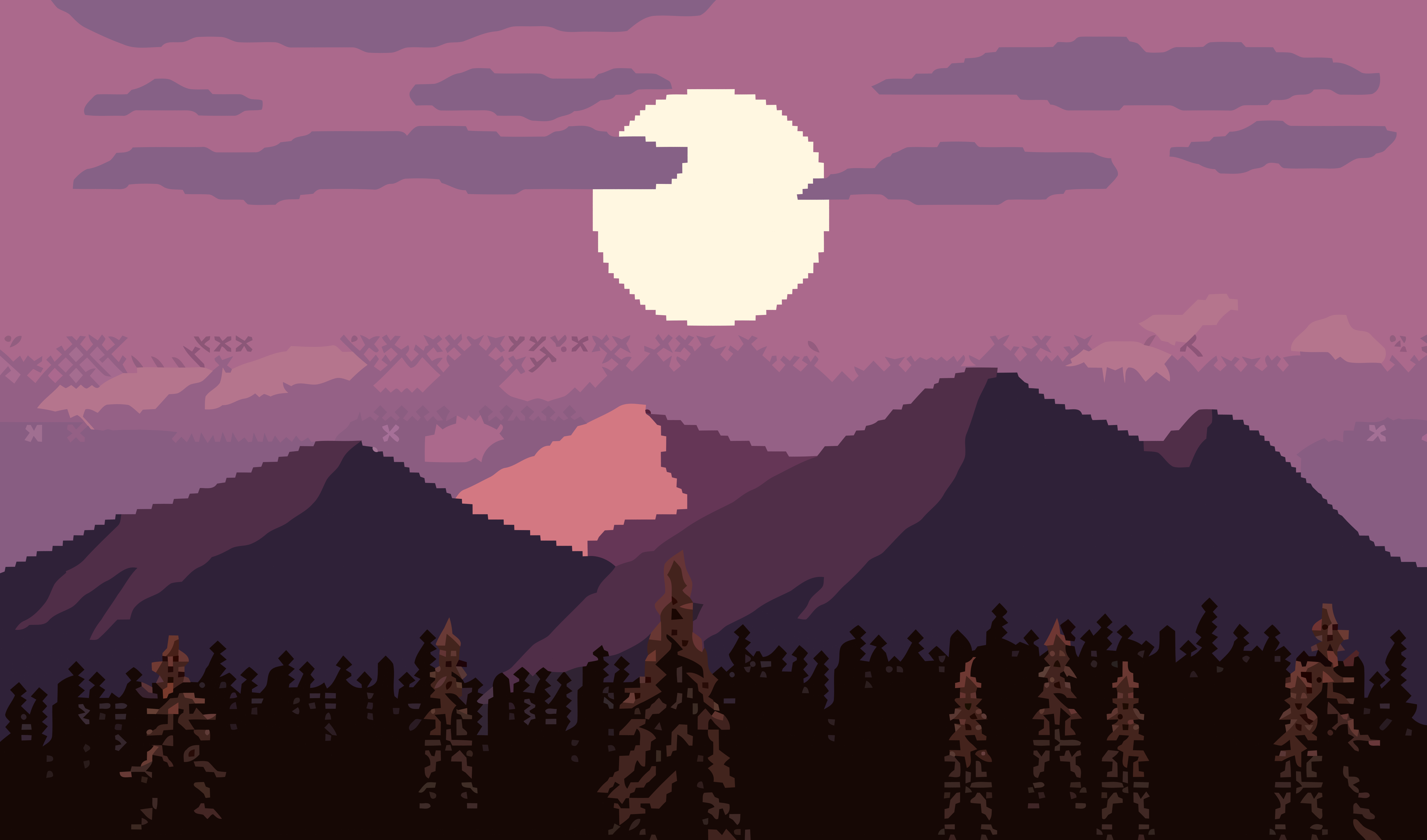} &
\includegraphics[width=1.6cm, height=1.5cm]{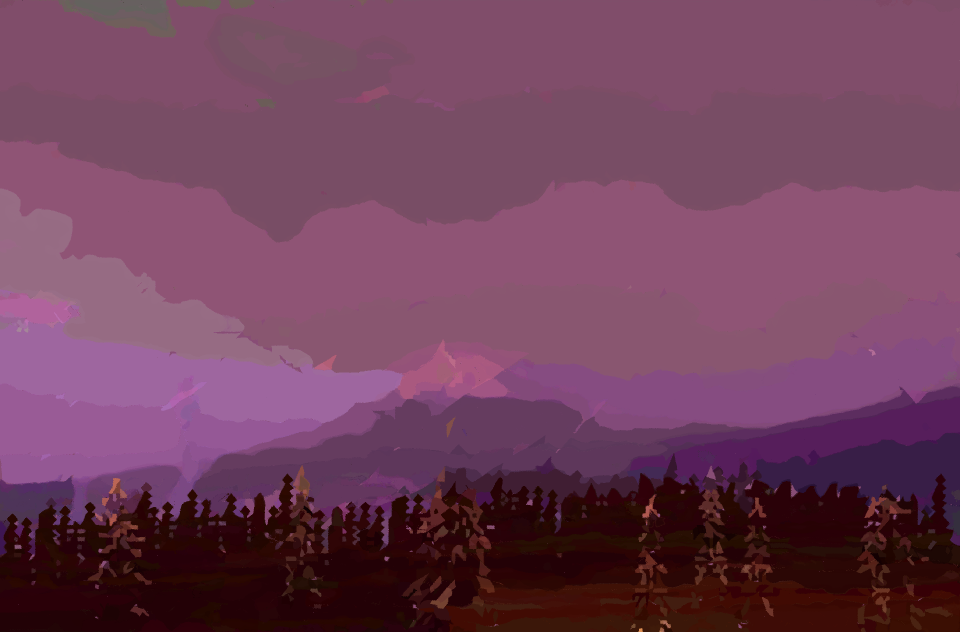} & \includegraphics[width=1.6cm, height=1.5cm]{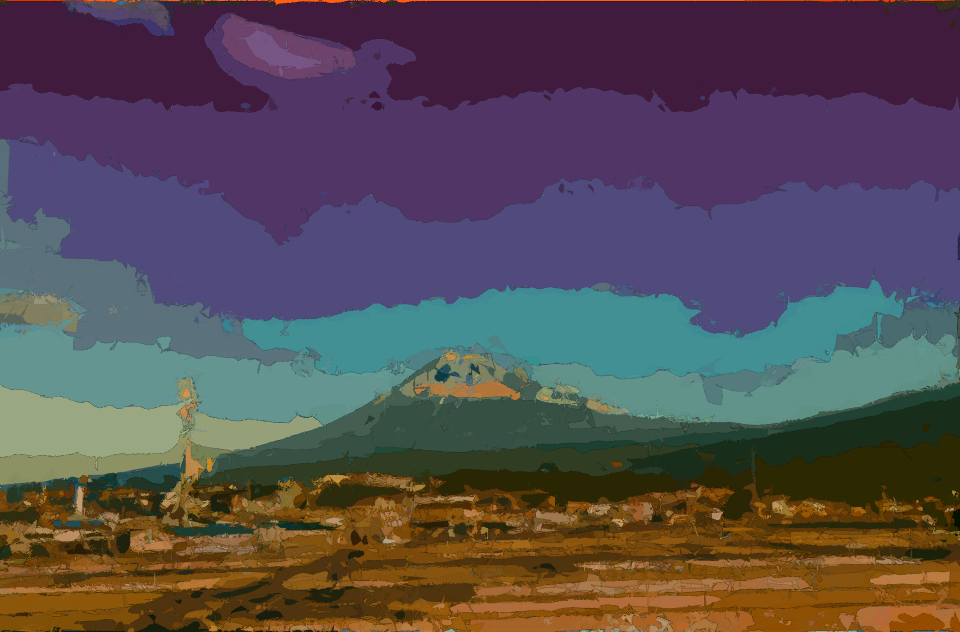} &
\includegraphics[width=1.6cm, height=1.5cm]{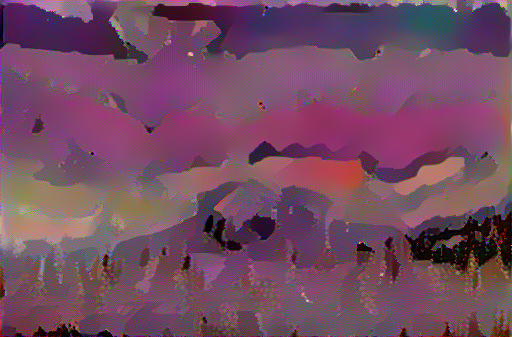} & \includegraphics[width=1.6cm, height=1.5cm]{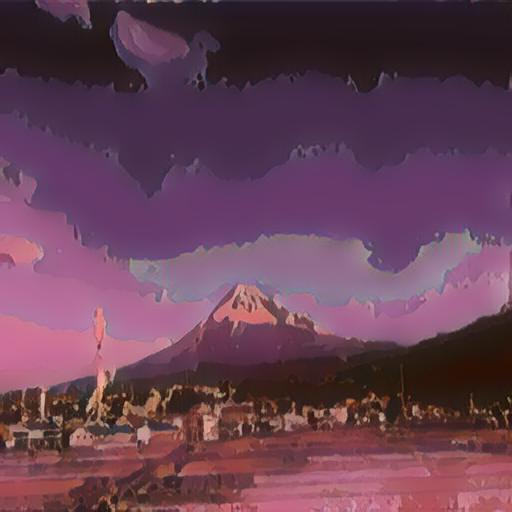} &
\includegraphics[width=1.6cm, height=1.5cm]{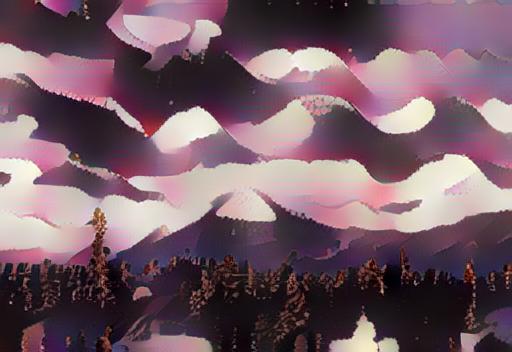} &
\includegraphics[width=1.6cm, height=1.5cm]{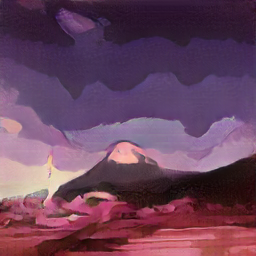} &
\includegraphics[width=1.6cm, height=1.5cm]{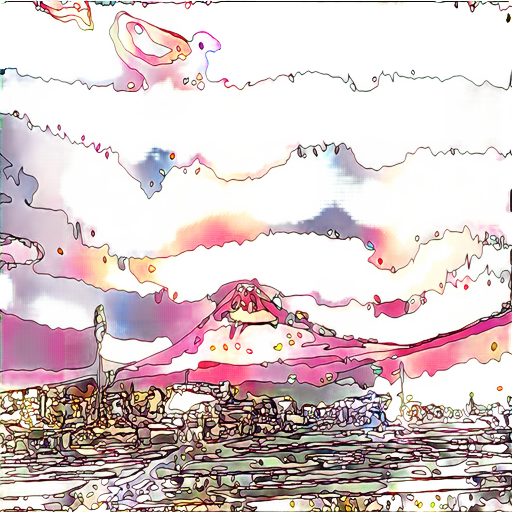}
\\
\includegraphics[width=1.6cm, height=1.5cm]{images/content/scene6.png} & \includegraphics[width=1.6cm, height=1.5cm]{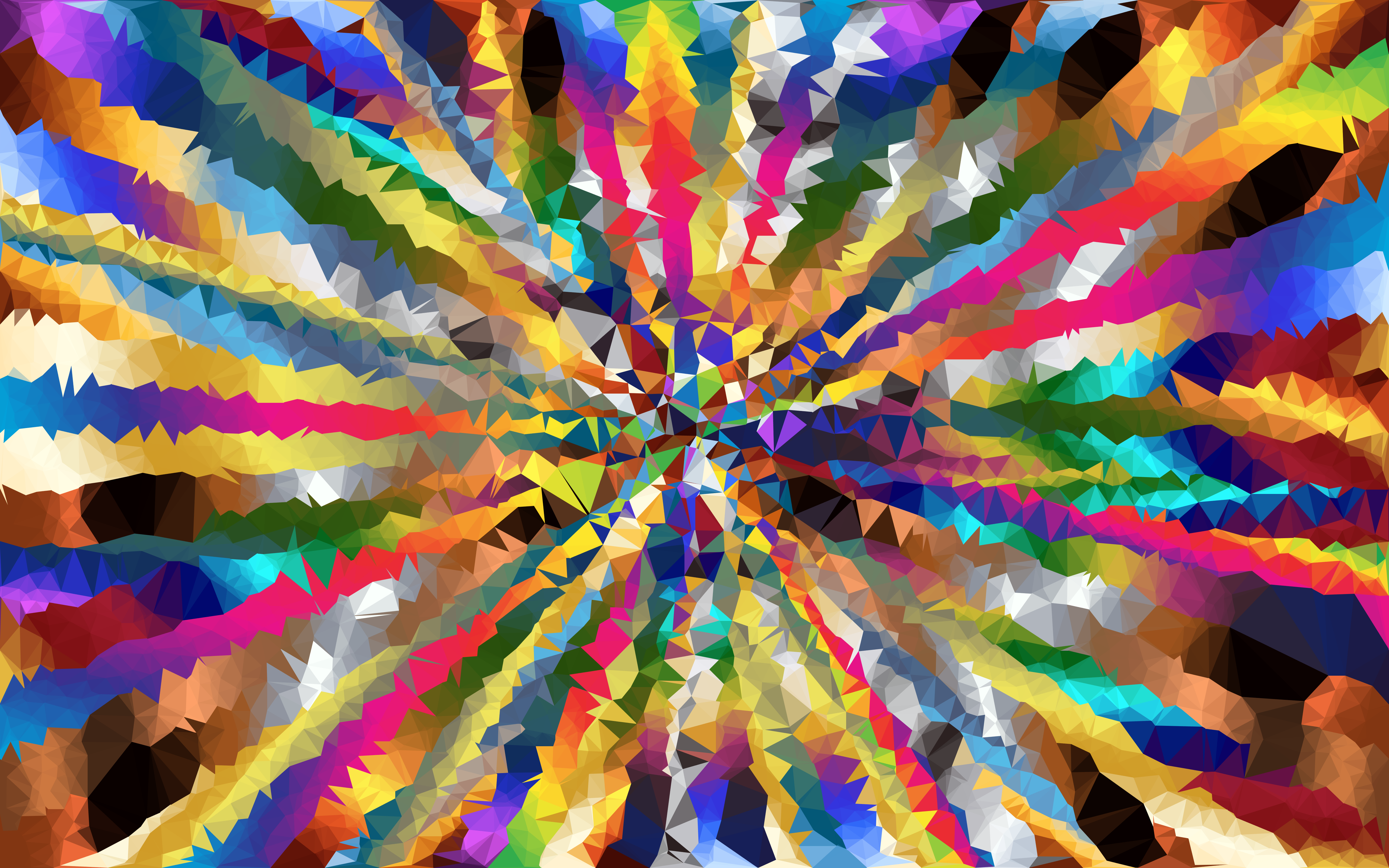} &
\includegraphics[width=1.6cm, height=1.5cm]{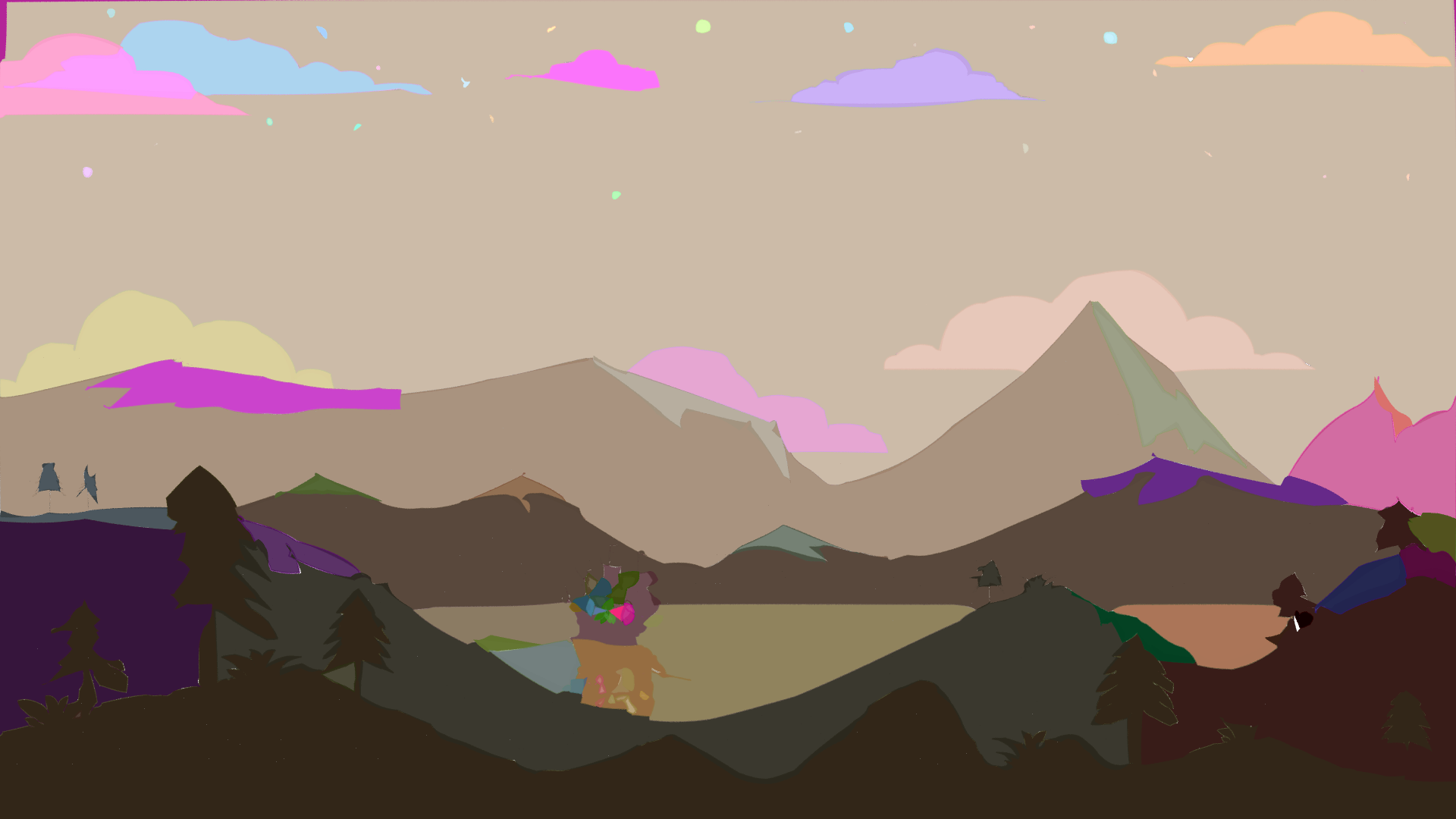} & \includegraphics[width=1.6cm, height=1.5cm]{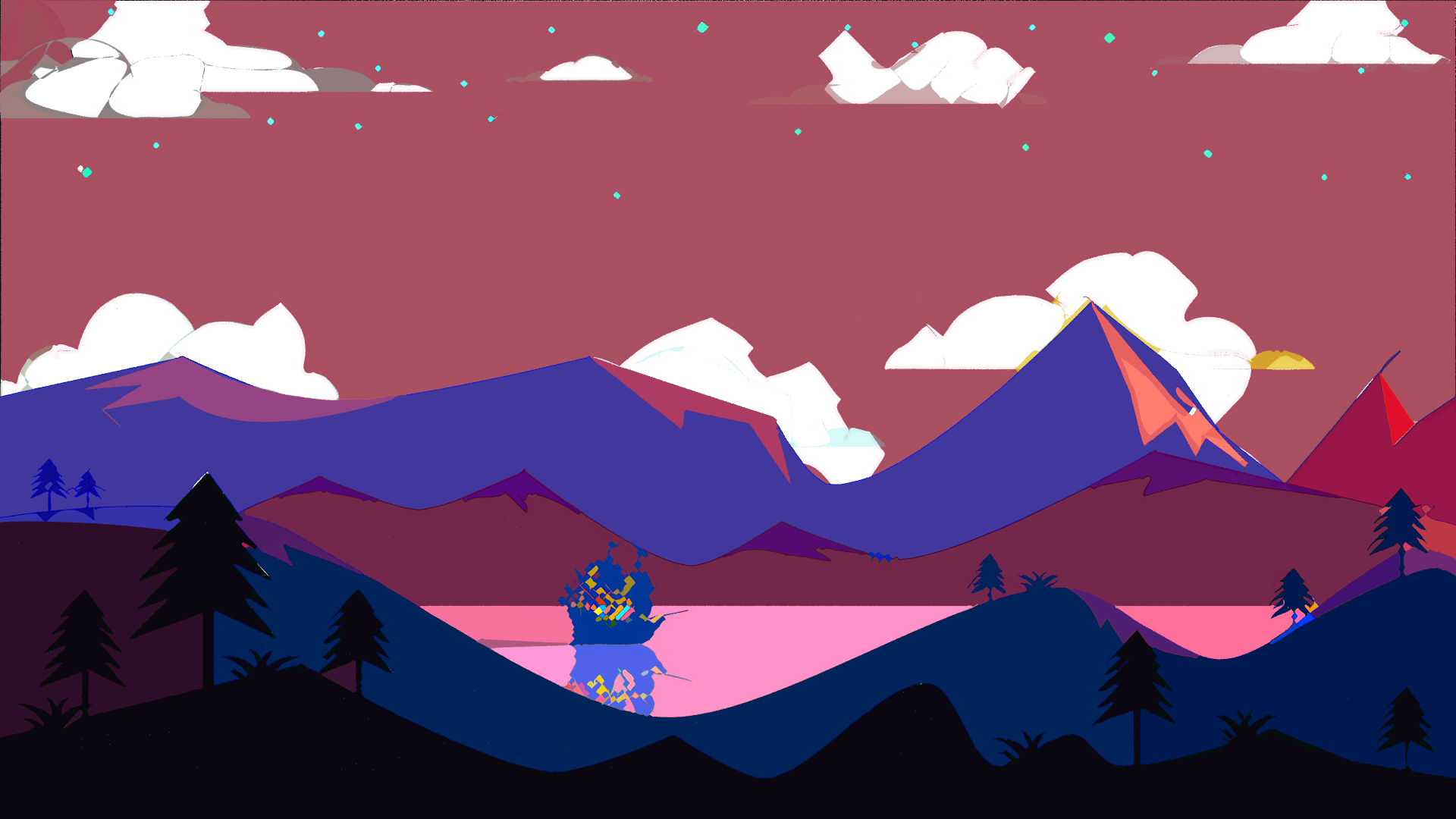} &
\includegraphics[width=1.6cm, height=1.5cm]{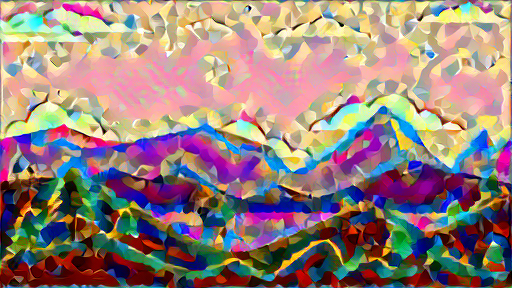} & \includegraphics[width=1.6cm, height=1.5cm]{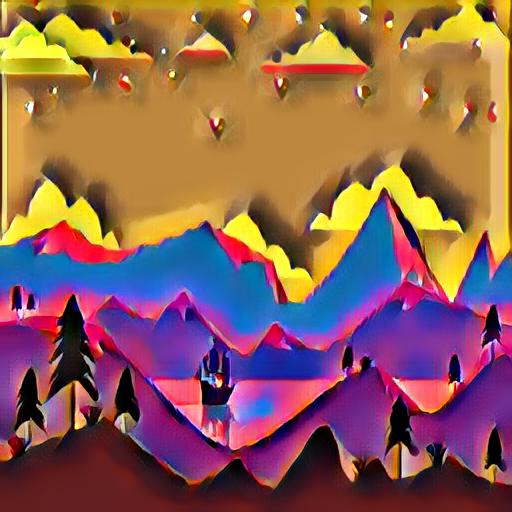} &
\includegraphics[width=1.6cm,
height=1.5cm]{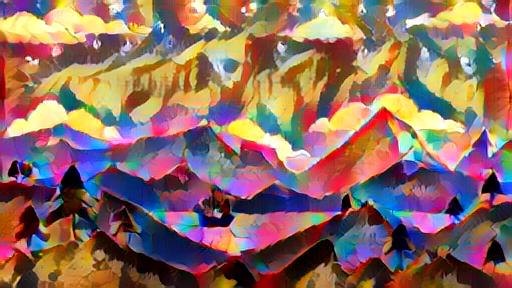} &
\includegraphics[width=1.6cm, height=1.5cm]{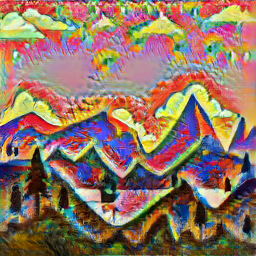} &
\includegraphics[width=1.6cm, height=1.5cm]{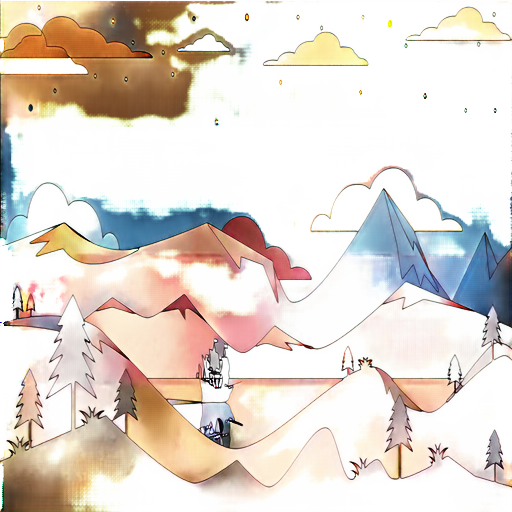}\\
\includegraphics[width=1.3cm, height=1.5cm]{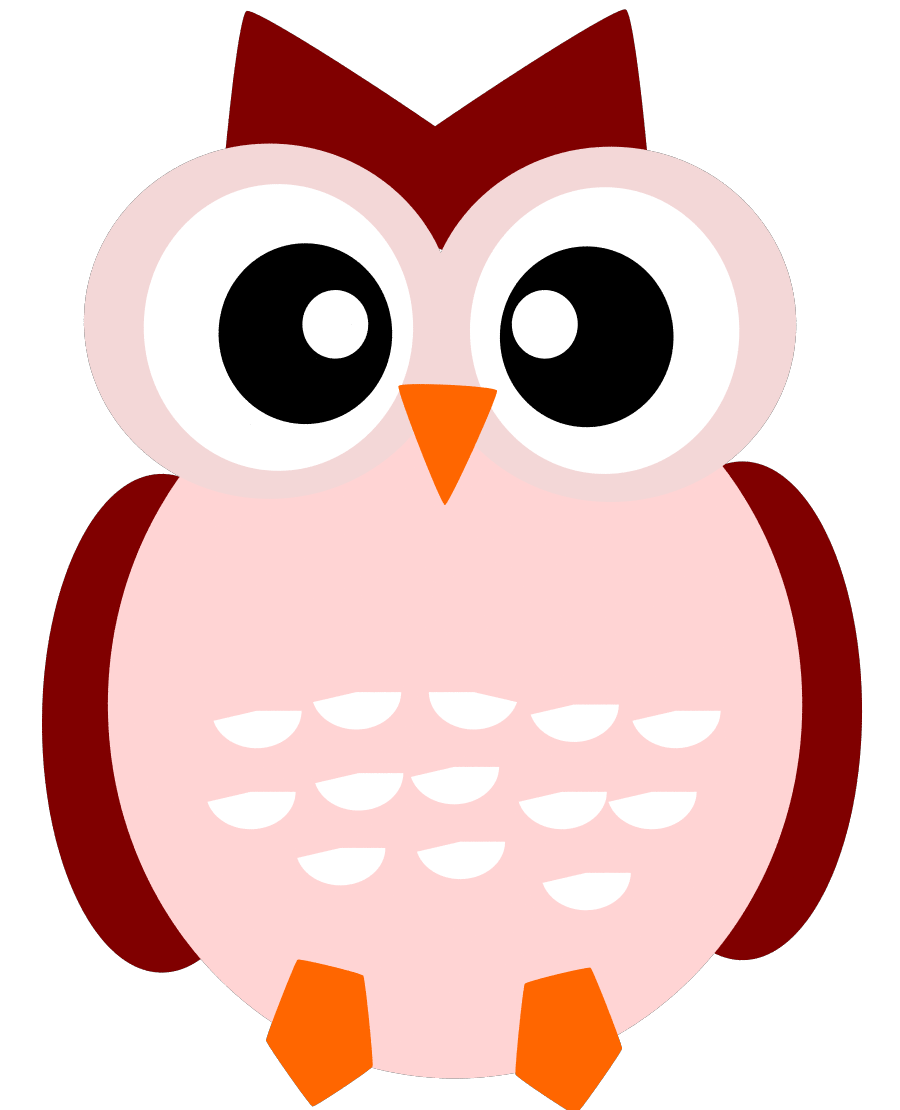} & \includegraphics[width=1.3cm, height=1.5cm]{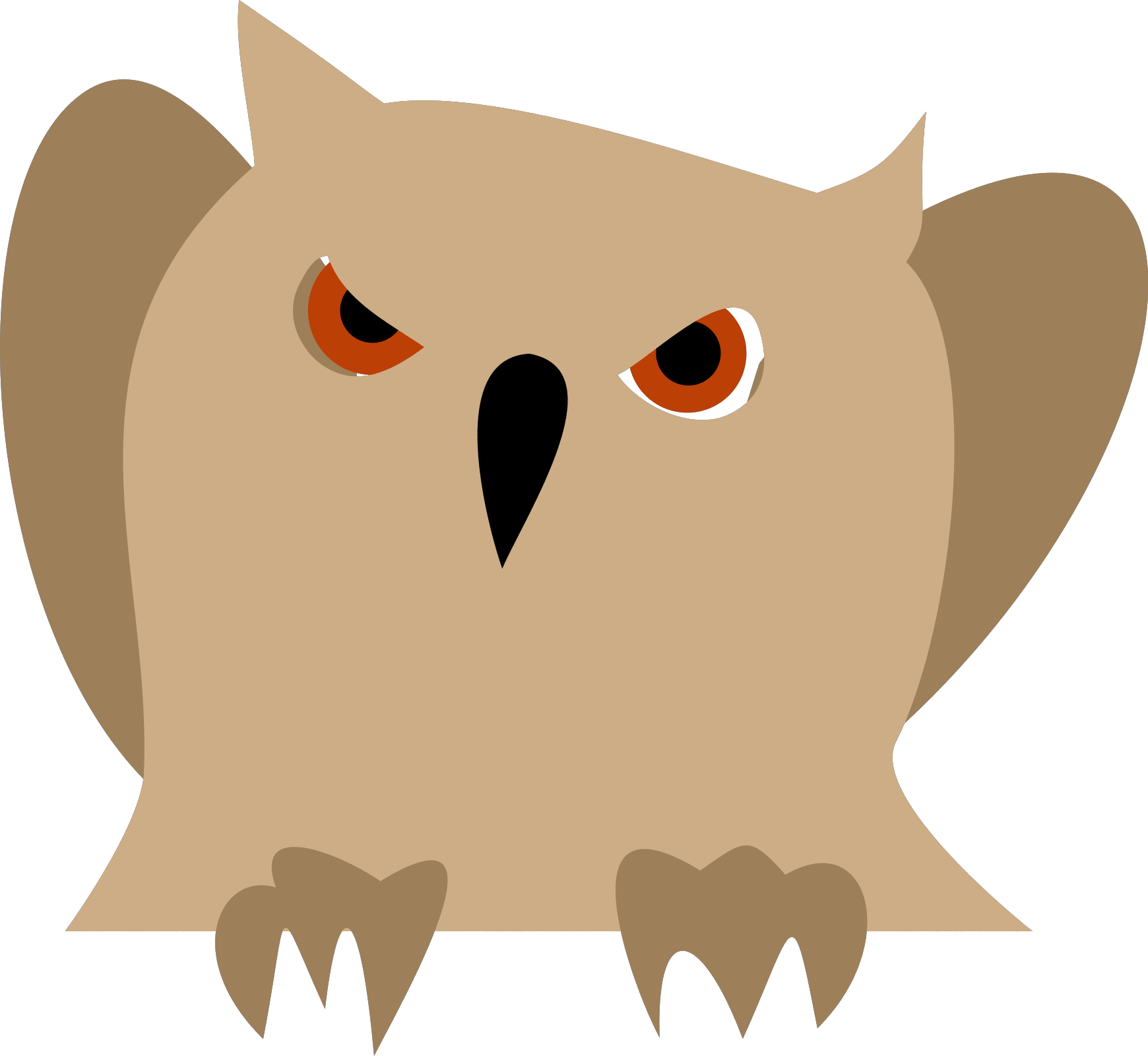} &
\includegraphics[width=1.3cm, height=1.5cm]{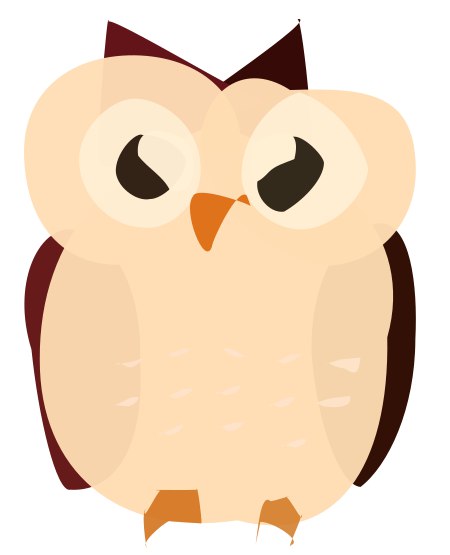} & \includegraphics[width=1.3cm, height=1.5cm]{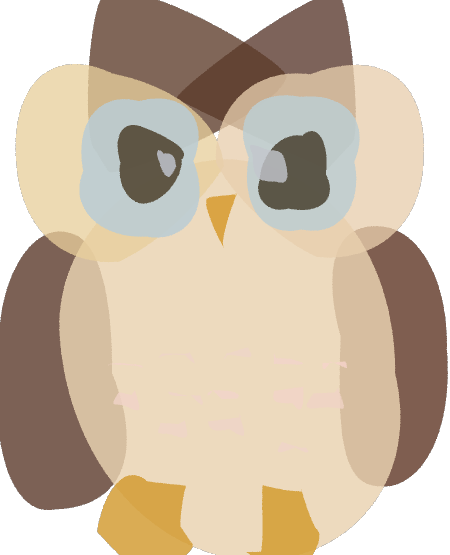} &
\includegraphics[width=1.3cm, height=1.5cm]{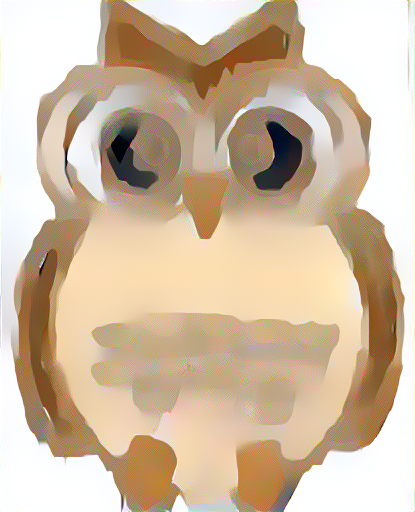} & \includegraphics[width=1.3cm, height=1.5cm]{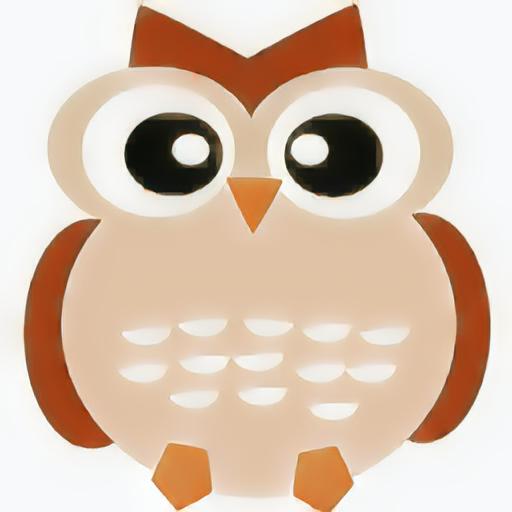} &
\includegraphics[width=1.3cm,
height=1.5cm]{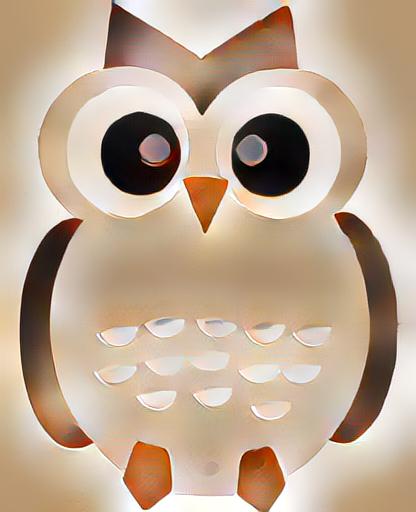} &
\includegraphics[width=1.3cm, height=1.5cm]{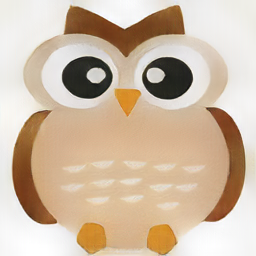} &
\includegraphics[width=1.3cm, height=1.5cm]{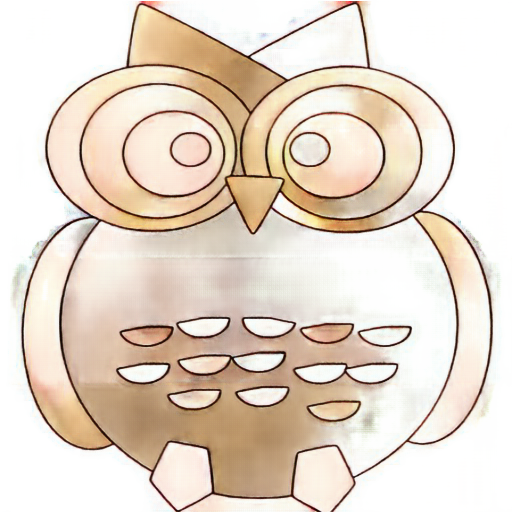}
\\
\includegraphics[width=1.6cm, height=1.2cm]{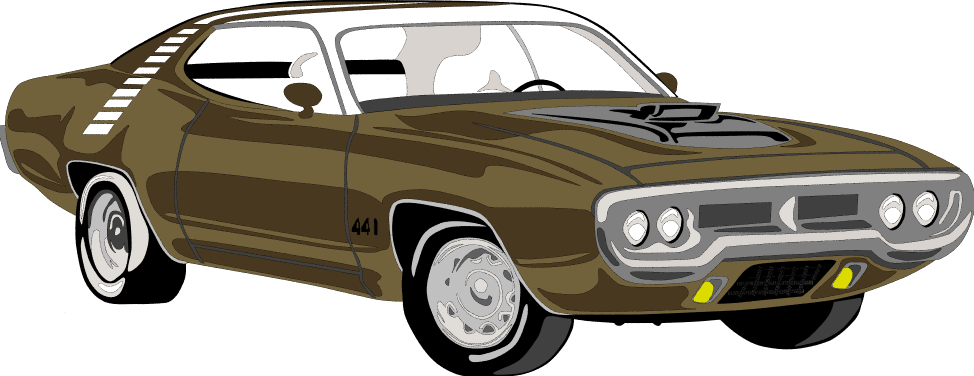} & \includegraphics[width=1.6cm, height=1.2cm]{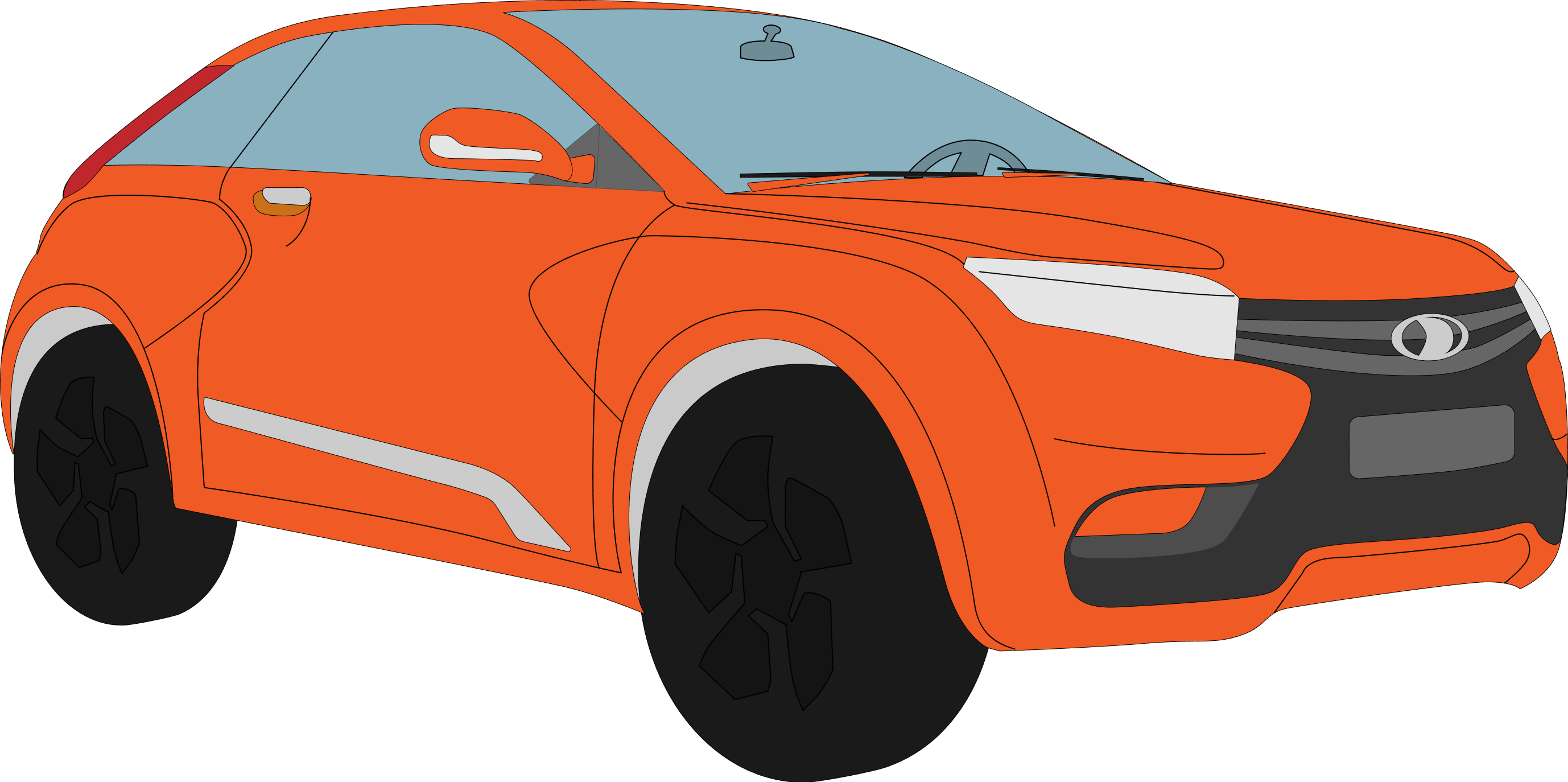} &
\includegraphics[width=1.6cm, height=1.2cm]{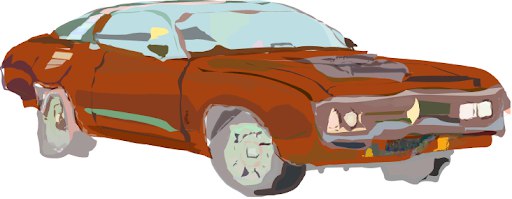} & \includegraphics[width=1.6cm, height=1.2cm]{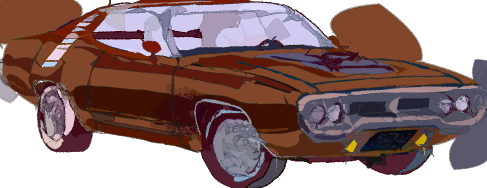} &
\includegraphics[width=1.6cm, height=1.2cm]{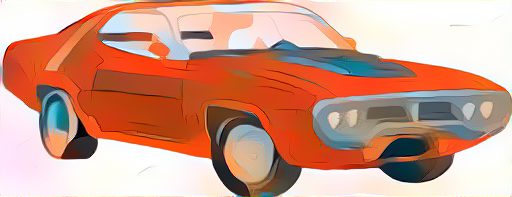} & \includegraphics[width=1.6cm, height=1.2cm]{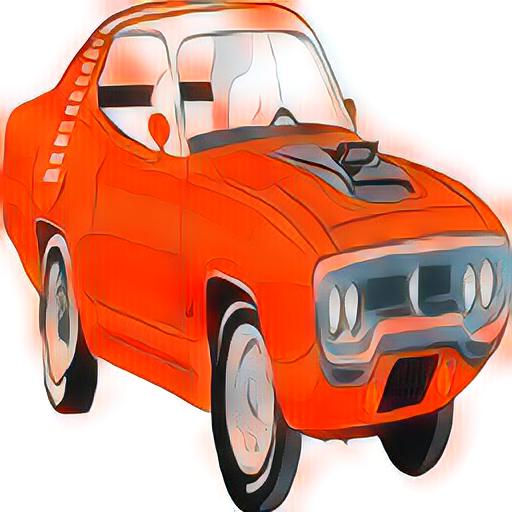} &
\includegraphics[width=1.6cm,
height=1.2cm]{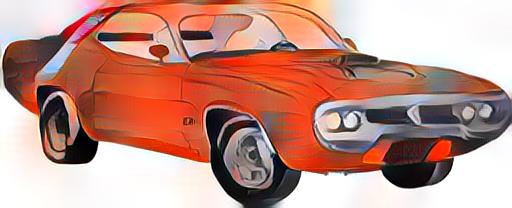} &
\includegraphics[width=1.6cm, height=1.2cm]{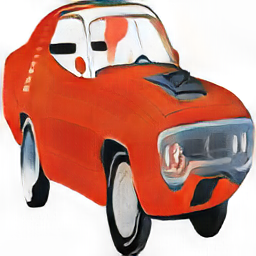} &
\includegraphics[width=1.6cm, height=1.2cm]{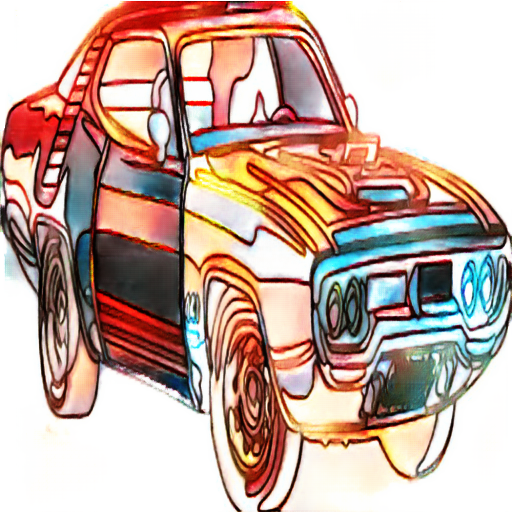}\\
\includegraphics[width=1.3cm, height=1.5cm]{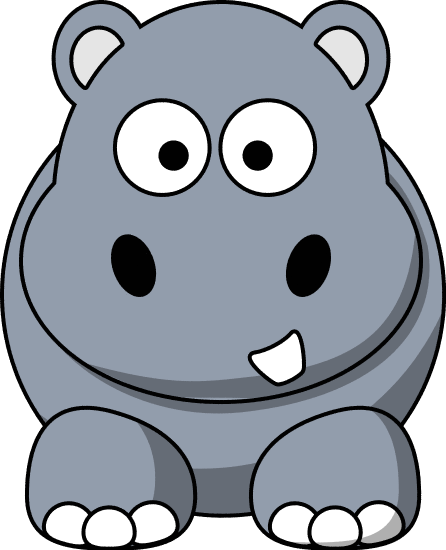} & \includegraphics[width=1.3cm, height=1.5cm]{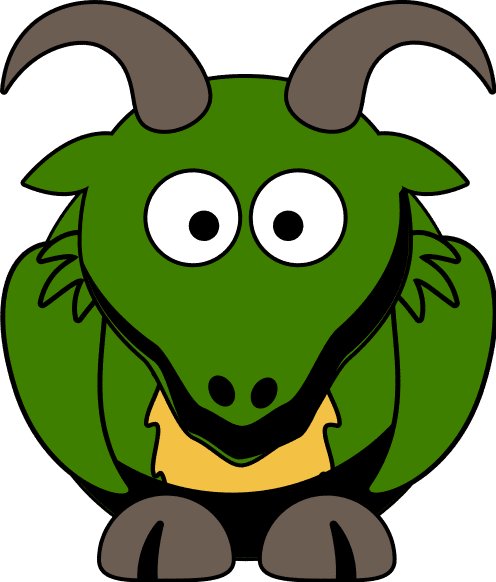} &
\includegraphics[width=1.3cm, height=1.5cm]{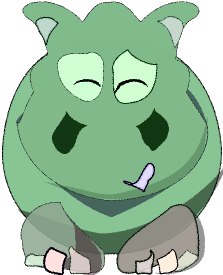} & \includegraphics[width=1.3cm, height=1.5cm]{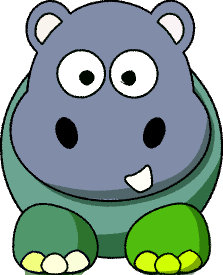} &
\includegraphics[width=1.3cm, height=1.5cm]{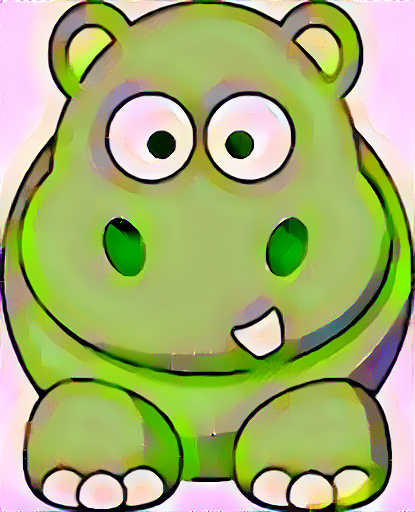} & \includegraphics[width=1.3cm, height=1.5cm]{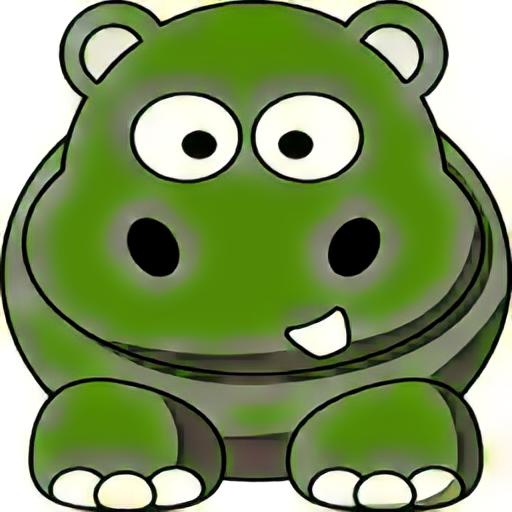} &
\includegraphics[width=1.3cm,
height=1.5cm]{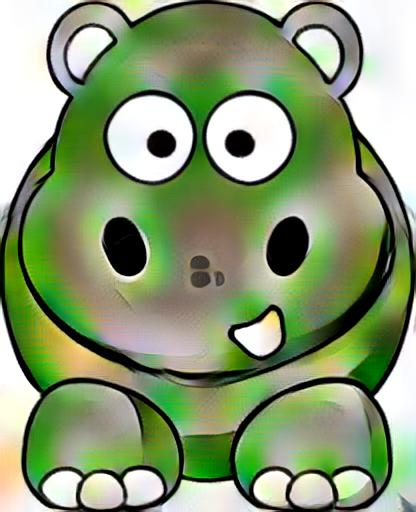} &
\includegraphics[width=1.3cm, height=1.5cm]{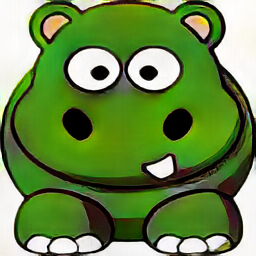} &
\includegraphics[width=1.3cm, height=1.5cm]{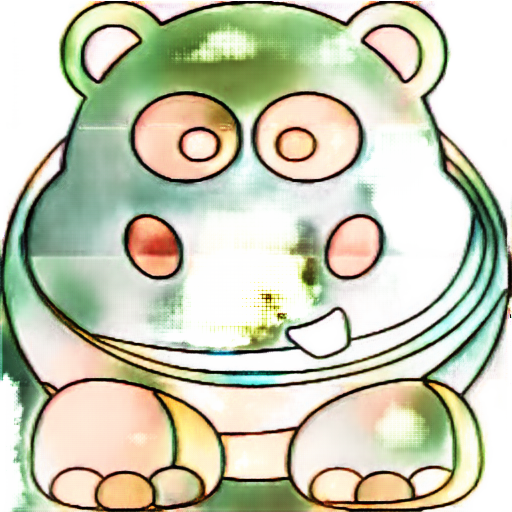}\\
\includegraphics[width=1.6cm, height=1.5cm]{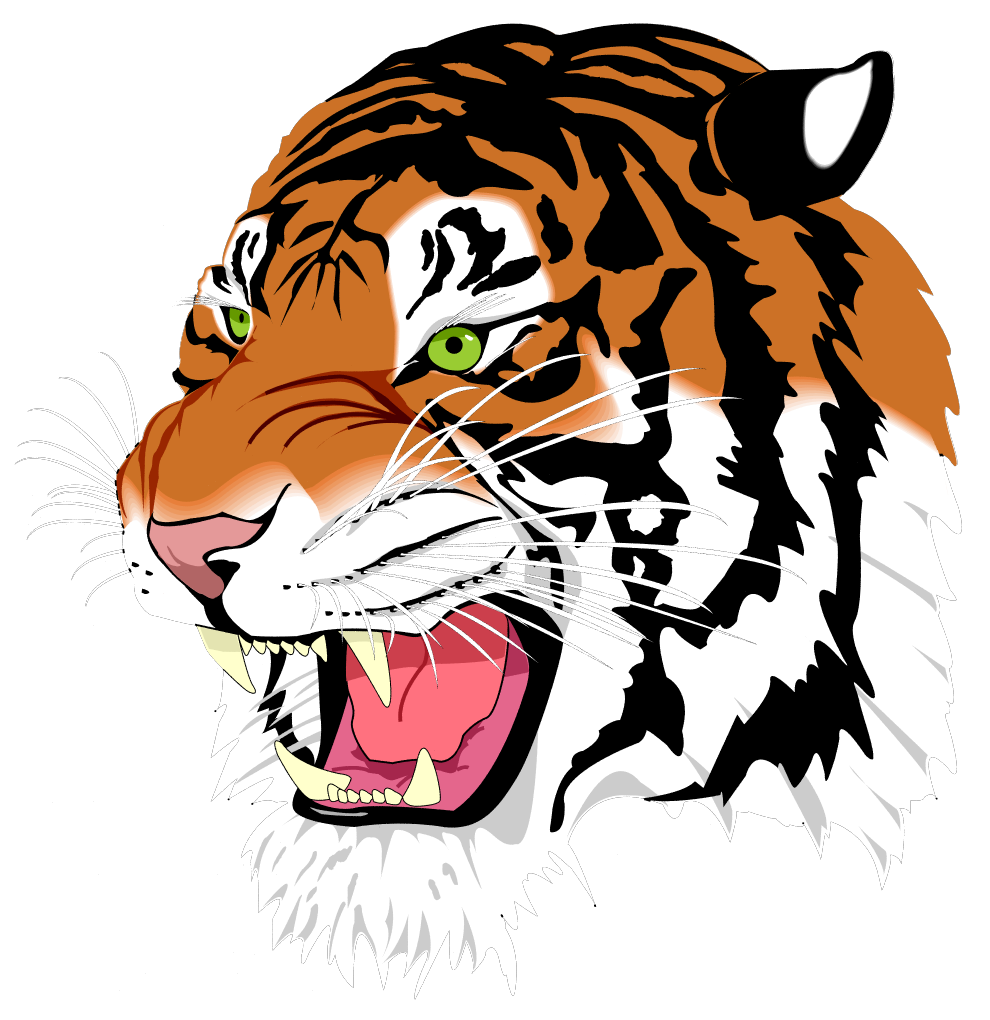} & \includegraphics[width=1.6cm, height=1.5cm]{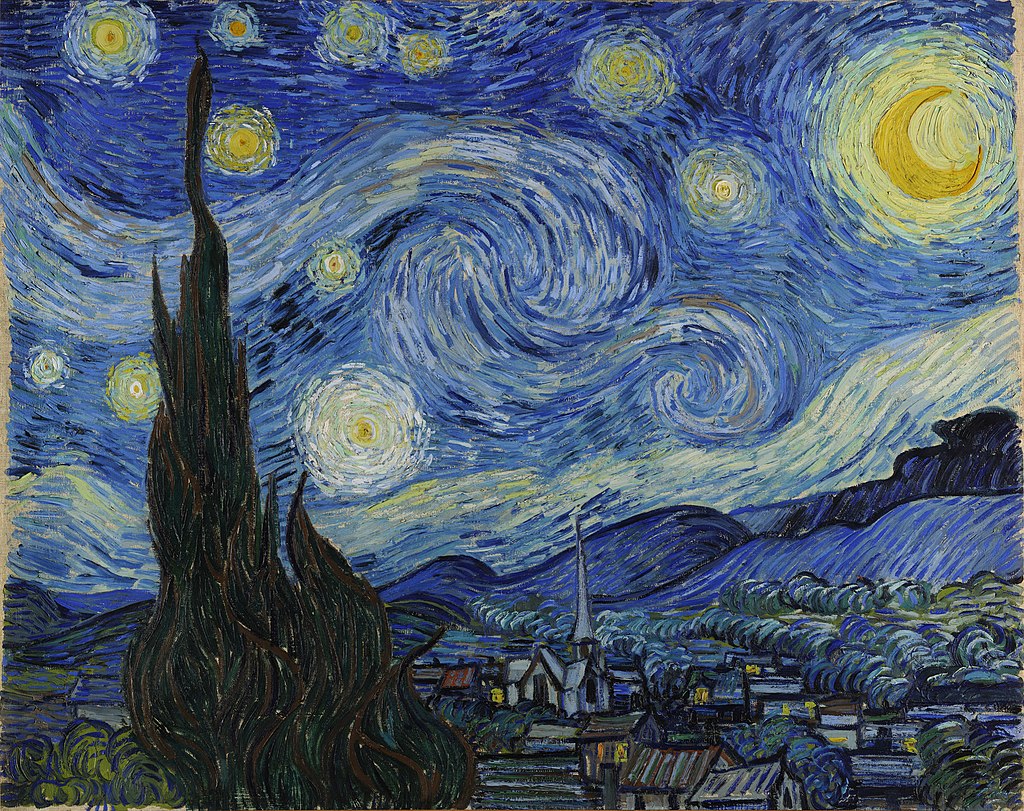} &
\includegraphics[width=1.6cm, height=1.5cm]{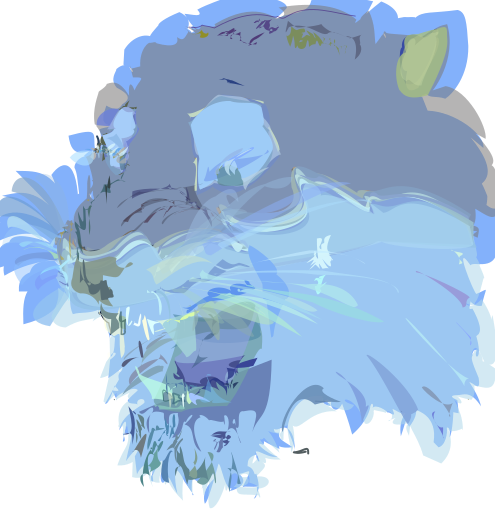} & \includegraphics[width=1.6cm, height=1.5cm]{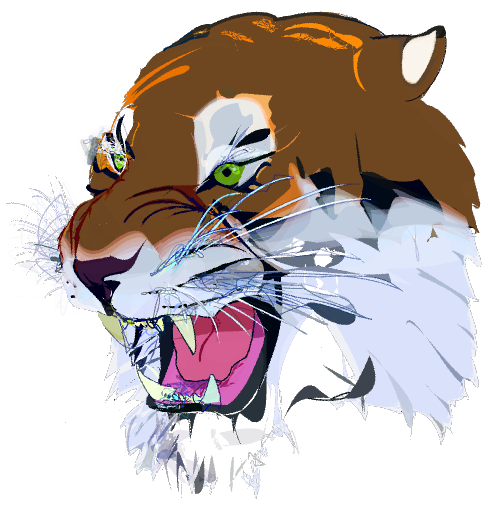} &
\includegraphics[width=1.6cm, height=1.5cm]{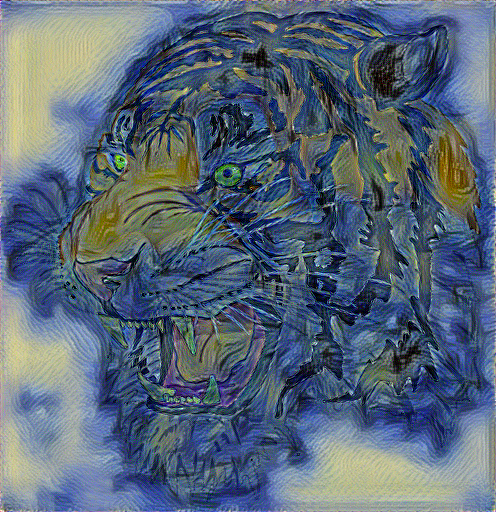} & \includegraphics[width=1.6cm, height=1.5cm]{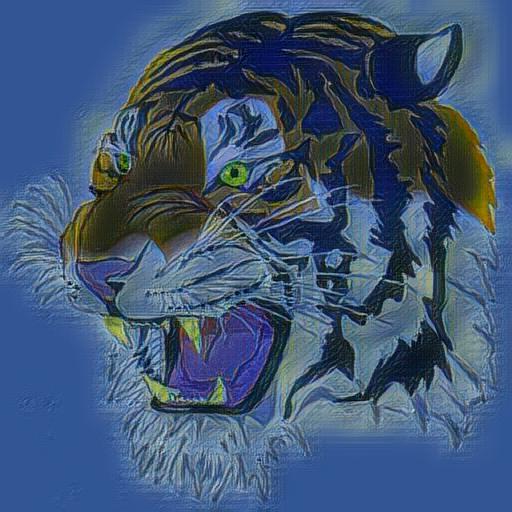} &
\includegraphics[width=1.6cm,
height=1.5cm]{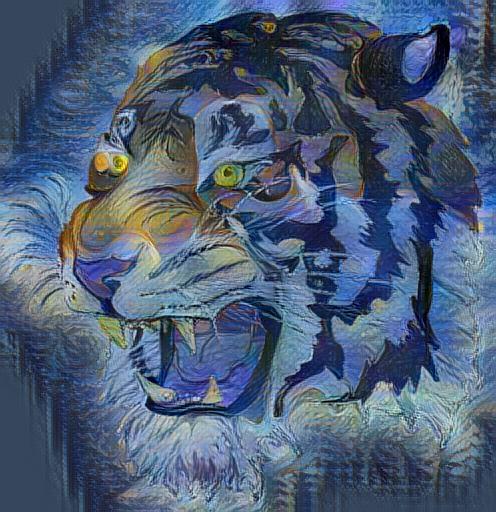} &
\includegraphics[width=1.6cm, height=1.5cm]{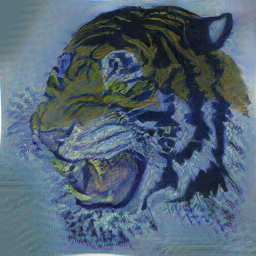} &
\includegraphics[width=1.6cm, height=1.5cm]{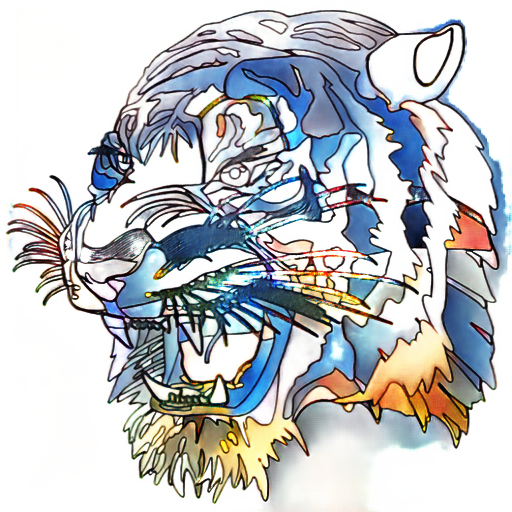}\\
\includegraphics[width=1.6cm, height=1.5cm]{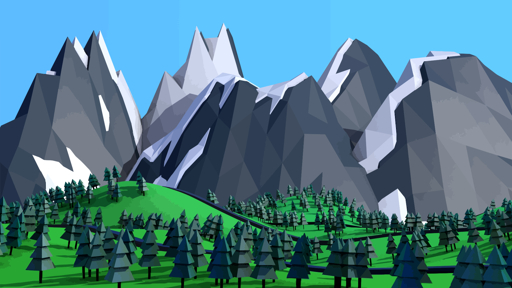} & \includegraphics[width=1.6cm, height=1.5cm]{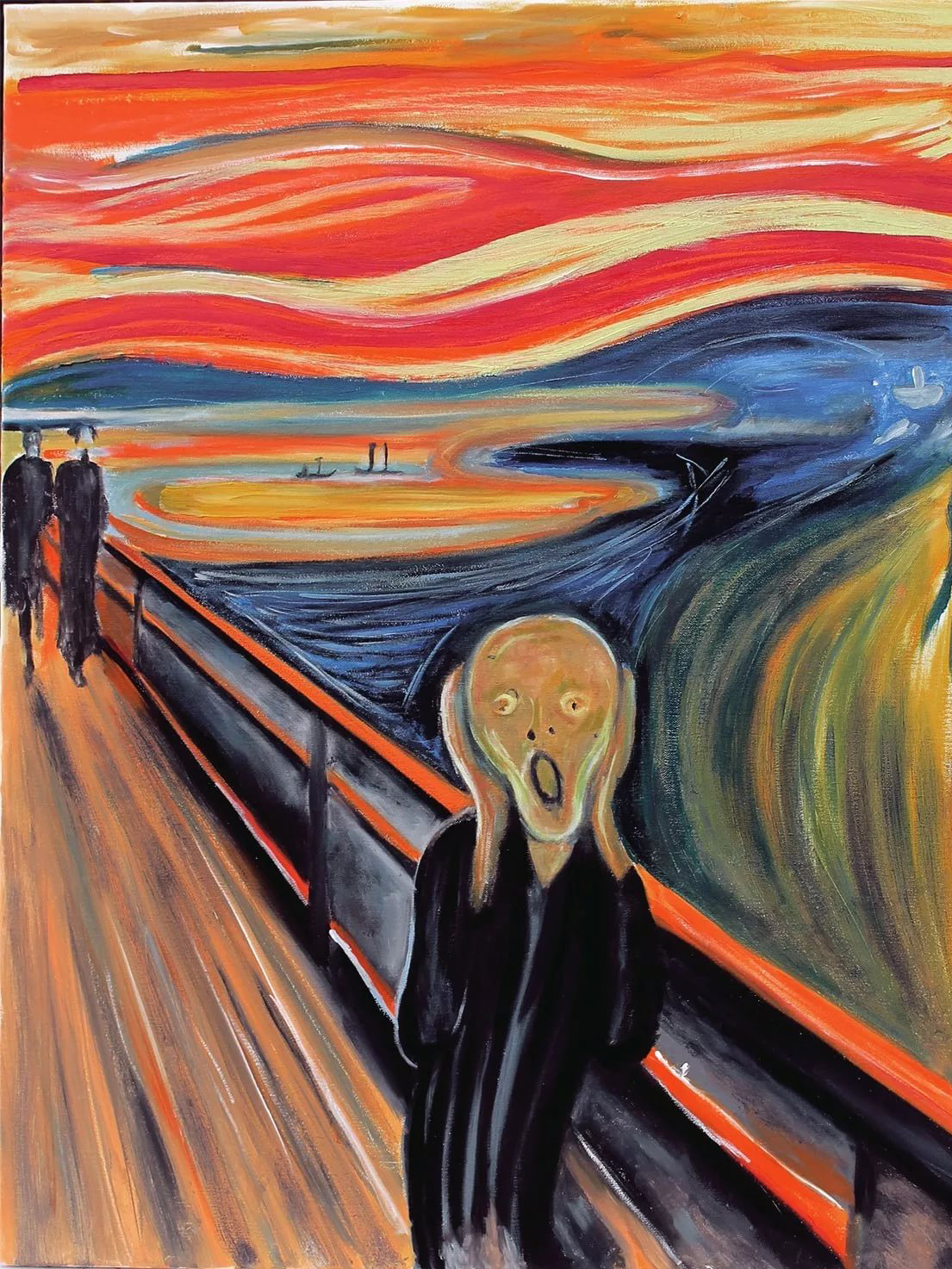} &
\includegraphics[width=1.6cm, height=1.5cm]{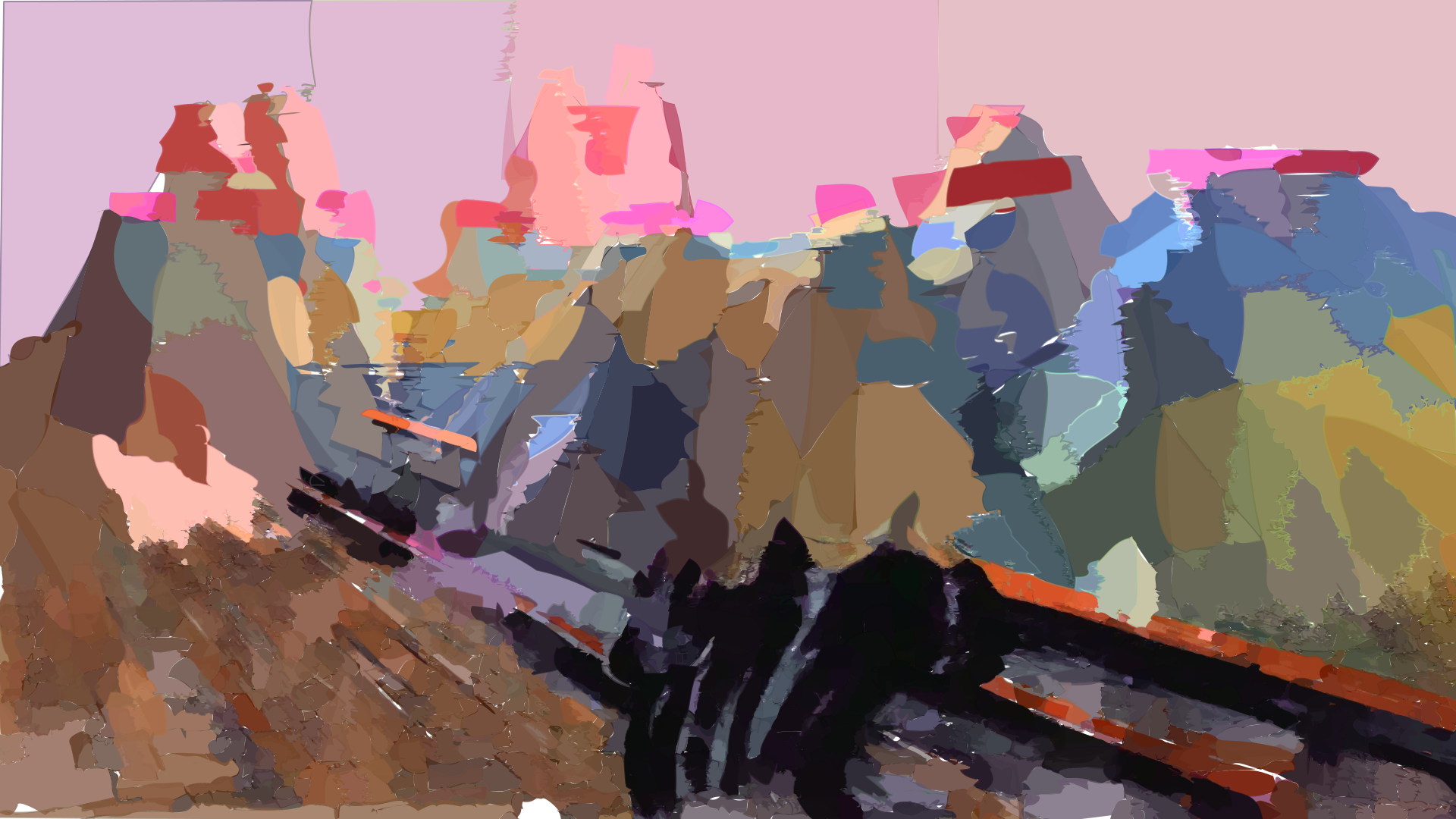} & \includegraphics[width=1.6cm, height=1.5cm]{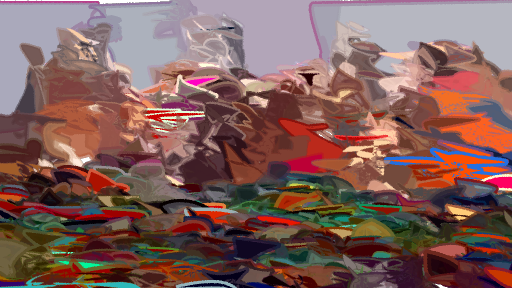} &
\includegraphics[width=1.6cm, height=1.5cm]{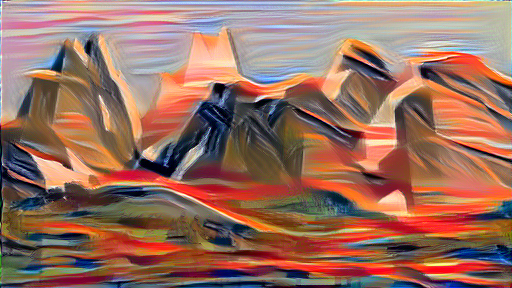} & \includegraphics[width=1.6cm, height=1.5cm]{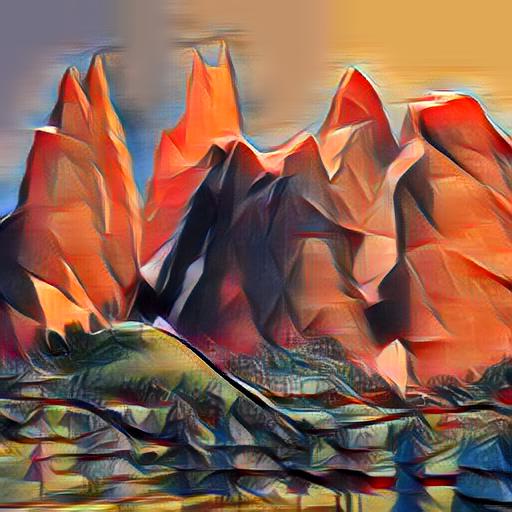} &
\includegraphics[width=1.6cm,
height=1.5cm]{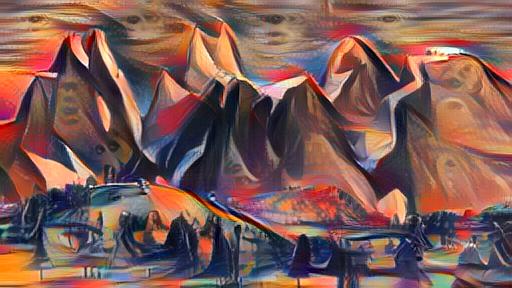} &
\includegraphics[width=1.6cm, height=1.5cm]{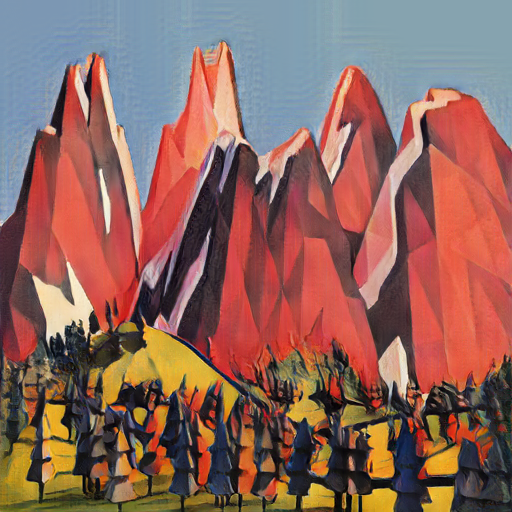} &
\includegraphics[width=1.6cm, height=1.5cm]{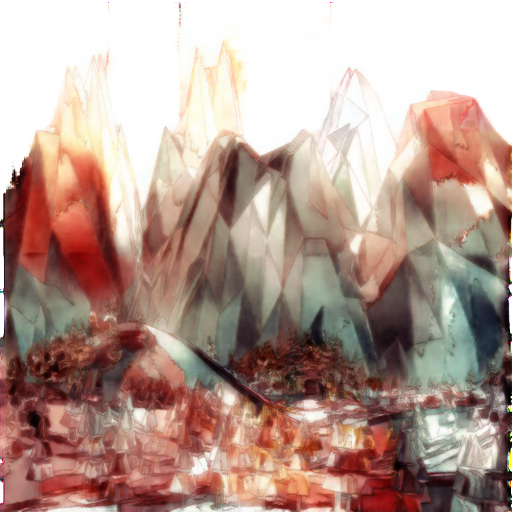}\\
\includegraphics[width=1cm, height=1.5cm]{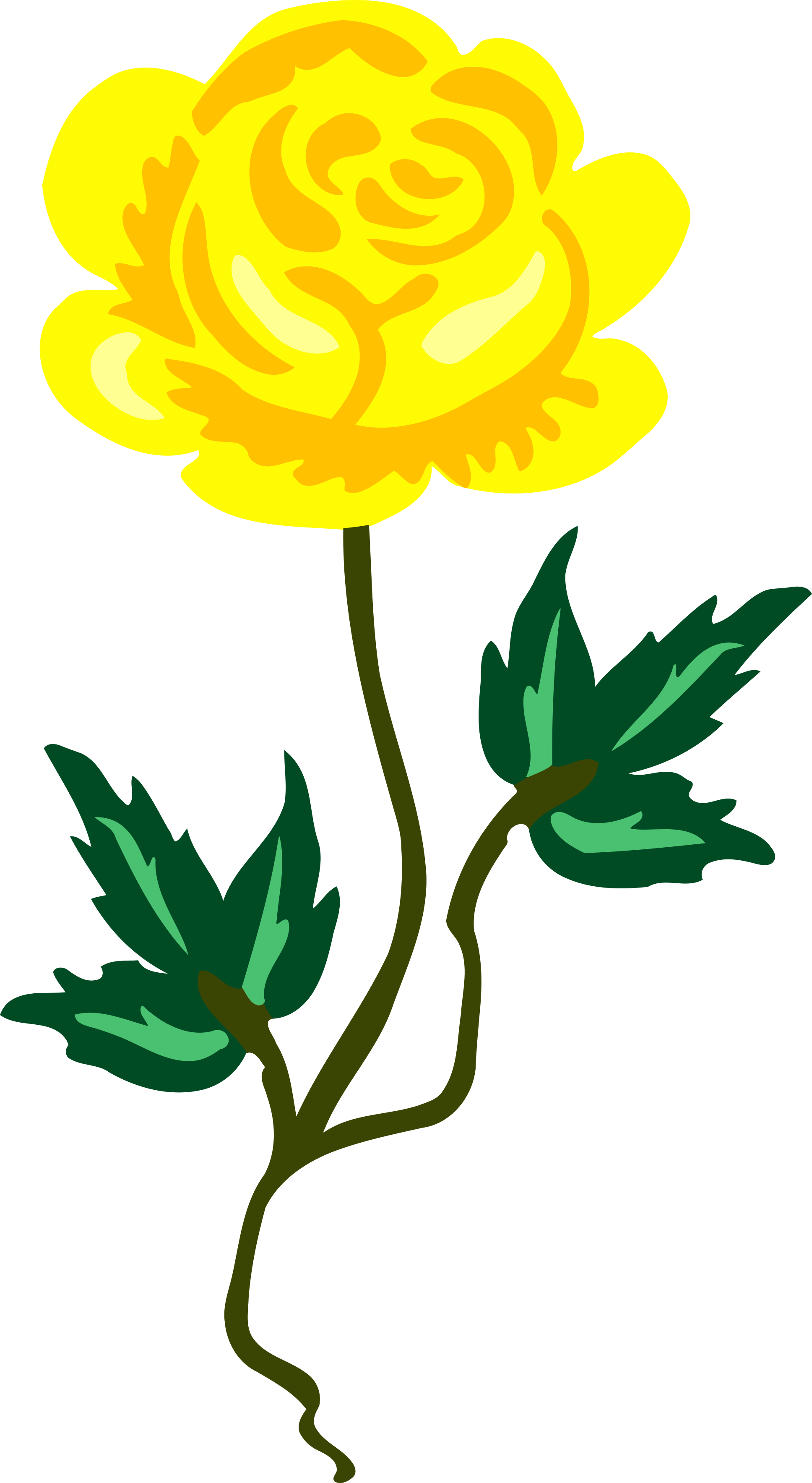} & 
\includegraphics[width=1.6cm, height=1.5cm]{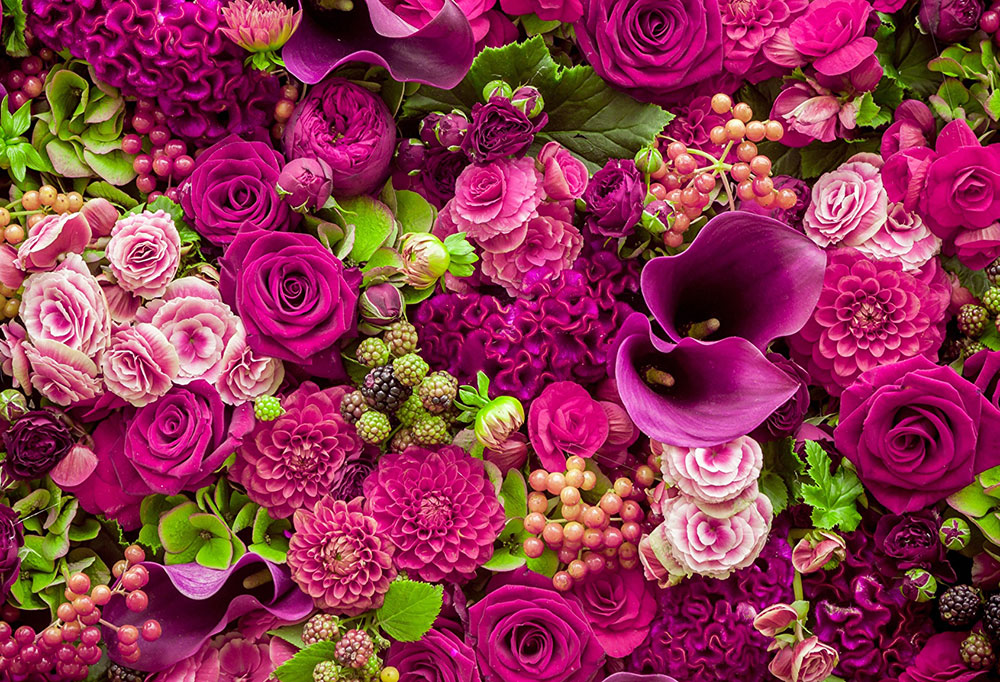} &
\includegraphics[width=1cm, height=1.5cm]{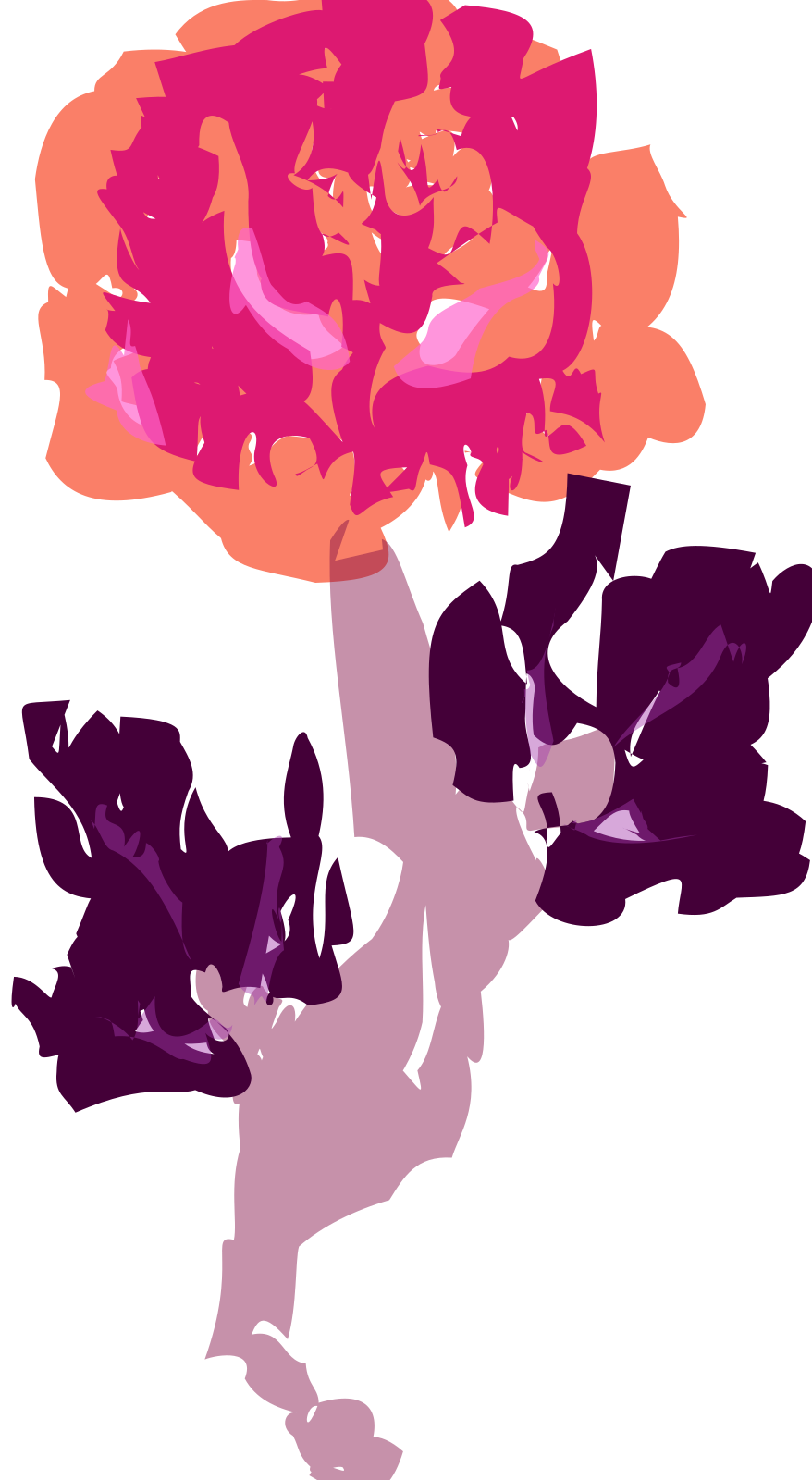} & 
\includegraphics[width=1cm, height=1.5cm]{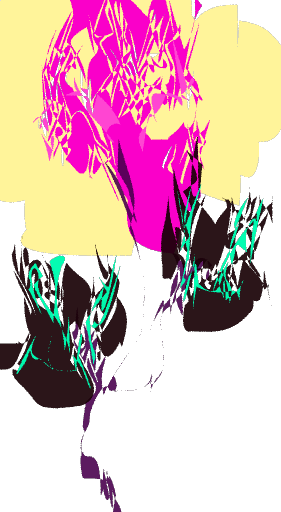} &
\includegraphics[width=1cm, height=1.5cm]{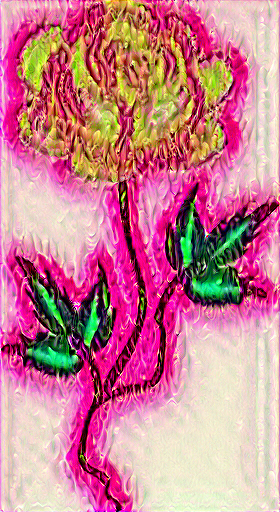} & \includegraphics[width=1cm, height=1.5cm]{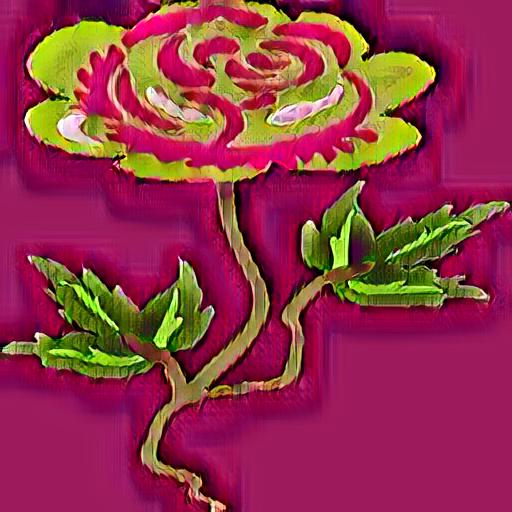} &
\includegraphics[width=1cm,
height=1.5cm]{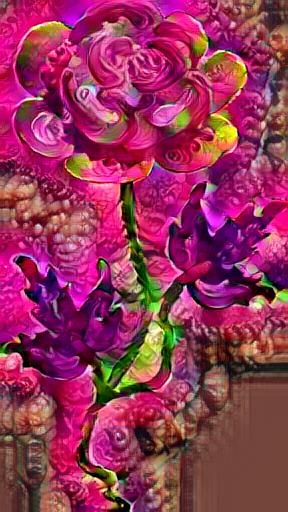} &
\includegraphics[width=1cm, height=1.5cm]{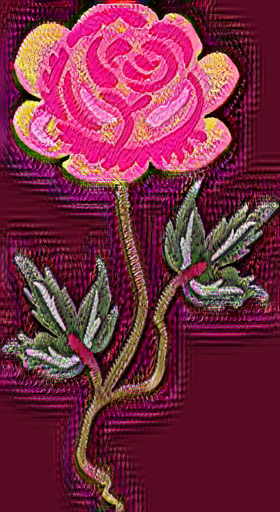} &
\includegraphics[width=1cm, height=1.5cm]{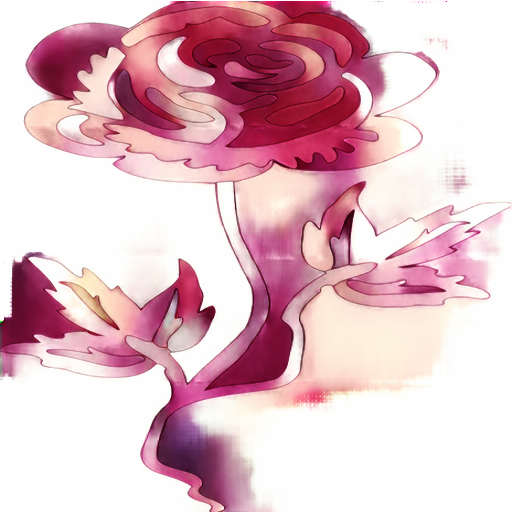}\\

\bottomrule
\end{tabular}
\end{center}
\caption{Qualitative comparisons of style transfer results using different methods}\label{fig:results-comparing}
\end{figure*}

The results of the application of the methods with various style and content image pairs are presented in Fig.~\ref{fig:results-comparing}.

As can be seen from Fig.~\ref{fig:results-comparing}, the Attentioned Deep Paint, SANet, StyTr$^2$, and CAST methods transfer the style but add a lot of artifacts to the images, while losing content patterns. All raster methods make uniform areas non-uniform. 
The StyTr$^2$ method achieves good stylization effects for the owl and hippo images, but at the same time, the stylized images of the tiger and the first landscape contain noticeable artifacts that distort the perception of the content. 
CAST preserve objects' contours, however, it adds unacceptable extra background. 

Although the DiffVG algorithm changes the colors of content images, it blurs the contours or adds distortions, and it cannot convey the style, which is clearly seen in the examples images of a tiger, a hippopotamus, a car, and landscapes. It produces much fewer artifacts, all contours are clear, the pictures are smooth, the color changes (but not everywhere), and the drawing is not transferred, that is, the image content almost does not change. 

Our method seeks a trade-off between following the style and freezing the content. It changes the shape and color of vector primitives to preserve the content as much as possible. The sharpness of the contours does not change.

\subsection{User Study}

We attracted $40$ assessors to evaluate the quality of images generated by VectorNST. We conducted a survey asking participants to assess $10$ images generated by each method on a scale of $1$ to $5$ ($1$~stands for completely inappropriate, $5$ stands for the perfect fit). The images were grouped by method without providing any information about the methods.
Survey results are presented in Tab.~\ref{tab:asessors-scores}.

\begin{table}
  \caption{Comparison of survey results to the proposed VectorNST with DiffVG, Gatys~\etal, StyTR$^2$, SANet, CAST, and Attentioned Deep Paint.}
  \begin{center}
  \begin{tabular}{@{}lc@{}}
    \toprule
    Method & Score\\
    \midrule
    VectorNST (ours) & $0.56 \pm 0.04$\\
    DiffVG & $0.44 \pm 0.05$\\
    Gatys~\etal & $0.42 \pm 0.06$\\
    StyTR$^2$ & $0.62 \pm 0.05$\\
    SANet & $0.43 \pm 0.06$\\
    CAST & $0.59 \pm 0.06$\\
    Attentioned Deep Paint & $0.11 \pm 0.04$\\
    \bottomrule
  \end{tabular}
  \end{center}
  \label{tab:asessors-scores}
\end{table}

\subsection{Time Comparison}

We compare the time required for processing a single image by our method, Gatys~\etal approach, and its implementation for vector graphics in DiffVG. Because three other methods use pre-trained networks, we excluded them from the comparison.

The speed of Gatys~\etal depends only on the size of the content image. On the contrast, the speed of our method and DiffVG depend on (1) the content image size (because how many points need to be sampled during rasterization depends on its size); (2) the number of paths (because when creating an image with contours, the number of paths is important and it determines the size of the image during rasterization); (3) the total number of parameters (the sum of the parameters of all three optimizers).

The results of the time comparison can be found in Tab.~\ref{tab:timings-table}. VectorNST is a bit slower than DiffVG because it spent time on computing the contour loss value. 
Gatys~\etal is considerably faster.

\begin{table}
  \caption{Timings in seconds. Small stands for $256 \times 256$ bitmap images and for vector images with a number of shapes less than $100$. Medium stands for $512 \times 512$ bitmap images and for vector images with the number of shapes less between $100$ and $700$. Big is for bitmap images $1024 \times 1024$ and greater and for vector images with more than $700$ shapes.}
  \begin{center}
  \begin{tabular}{@{}lccc@{}}
    \toprule
    Method & Small & Medium & Big\\
    \midrule
    Gatys~\etal & $1.61$ & $4.14$ & $11.59$\\
    DiffVG & $4.20$ & $26.21$ & $98.57$\\
    VectorNST & $5.93$ & $33.52$ & $112.10$\\
    \bottomrule
  \end{tabular}
  \end{center}
  \label{tab:timings-table}
\end{table}

\subsection{Ablation Study}

\begin{figure*}[h] 
\begin{center}
\begin{tabular}{>{\centering\arraybackslash}m{1.6cm}
>{\centering\arraybackslash}m{1.6cm}
>{\centering\arraybackslash}m{1.6cm}
>{\centering\arraybackslash}m{1.6cm}
>{\centering\arraybackslash}m{1.6cm}
>{\centering\arraybackslash}m{1.6cm}
>{\centering\arraybackslash}m{1.6cm}
>{\centering\arraybackslash}m{1.6cm}
>{\centering\arraybackslash}m{1.6cm}}
    \includegraphics[width=1.6cm, height=1.9cm]{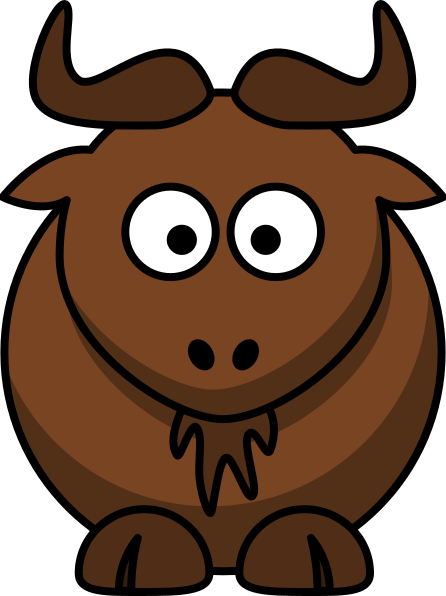}&
    \includegraphics[width=1.6cm, height=1.9cm]{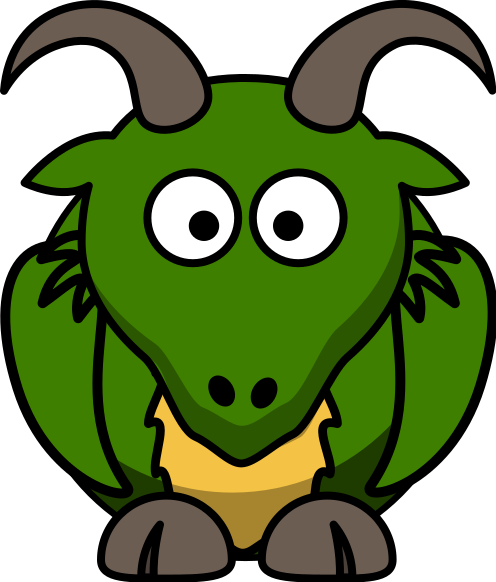}&
    \includegraphics[width=1.6cm, height=1.9cm]{images/content/gnu.png}&
    \includegraphics[width=1.6cm, height=1.9cm]{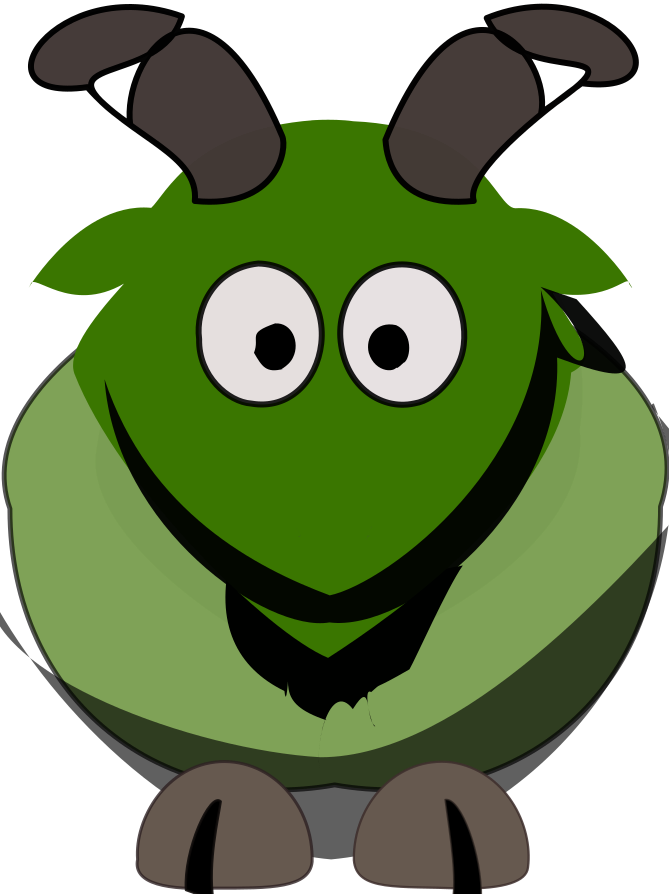}&
    \includegraphics[width=1.6cm, height=1.9cm]{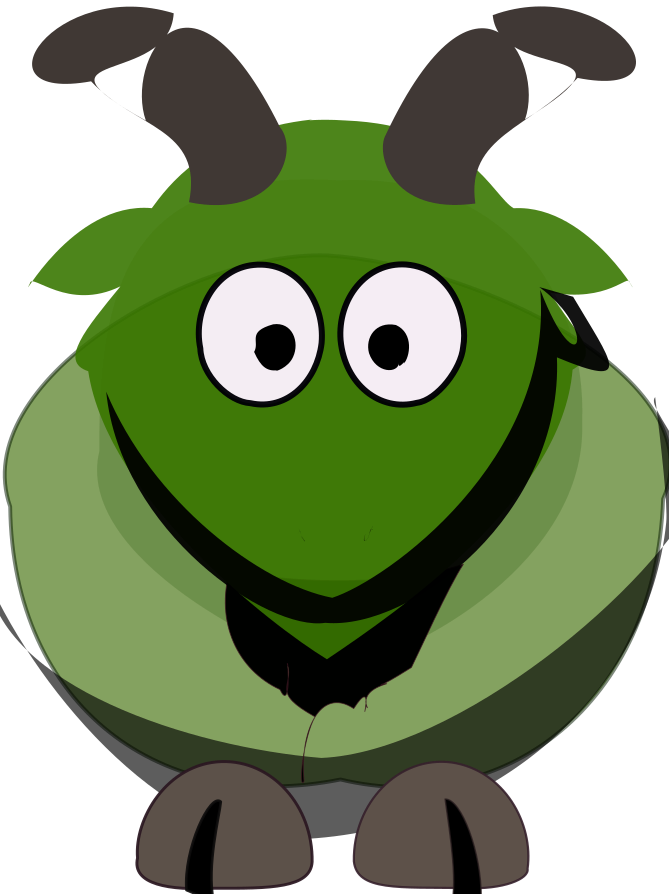}&
    \includegraphics[width=1.6cm, height=1.9cm]{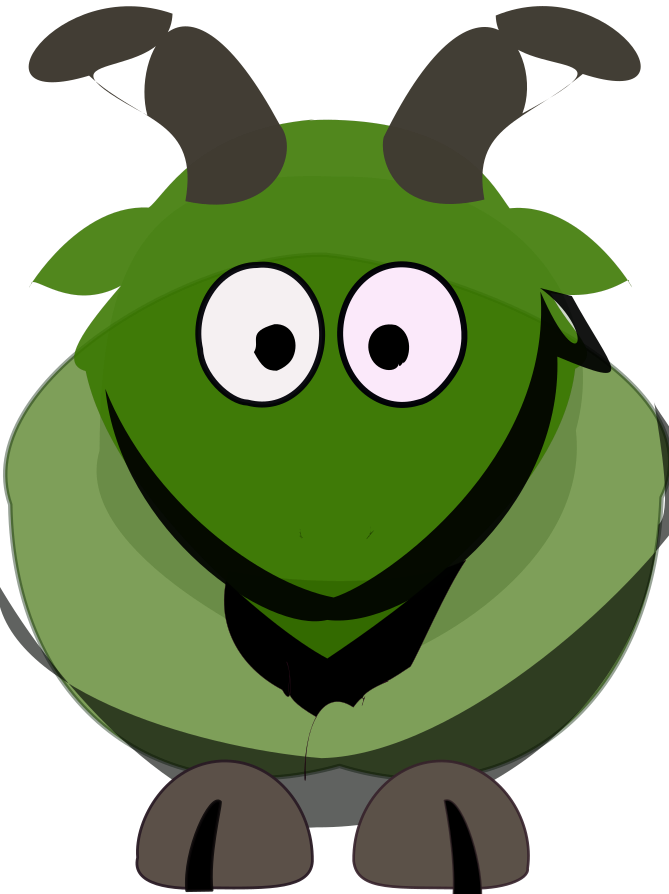}&
    \includegraphics[width=1.6cm, height=1.9cm]{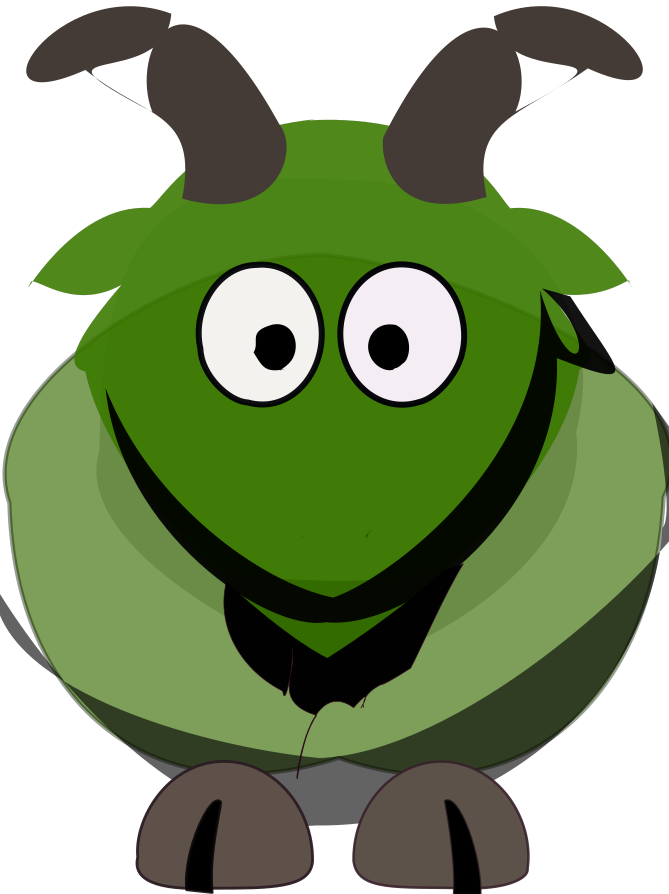}&
    \includegraphics[width=1.6cm, height=1.9cm]{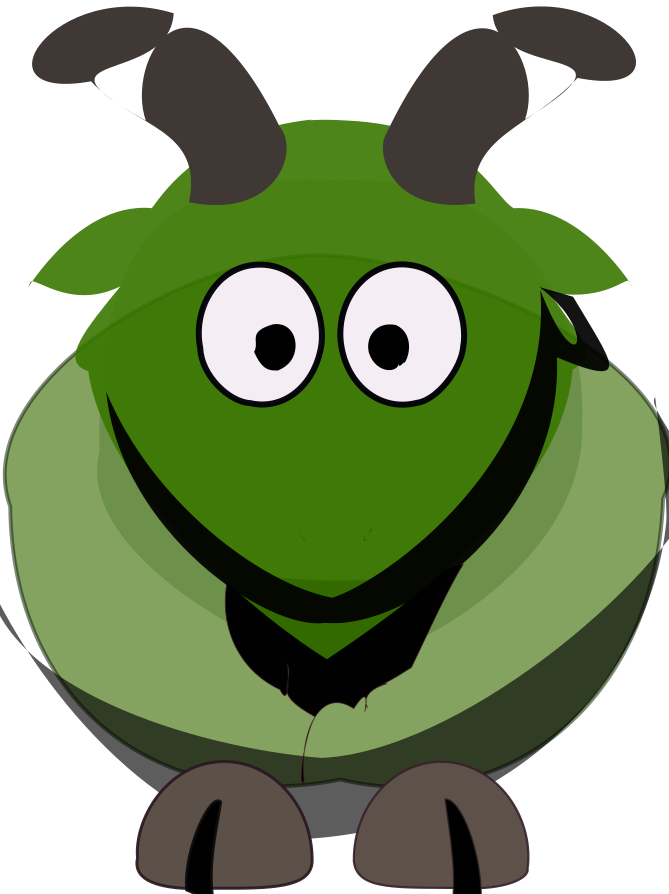}&
    \includegraphics[width=1.6cm, height=1.9cm]{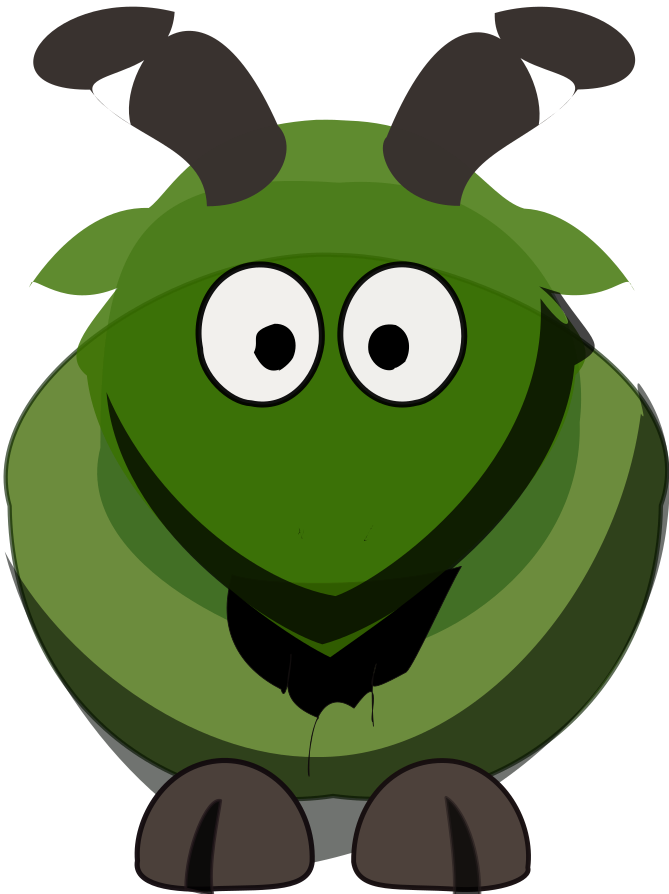}\\
    
    \midrule
    \includegraphics[width=1.6cm, height=1.5cm]{images/content/scene4.png}&
    \includegraphics[width=1.6cm, height=1.5cm]{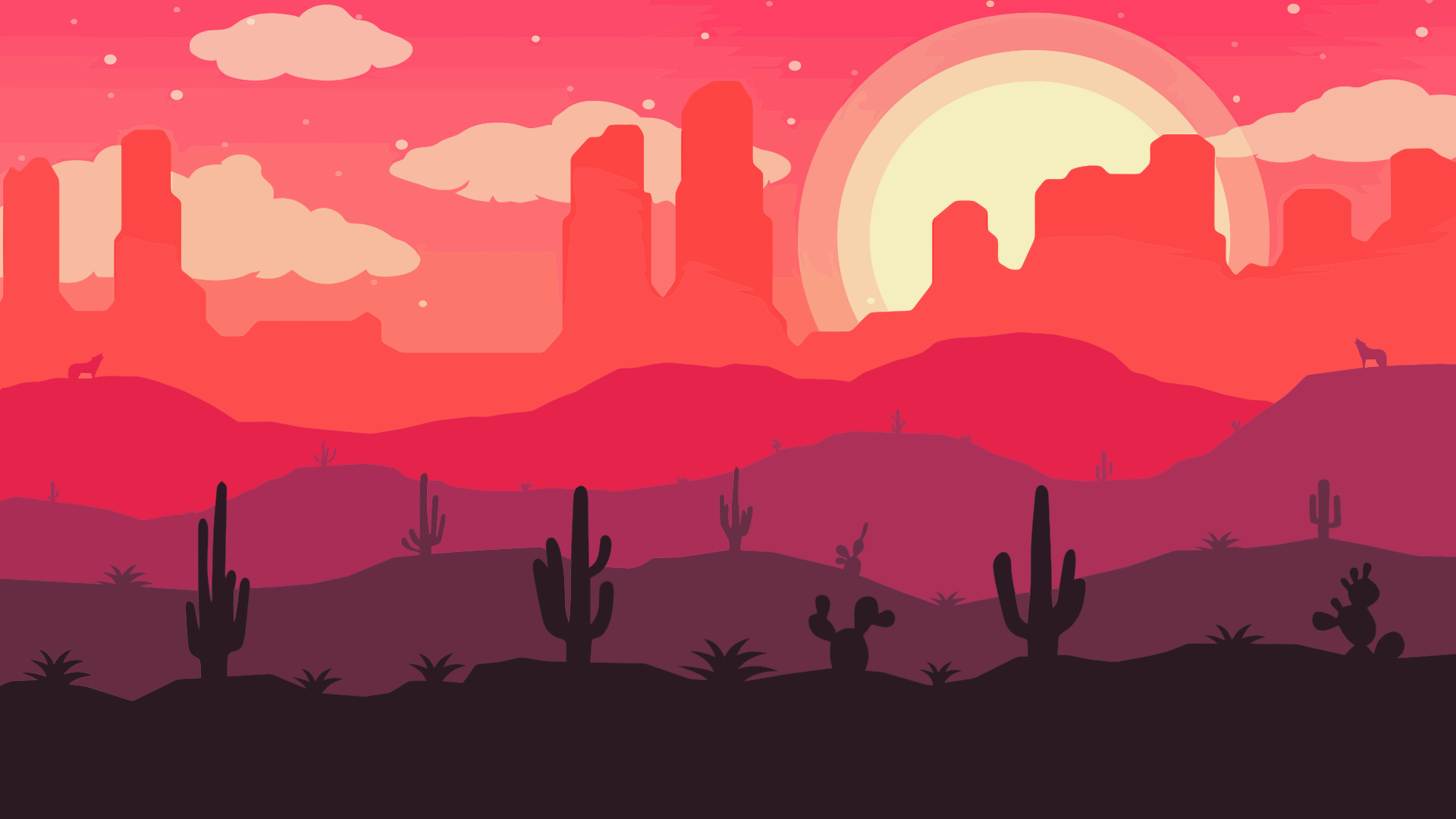}&
    \includegraphics[width=1.6cm, height=1.5cm]{images/content/scene4.png}&
    \includegraphics[width=1.6cm, height=1.5cm]{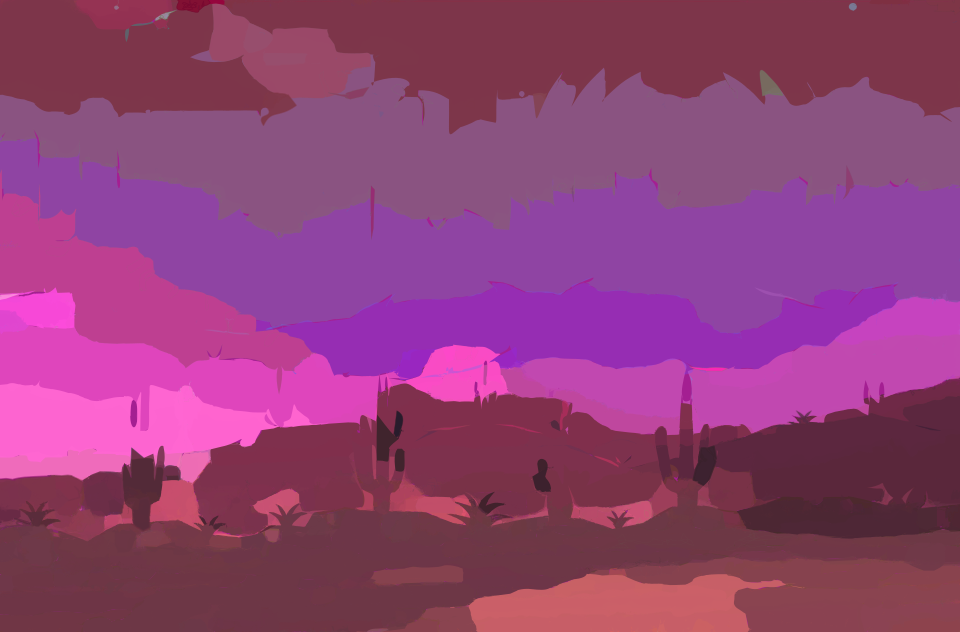}&
    \includegraphics[width=1.6cm, height=1.5cm]{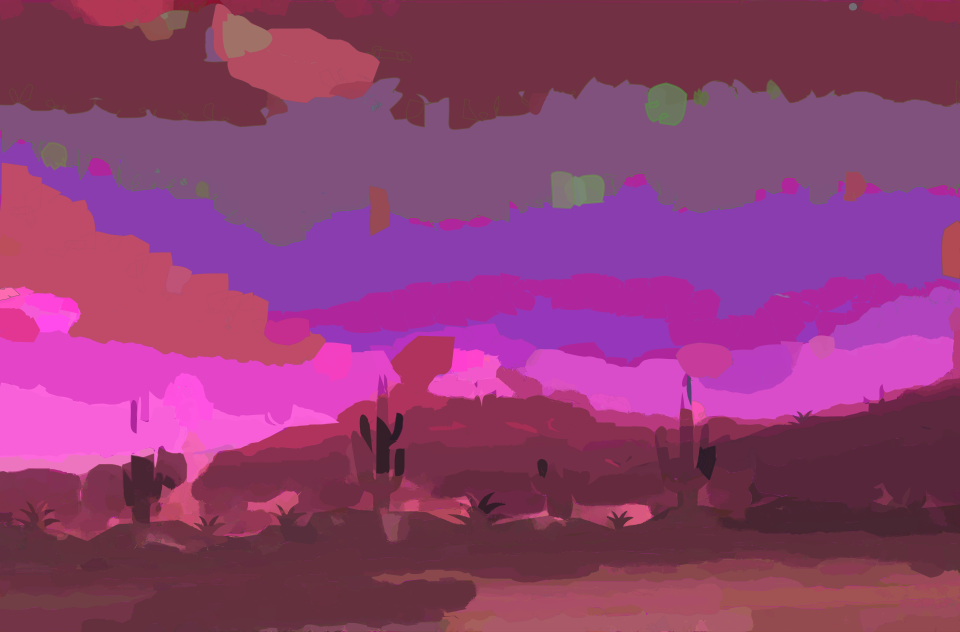}&
    \includegraphics[width=1.6cm, height=1.5cm]{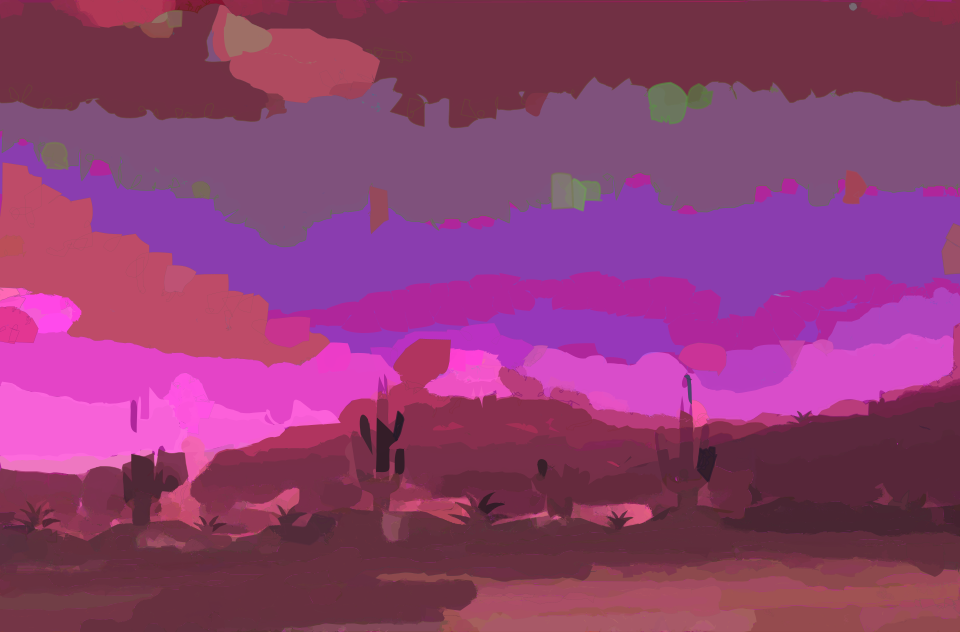}&
    \includegraphics[width=1.6cm, height=1.5cm]{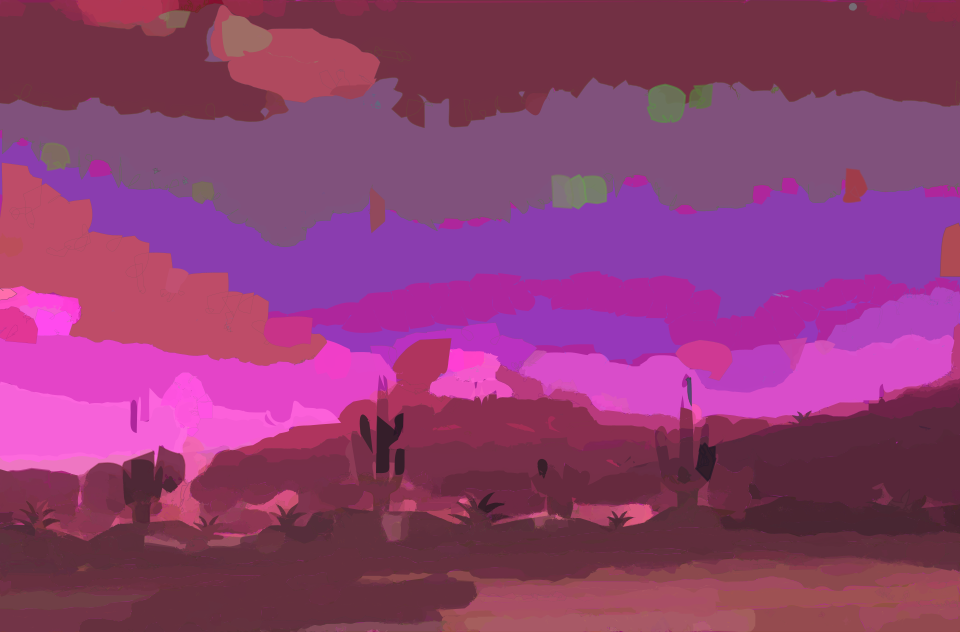}&
    \includegraphics[width=1.6cm, height=1.5cm]{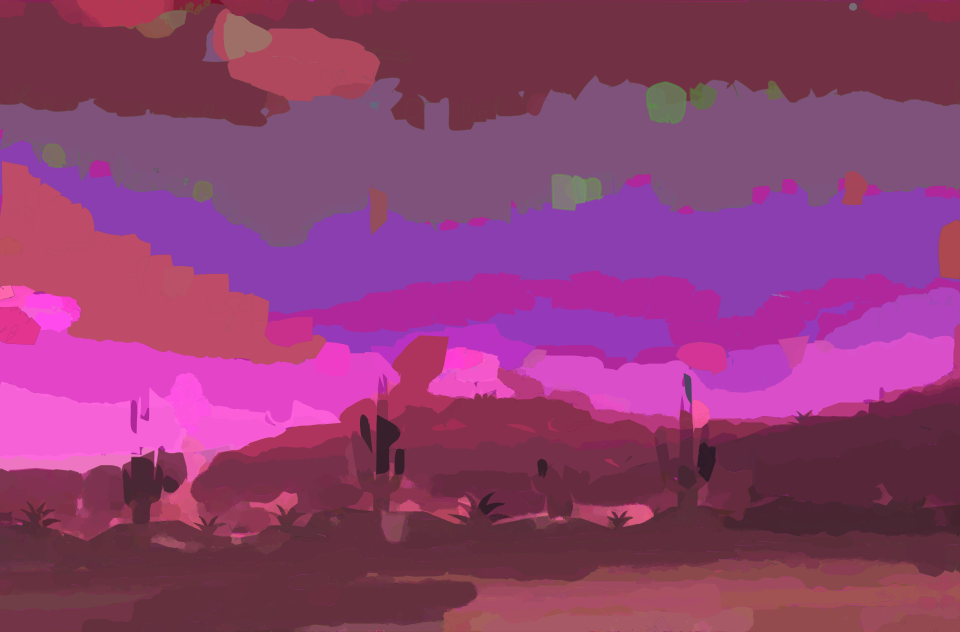}&
    \includegraphics[width=1.6cm, height=1.5cm]{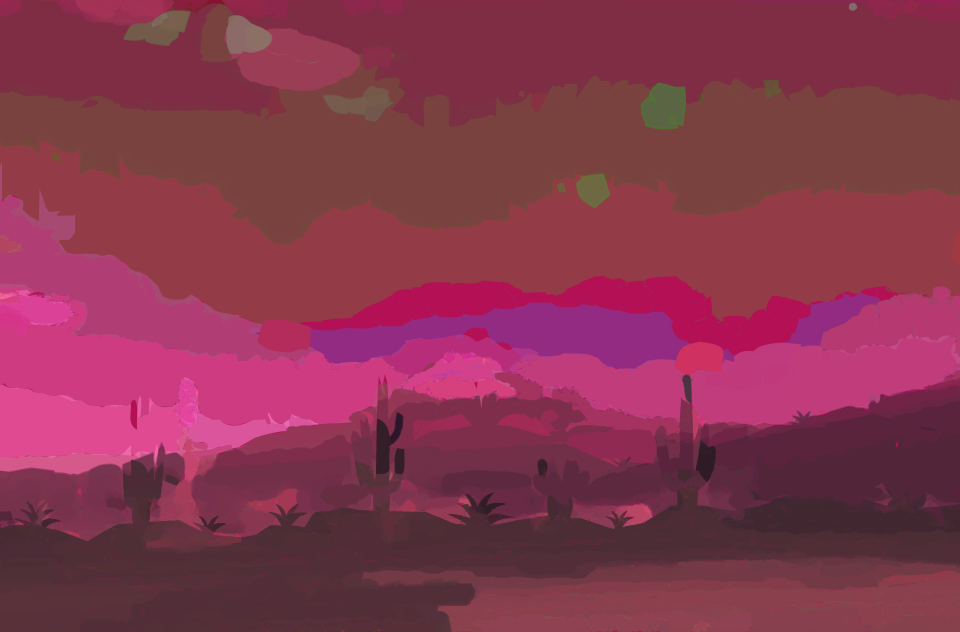}\\
    
    \bottomrule
    Content Image&
    Style Image&
    without $[0,4)$&
    with $[0,4)[4,9)$&
    with $[0,4)[9,16)$&
    with $[0,4)[16,23)$&
    with $[0,4)[23,30)$&
    with $[0,4)[30,36)$&
    without $[4,9)$
\end{tabular}
\end{center}
\caption{Comparison of the effect of selecting different deep embedding subsets on style transfer quality}\label{fig:different-buckets}
\end{figure*}

\textbf{Localization of Style.}
The $37$ deep embeddings were grouped in LPIPS code\footnote{\url{https://github.com/mgharbi/ttools/blob/master/ttools/modules/losses.py}} into consequent intervals with respect to their indices. 
We use this interval structure.
We investigated, which subset of these intervals is the best in terms of preserving styles.
We found out that if we do not take into account deep embeddings from $0$ to $4$ there is no style transfer at all.
The results of our search are shown in Fig.~\ref{fig:different-buckets}. 

We excluded deep embeddings from $4$ to $8$ as they lead to the appearance of broken contours. 
This is exactly the same configuration that was used in LPIPS.

\textbf{Loss functions.}
While conducting experiments with our method, we noted a certain drawback, some erroneous generations of vector shapes. 
Previously, some researchers suggested using contour loss functions. 
For instance, Chen~\etal~\cite{chen2019learning} proposed a segmentation loss function based on the idea of an active contour model, firstly proposed by Kass~\etal~\cite{kass1988snakes}. 
It allowed for obtaining a more accurate segmentation and avoiding ragged edges of the segmentation contour in comparison with the cross entropy loss. 

\begin{figure}[h] 
\begin{center}
\begin{tabular}{>{\centering\arraybackslash}m{1.3cm}
>{\centering\arraybackslash}m{1.3cm}
>{\centering\arraybackslash}m{1.3cm}
>{\centering\arraybackslash}m{1.3cm}
>{\centering\arraybackslash}m{1.3cm}}
\includegraphics[width=1.1cm, height=1.3cm]{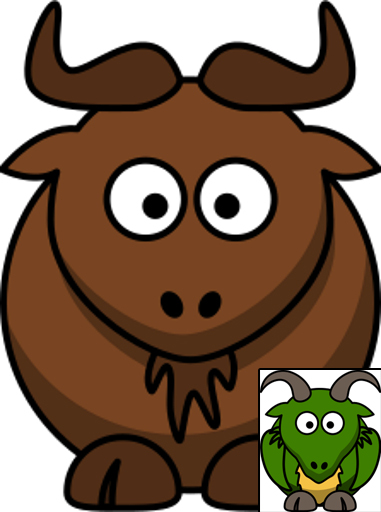} &
\includegraphics[width=1.1cm,  height=1.3cm]{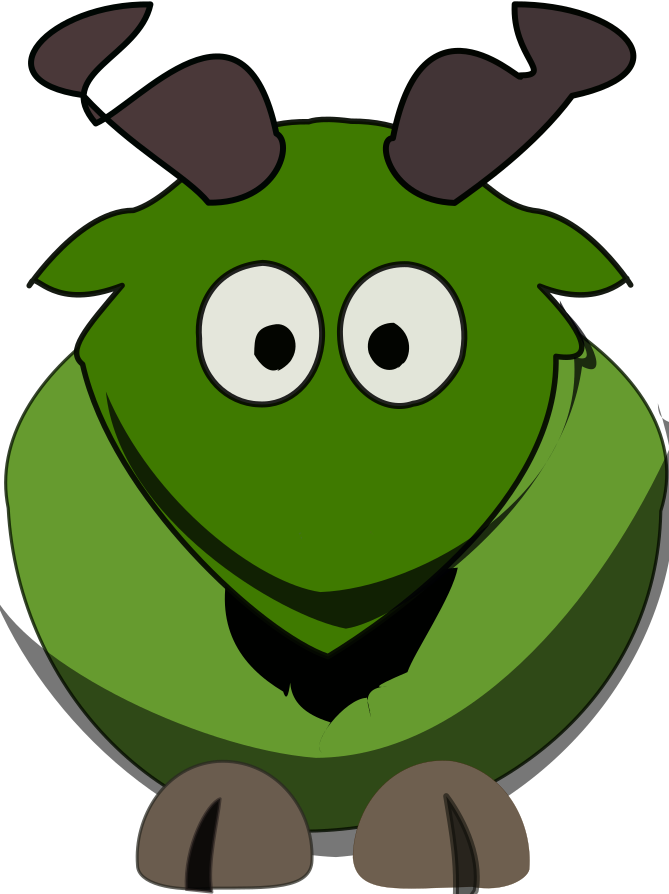} &
\includegraphics[width=1.1cm, height=1.3cm]{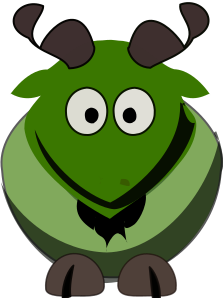} &
\includegraphics[width=1.1cm, height=1.3cm]{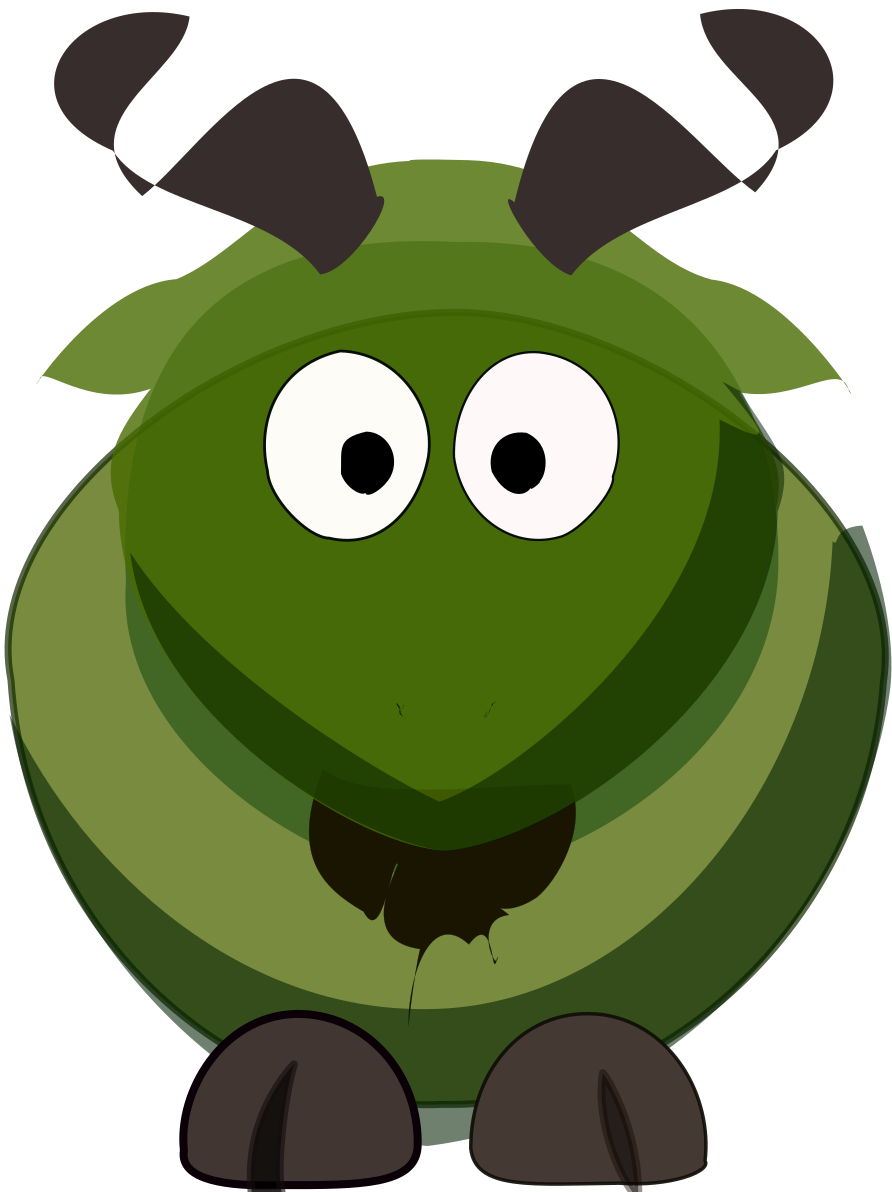} &
\includegraphics[width=1.1cm, height=1.3cm]{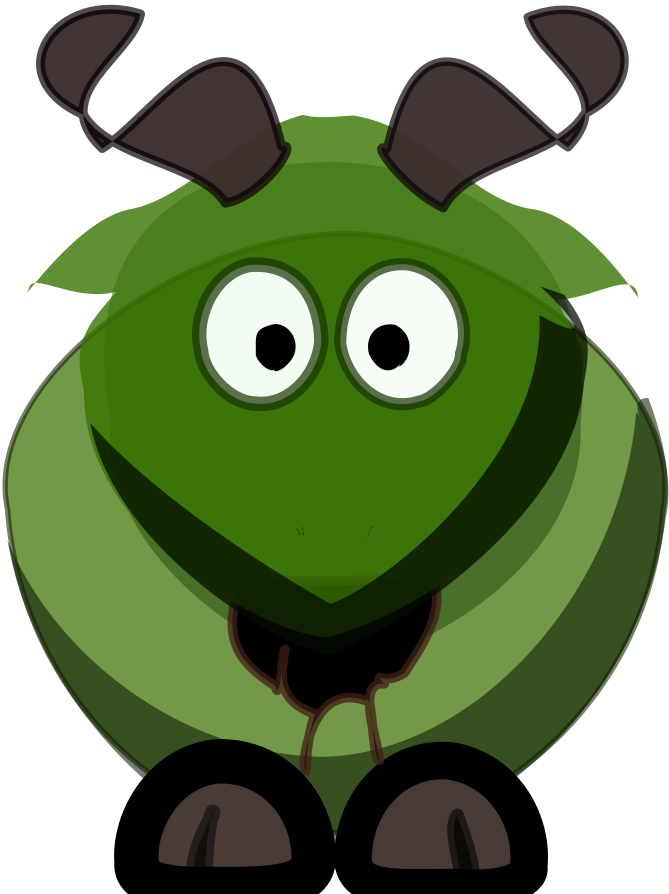}\\

\midrule
\includegraphics[width=1.3cm, height=1.1cm]{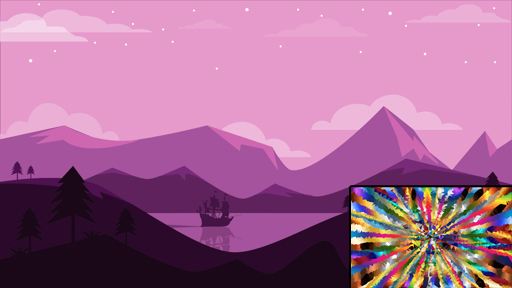} &
\includegraphics[width=1.3cm,  height=1.1cm]{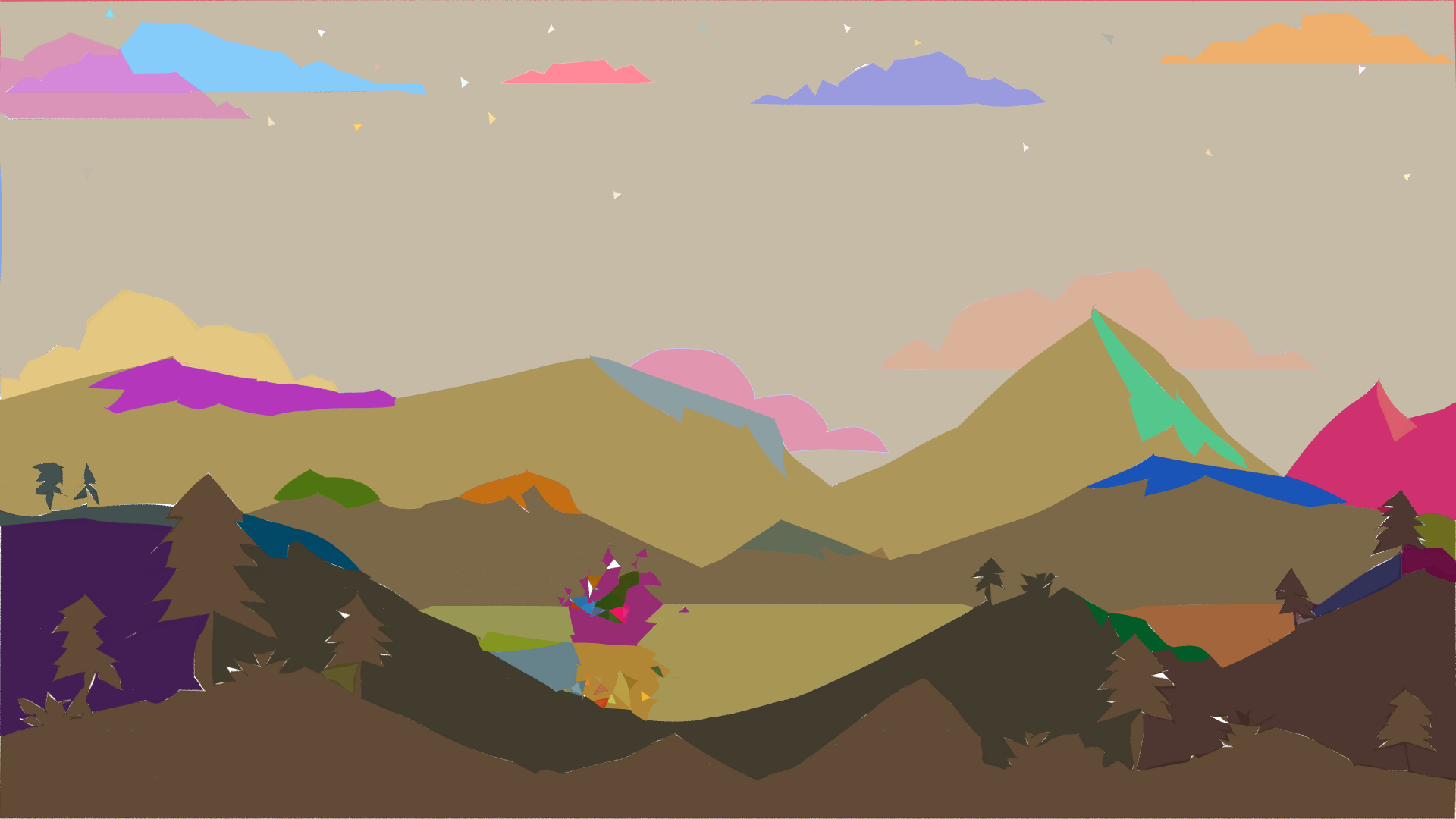} &
\includegraphics[width=1.3cm, height=1.1cm]{images/l1_contour/scene6.png} &
\includegraphics[width=1.3cm, height=1.1cm]{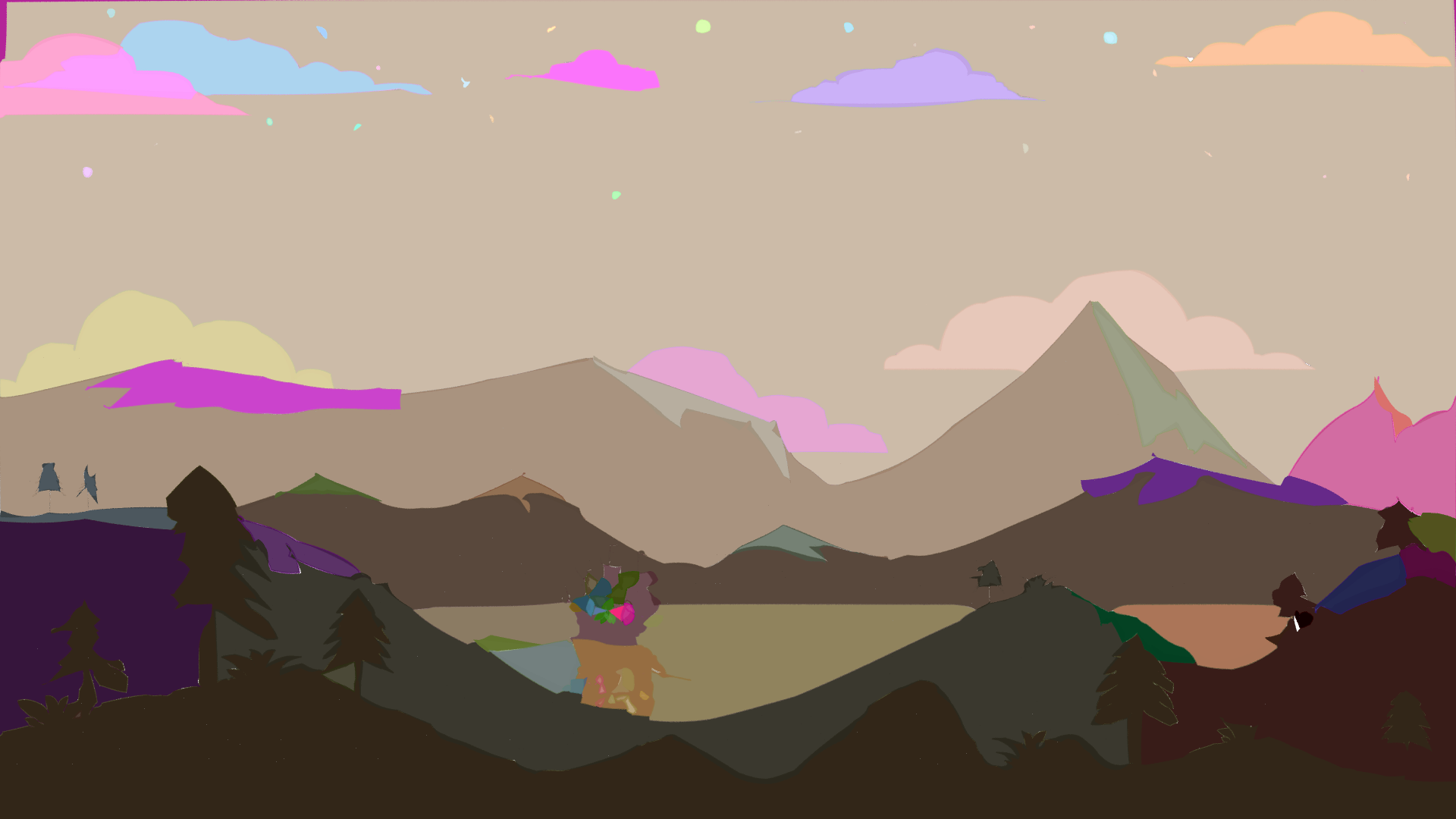} &
\includegraphics[width=1.3cm, height=1.1cm]{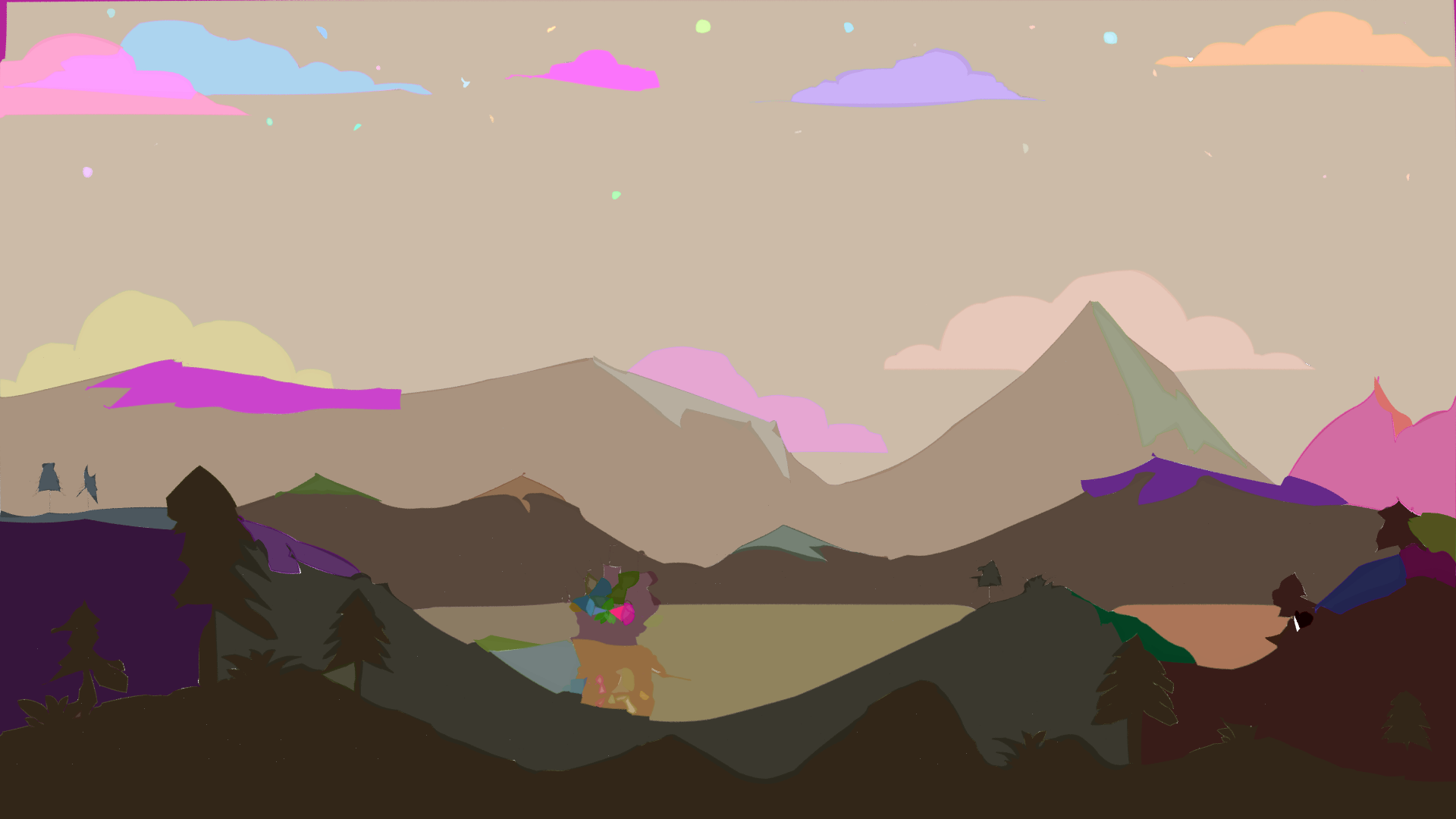}\\

\bottomrule
Input&
W/o $\mathcal{L}_{contour}$&
$L_1$&
$L_2$&
Segment. loss
\end{tabular}
\end{center}
\caption{Comparison of the effect of different loss functions}\label{fig:different-losses}
\end{figure}

We conducted an ablation study involving $L_1$ loss we use in our method, $L_2$ loss, and the segmentation loss. Despite the latter improves the smoothness of the image, however, does not allow achieving a better quality of stylization than $L_1$ loss. 
Examples of stylized images generated with various loss functions are shown in Fig.~\ref{fig:different-losses}. 

As it can be seen, only $L_1$ can properly assign transparency of the regions. The shapes are more complete and do not contain redundant strokes compared to other losses. $L_2$ and the Segmentation loss do not preserve the color brightness. Moreover, the segmentation loss adds redundant strokes. 

\textbf{Content Leak Study.}
Style transfer methods are known to suffer from the so-called ``content leak'' problem, which is losing substantial information about the content in the resulting image~\cite{deng2022stytr2}.
To evaluate to which degree this is the problem for vector style transfer, we compared the content leak between $3$ contour losses after several rounds of the various stylizations.

In the first setting, we transferred the style of a content image only once and then applied the style of the original image.
In the second setting, we first applied $9$ consequent transfers to various styles different from the style of the original image. Then we applied $9$ consequent transfers to the original image style.

\begin{figure}[h] 
\begin{center}
\begin{tabular}{>{\centering\arraybackslash}m{1.3cm}
>{\centering\arraybackslash}m{1.3cm}
>{\centering\arraybackslash}m{1.3cm}
>{\centering\arraybackslash}m{1.3cm}
>{\centering\arraybackslash}m{1.3cm}}

    \includegraphics[width=1cm, height=1.2cm]
    {images/content/gnu.png} &
    \includegraphics[width=1cm,  height=1.2cm]{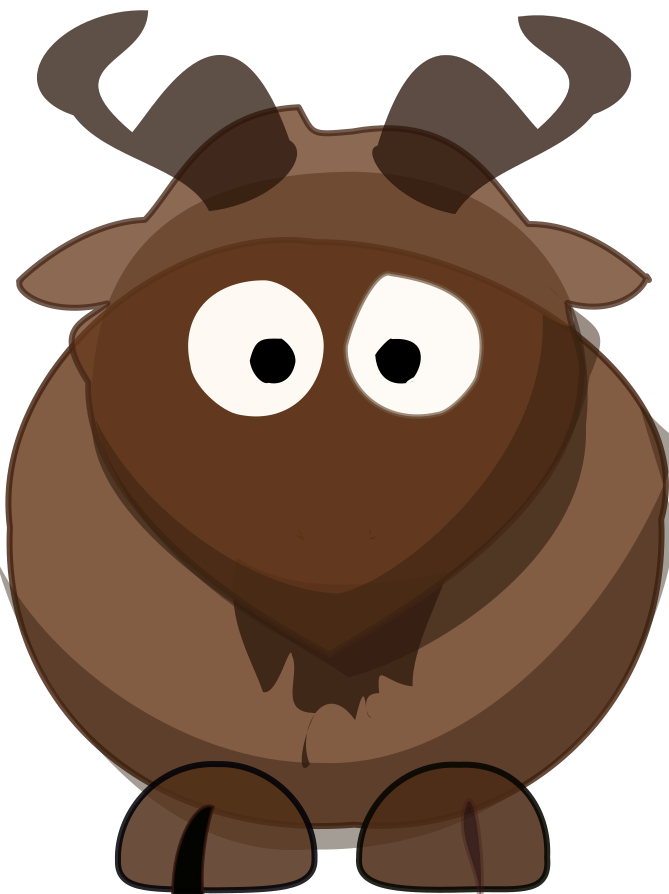} &
    \includegraphics[width=1cm, height=1.2cm]{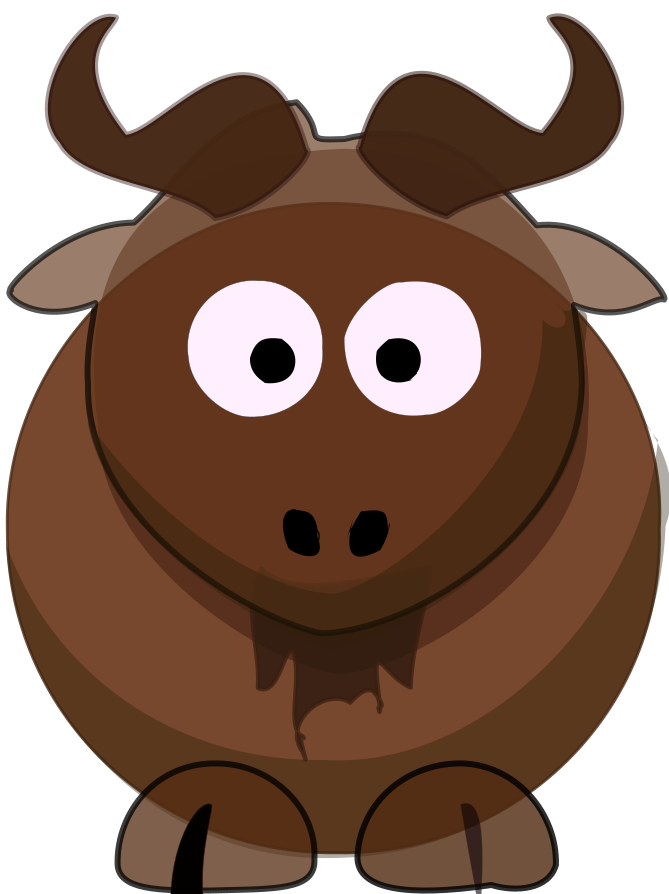} &
    \includegraphics[width=1cm, height=1.2cm]{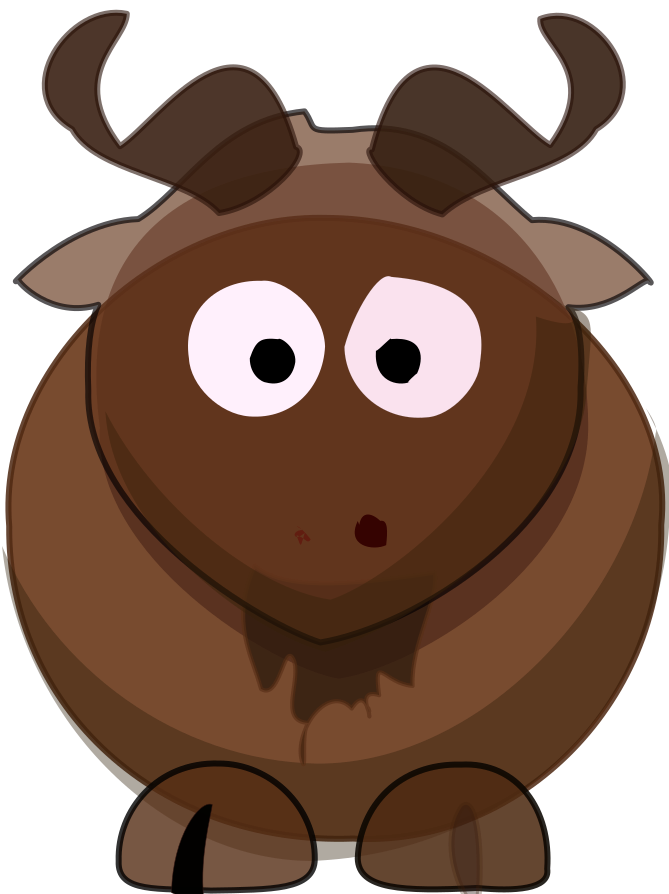} &
    \includegraphics[width=1cm, height=1.2cm]{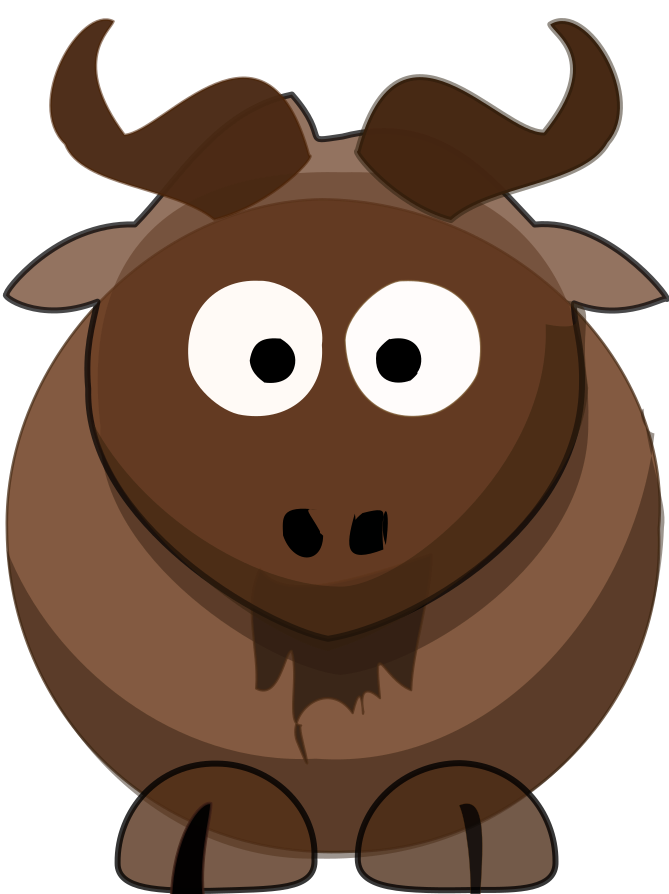}\\  
    \midrule
    \includegraphics[width=1cm, height=1.2cm]
    {images/content/gnu.png}&
    \includegraphics[width=1cm,  height=1.2cm]{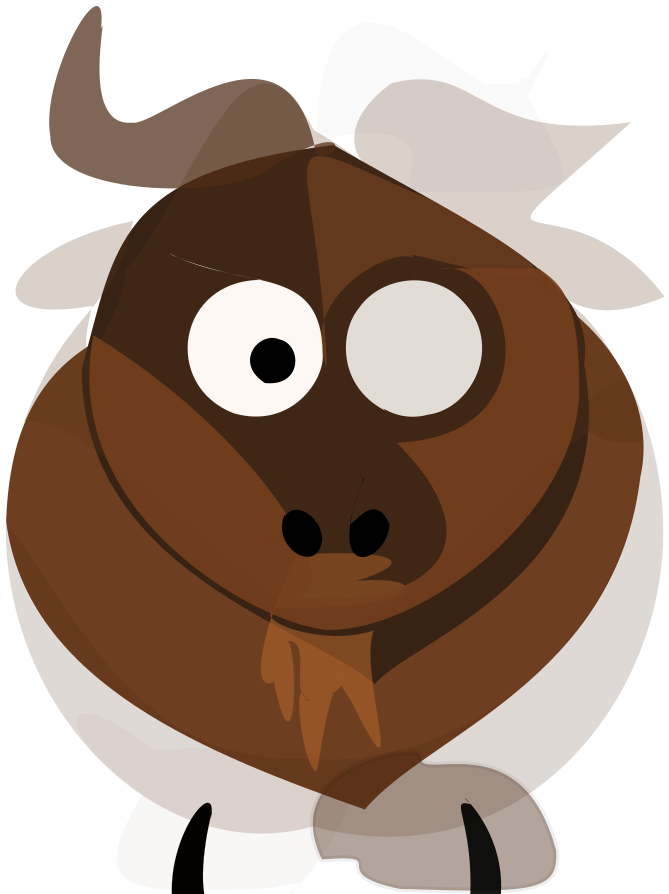} &
    \includegraphics[width=1cm, height=1.2cm]{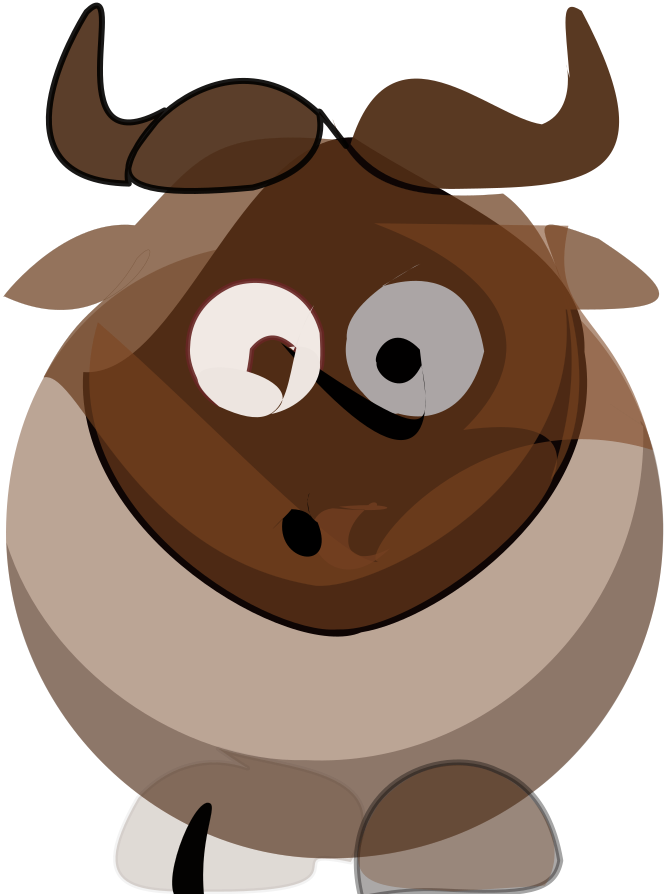} &
    \includegraphics[width=1cm, height=1.2cm]{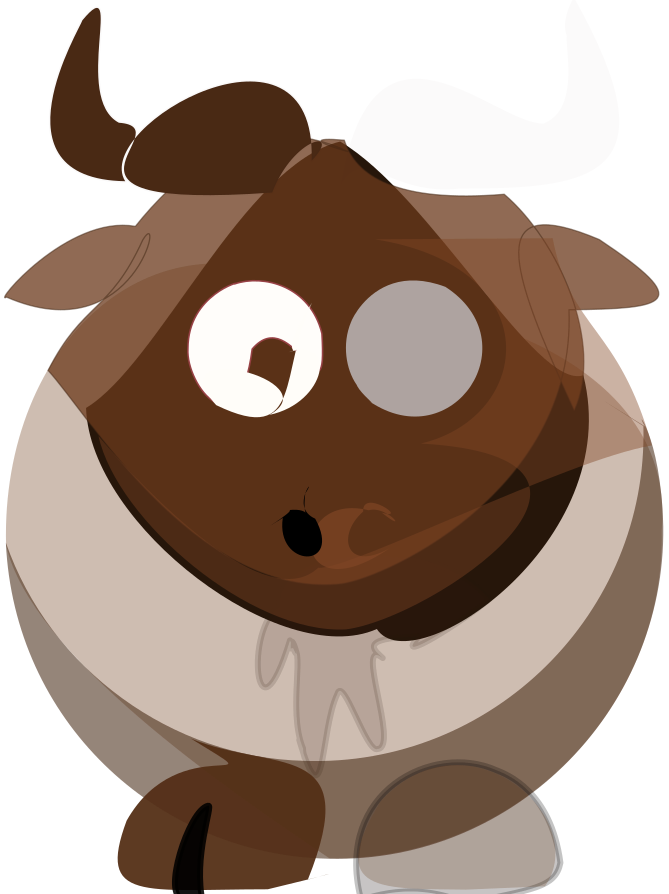} &
    \includegraphics[width=1cm, height=1.2cm]{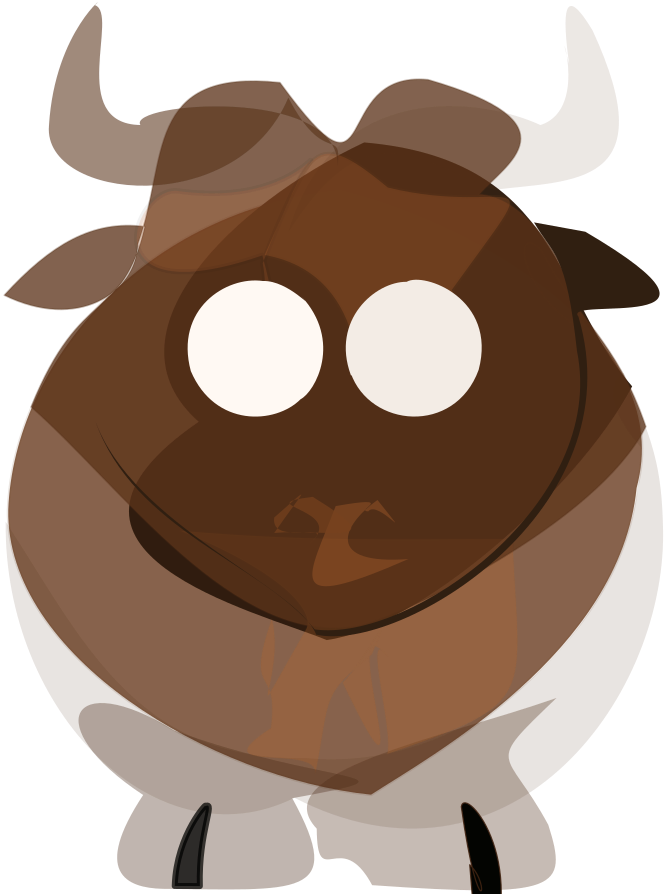}\\
    \bottomrule
    Content Image&
    W/o $\mathcal{L}_{contour}$&
    $L_1$&
    $L_2$&
    Segment. loss
\end{tabular}
\end{center}
\caption{Comparison of content leak between various contour regularizations}
\label{fig:multiple-style-transfer}
\end{figure}

The result of the first experiment are presented in the first line, and the result of the second experiment are presented in the second line of Fig.~\ref{fig:multiple-style-transfer}.

As can be seen, in the single round scenario, $L_1$ and the Segment Loss perform almost identically and preserve the most details of the image. However, in the nine rounds scenarios, $L_1$ remains the only loss that preserves the most details of the original image and keeps it recognizable.

\section{Discussion}\label{sec:lim}

\subsection{Suitability of Content Loss for Vector Images} 
\begin{figure}[!h]
\begin{center}
\begin{tabular}{
>{\centering\arraybackslash}m{1.3cm}
>{\centering\arraybackslash}m{1.3cm}
>{\centering\arraybackslash}m{1.3cm}
>{\centering\arraybackslash}m{1.3cm}
>{\centering\arraybackslash}m{1.3cm}}
    \toprule
    & &
    Gatys~\etal step $100$ &
    Our method step $151$ &
    DiffVG step $30$\\
     &  &
    \includegraphics[width=1.4cm, height=1.4cm]{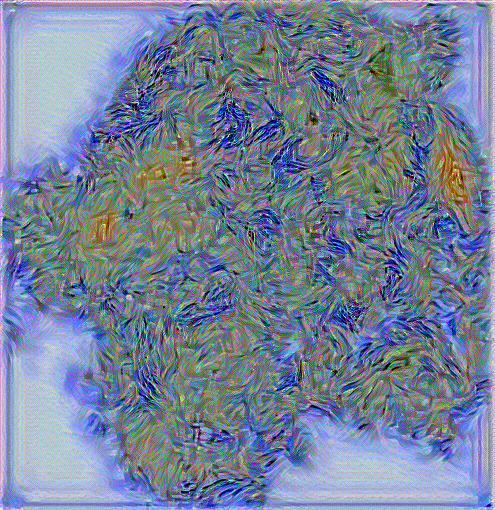} &
    \includegraphics[width=1.4cm, height=1.4cm]{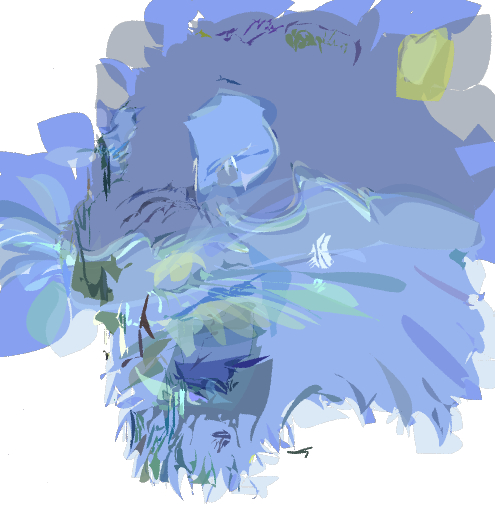} &
    \includegraphics[width=1.4cm, height=1.4cm]{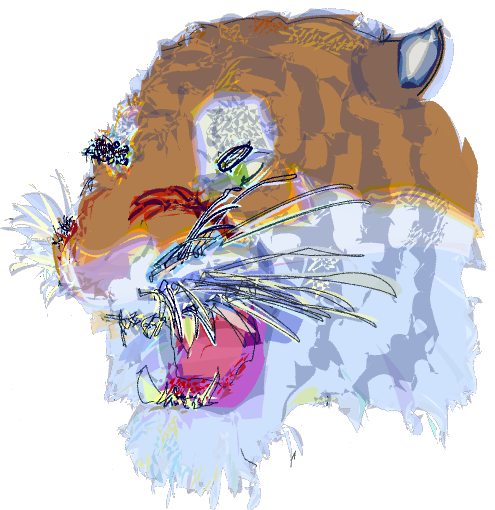}
    \\
    Content Losses &
    \includegraphics[width=1.4cm, height=1.4cm]{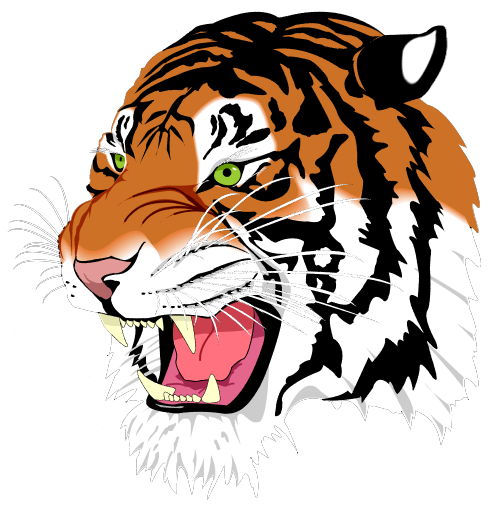} &
    58.99 &
    54.78 &
    61.46
    \\
    Style Losses &
    \includegraphics[width=1.4cm, height=1.4cm]{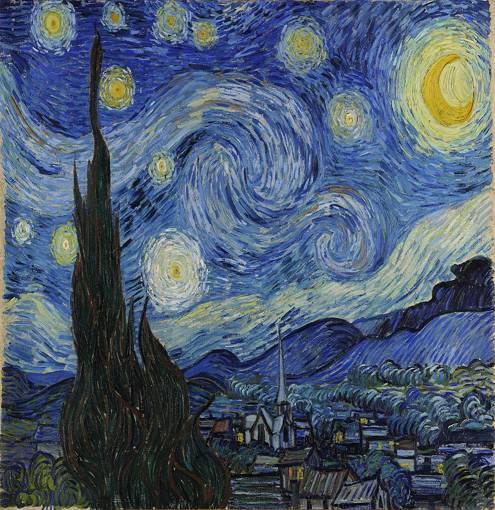} &
    67.49 &
    22662.99 &
    7841.42
    \\
    \bottomrule
\end{tabular}
\end{center}
\caption{Analysis of content and style losses of different images} 
\label{fig:bad-losses-table}
\end{figure}

StyTr$^2$ paper provides an assessment based on content and style losses. 
We intended to compare methods for vector style transfer in the same way, but it turned out that this approach is not applicable in this domain. 
In Fig.~\ref{fig:bad-losses-table}, some samples generated by Gatys~\etal and DiffVG methods are depicted, which implicitly use style and content losses for training. 
With frozen weights and hyperparameters of the methods, we calculated these losses and came to the conclusion that the values of these loss functions are unrepresentative and difficult to explain. 
In addition, we have revealed cases when an image containing a lot of noise and clearly not having the style of a given style image received a very low error in contrast to reliably high-quality images having a significantly higher estimate of the loss function. 
This is because the values of loss functions do not always correspond precisely to the quality of output images~\cite{li2018literature}.
Therefore, as mentioned earlier, the evaluation of NST methods still remains an open problem.


\subsection{Vectorization of raster stylized images}\label{subsec:rtv}
Special attention should be paid to the possibility of obtaining a vector image by generating a bitmap image with subsequent vectorization. This approach leads to many problems. 
First, bitmap images usually contain a lot of details, thus, their vectorizations consist of a large number of shapes, which in total begin to require a lot of memory and extensively consume resources during rendering. 
Second, as a result, such vector images become very complex for designers to edit and post-process. 
Third, in a case when the content image contains a single object on a white or transparent background but the style image does not, bitmap approaches, unlike our method, often add an unacceptable extra background. For example, this happened when transferring the style to a tiger and a flower, as can be seen in Fig.~\ref{fig:results-comparing}.
It is noteworthy that raster methods for NST do not sufficiently change the shapes of figures and objects, because they do not have the ability to semantically separate objects and directly influence their shapes.


\subsection{Limitations}
\textbf{DiffVG limitations.}
Our work is largely based on the DiffVG method and therefore fully inherits the constraints and limitations of this differentiable rasterizer. Here are the most important ones.
(1) One of the main limitations is the impossibility of optimizing the topology of a vector image, namely the number of B\'ezier curves and their segments. 
In particular, it is impossible to choose the number of shapes in the vector image, because this is a discrete choice that cannot be optimized. 
(2) It is not possible either to group/combine/rearrange shapes and it is impossible to choose the type of generated shape. 
(3) DiffVG does not fully meet the SVG specification, because it cannot optimize a text tag, a radial gradient, or other tags.
(4) DiffVG takes as an input the parameters of vector shapes of the entire image and outputs a bitmap image in the RGBA format. The loss function we use is based on the VGG network, which receives an RGB image as input. Therefore, it becomes necessary to transform the RGBA image obtained after rasterization into RGB. There is no standard differentiable method for this operation. 
Any transition from RGBA to RGB format may cause the loss of information, which leads to the necessity of using a modified image during optimization, not the original one.

\textbf{Feature extractor limitations.}
The method proposed is based on a convolutional network (VGG-19) as a feature extractor, which was trained on bitmap images from ImageNet. 
Therefore, it cannot learn vector images peculiarities including large outlined segments of the same color, well-defined object contours, transparent background, or a smaller percentage of tiny image details.  

Another essential limitation to mention is that it was trained on RGB images, while DiffVG processes RGBA images. We thus lose image opacity.

\section{Conclusion}\label{sec:concl}
In this paper, we proposed a novel neural style transfer method for vector graphics, VectorNST, which allows processing illustrations such as sketchy animals, cars, and landscapes.   
We introduced a loss function consisting of two parts, an adapted LPIPS loss and a contour loss, the latter providing more accurate style transfer and content information preservation.
Experimental results demonstrated that our method generates gorgeous stylized vector images and achieves higher human assessment results compared to SANet, Attentioned Deep Paint, and DiffVG methods.

Further improvement of our method would include adding a transformer-based model for more accurate preservation of the vector image contours. Another direction would be to overcome the limitation rooted in DiffVG by making the model capable of changing the input parameters of a number of curves or anchor points via backpropagation.

Another direction of future work may be collecting a vector image dataset for improving style transfer inference time as it can be done offline using a pre-trained style network.




{\small
\bibliographystyle{ieee_fullname}
\bibliography{egbib}
}

\end{document}